\documentclass[10pt,twocolumn,letterpaper]{article}
\pdfoutput=1
\usepackage{cvpr}
\usepackage{times}
\usepackage{epsfig}
\usepackage{graphicx}
\usepackage{amsmath}
\usepackage{amssymb}
\usepackage{subfig}
\usepackage{epsfig}
\usepackage{multirow}
\usepackage{graphbox}

\usepackage{hyperref}
\cvprfinalcopy 

\def\cvprPaperID{4851} 

\ifcvprfinal\fi
\begin{document}

\title{3D Appearance Super-Resolution with Deep Learning}

\author{Yawei Li$^1$, Vagia Tsiminaki$^2$, Radu Timofte$^1$, Marc Pollefeys$^{2,3}$, and Luc van Gool$^1$\\
$^1$Computer Vision Lab, ETH Zurich, Switzerland\\
{\tt\small \{yawei.li, radu.timofte, vangool\}@vision.ee.ethz.ch}\\
$^2$Computer Vision and Geometry Group, ETH Zurich, Switzerland, $^3$Microsoft, USA\\
{\tt\small \{vagia.tsiminaki, marc.pollefeys\}@inf.ethz.ch}
}

\maketitle

\begin{abstract}
{We tackle the problem of retrieving high-resolution (HR) texture maps of objects that are captured from multiple view points. In the multi-view case, model-based super-resolution (SR) methods have been recently proved to recover high quality texture maps.  On the other hand, the advent of deep learning-based methods has already a significant impact on the problem of video and image SR. Yet, a deep learning-based approach to super-resolve the appearance of 3D objects is still missing. The main limitation of exploiting the power of deep learning techniques in the multi-view case is the lack of data. We introduce a 3D appearance SR (3DASR) dataset based on the existing ETH3D~\cite{schops2017multi}, SyB3R~\cite{ley2016syb3r}, MiddleBury, and our Collection of 3D scenes from TUM~\cite{goldlucke2014super}, Fountain~ \cite{Zhou-Koltun-TOG-2014} and Relief~\cite{Zollhofer-et-al-TOG-2015}. We provide the high- and low-resolution texture maps, the 3D geometric model, images and projection matrices. We exploit the power of 2D learning-based SR methods and design networks suitable for the 3D multi-view case. We incorporate the geometric information by introducing normal maps and further improve the learning process. Experimental results demonstrate that our proposed networks successfully incorporate the 3D geometric information and super-resolve the texture maps.}
\end{abstract}

\begin{figure}[t]
\begin{center}
  \includegraphics[width=1\linewidth]{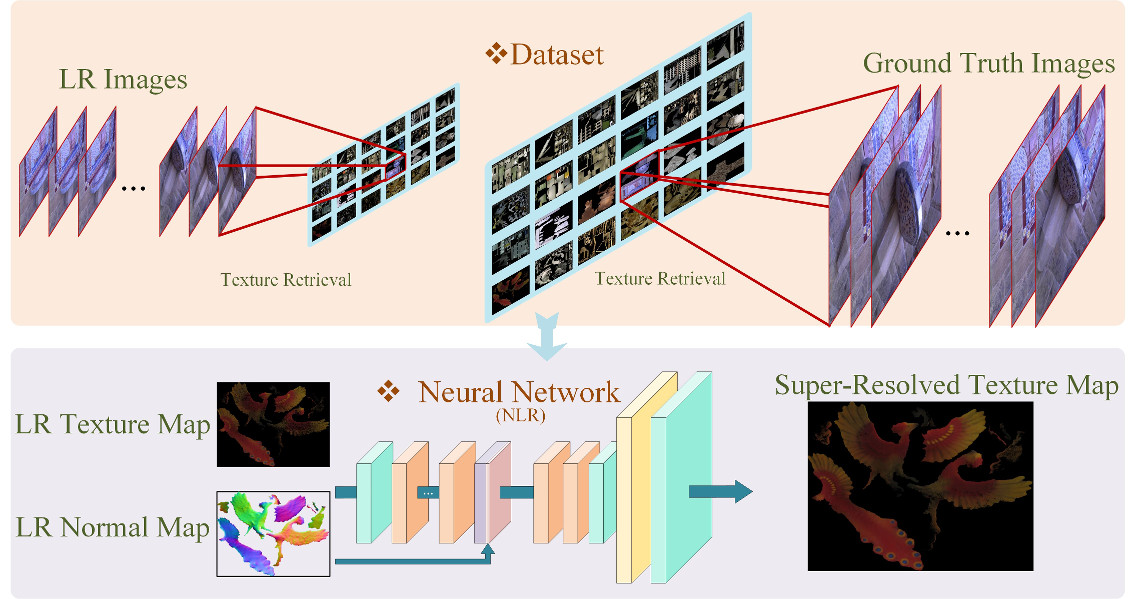}
\end{center}
\vspace{-0.4cm}
\caption{We introduce the 3DASR, a 3D appearance SR dataset  and a deep learning-based approach to super-resolve the appearance of 3D objects.}
\vspace{-0.6cm}
\label{fig:contribution}
\end{figure}

\section{Introduction}
\label{sec:introduction}

Retrieving efficiently the appearance information of objects through multi-camera observations is of a great importance for the final goal of creating realistic 3D content. To increase the realism of the reconstructed 3D object a detailed appearance needs to be added on top of geometry. This high quality 3D content is used in applications such as movie production, video games and digital culture heritage preservation. Yet, even with highly accurate 3D geometric reconstruction, simply re-projecting the images onto the geometry does not guarantee detailed appearance coverage.

To regain details from the low-resolution (LR) images, model-based super-resolution (SR) techniques have been introduced in the multi-view case~\cite{goldlucke2009superresolution,goldlucke2014super,tsiminaki2014high}. These methods introduce a single coherent texture space to define a common texture map and they
model the captured image as a downgraded version of this high-resolution (HR) texture map. Through image formation model they exploit the visual redundancy of the overlapping views~\cite{goldlucke2009superresolution,goldlucke2014super} and of video frames~\cite{tsiminaki2014high}. Although these model-based SR techniques recover successfully high quality texture maps, they are computationally demanding.

On the other hand, 2D example-based SR methods have be shown to outperform the model-based methods.  The basic assumption of example-based SR is the recurrence of similar patches in different parts of an image or in different images~\cite{freeman2002example}. In particular, recent deep learning-based techniques have been proposed to learn the mapping between the LR and HR images.
Different networks are trained on large image datasets that contain pairs of HR and LR images. Super-resolving LR images is then realized with a feed forward step. Yet, a deep learning-based approach to super-resolve the appearance of 3D objects is still missing.

In this paper, our goal is to introduce deep learning techniques into the problem of appearance SR in the multi-view case. To exploit the capacity of  2D deep learning techniques, we first provide a 3D appearance dataset. Similar to the model-based SR methods, we introduce a common texture space and define a single coherent texture map. This texture map is first mapped onto the geometry. Then the textured surface is projected into the image space. We express the concatenation of these two mappings through the image formation model (Fig.~\ref{fig:image_formation_model}). Through this image generation process and using captured images of multiple scaling factors we can then recover the corresponding texture maps. We provide a  dataset that contains ground truth HR texture maps together with LR texture maps of down-scaling factor $\times2$, $\times3$, and $\times4$. The dataset covers both synthetic scenes SyB3R~\cite{ley2016syb3r} and real scenes ETH3D~\cite{schops2017multi}, MiddleBury, and our Collection of 3D scenes from TUM~\cite{goldlucke2014super}, \emph{Fountain}~\cite{Zhou-Koltun-TOG-2014} and \emph{Relief}~\cite{Zollhofer-et-al-TOG-2015}. We then leverage the capacity of 2D learning-based methods~\cite{lim2017enhanced}
and design two architectures suitable for the 3D multi-view case. Similar to~\cite{lahner2018deepwrinkles} we introduce normal maps to capture the local structure of the 3D model and incorporate the 3D geometric information into the 2D SR network. To our knowledge, our work is the first that introduces deep learning approaches for the appearance SR in the multi-view case. Using our provided dataset, we evaluate  different texture map SR methods including interpolation-based, model-based, and learning-based. In summary, the contributions of our paper are:
\begin{enumerate}
    \item a 3D texture dataset that contains pairs of HR and LR textures of 3D objects. With this dataset we facilitate the integration of deep learning techniques into the problem of appearance SR in the multi-view case and we open up a promising novel research direction. We refer to the dataset as 3DASR.
    \item the first appearance SR framework that elegantly combines the power of 2D deep learning-based techniques with the 3D geometric information in the multi-view setting.
\end{enumerate}
The rest of the paper is organized as follows. Sec.~\ref{sec:related_works} introduces related works of this paper. Sec.~\ref{sec:texture_retrieval_model} describes how the texture maps are retrieved. Sec.~\ref{sec:dataset} explains the generation process of the dataset. Sec.~\ref{sec:learning_methods} explores the introduction of normal information into neural networks to super-resolve LR texture maps. Sec.~\ref{sec:results} shows the evaluation results of different methods. Sec.~\ref{sec:conclusion} concludes the paper.

\section{Related Works}
\label{sec:related_works}

\subsection{2D image super-resolution}
\label{subsec:super_resolution}

2D image SR has been extensively studied and it can be classified into three categories, i.e.\ interpolation-based, model-based, and example-based{~\cite{park2003super, farsiu2004fast, yang2010image, freeman2002example}. Although a comprehensive review of these methods is beyond the scope of this paper, we present the underlying concepts of each of them. Interpolation-based methods~\cite{allebach1996edge,li2001new} increase the resolution by computing pixel values using the neighbouring information. But leveraging only the local information within the image cannot guarantee the recovery of high-frequency details. Model-based approaches  
describe the LR image as downgraded version of the HR image and express analytically the forward degradation system. Solving for the inverse problem prior knowledge over the unknown HR image such as smoothness and non-local similarity~\cite{buades2005non,li2017modified} is imposed. Treating the problem as a stochastic process, maximum likelihood~\cite{farsiu2004fast} or maximum a posterior~\cite{fu2016frequency} approach is followed. Although these methods successfully recover high-frequency details, they require elegant optimization techniques. Most of the times they correspond to iterative approaches that are computationally heavy and time-consuming. Learning-based methods shift this computational burden to the learning phase and using the trained network they super-resolve the image through a feed forward step. Due to the availability of large datasets, carefully designed network architectures can learn the mapping from LR to HR image
and achieve state-of-the-art performance~\cite{dong2014learning,timofte2014a+,ledig2017photo,lim2017enhanced,li2018carn,zhang2018image}.}
Our work, introduces deep learning-based approach in the multi-view case to retrieve the fine texture of 3D objects.

\subsection{Texture retrieval}
\label{texture_retrieval}

Adding a high quality texture layer onto the 3D geometry plays an essential role in the final realism. This is a challenging step since in the multi-view case there are additional sources of variation that we need to account for, namely occlusions, calibration and reconstruction inaccuracies. Several methods have been proposed in the literature~\cite{heckbert1986survey} to efficiently exploit all the available color information and to address the aforementioned challenges. 

\textbf{Single view selection.} 
To cope with different geometric inaccuracies, several methods use only one view to assign texture to each face.  Lempitsky and Ivanon~\cite{lempitsky2007seamless} compensate for seams between the boundaries of each face by solving a discrete labeling problem. Gal~\etal~\cite{gal2010seamless} incorporate in their optimization the effect of foreshortening, image resolution, and blur by modifying the weighting function. Waechter~\etal~\cite{waechter2014let} 
add an additional smoothness term to  penalize inconsistencies between adjacent faces. By choosing a single view, these methods disregard the multiple color information that exists in the multi-view setting.

\textbf{Multi-view selection.}
To leverage the multiple color information over views, several methods blend the images for each face. Debevec~\etal~\cite{debevec1996modeling} reproject and blend view contributions according to visibility and viewpoint-to-surface angle. To capture  view dependent shading effects
Buehler~\etal~\cite{buehler2001unstructured}  model and approximate the plenoptic function for the scene object. Some hybrid approaches~\cite{allene2008seamless,chen20123d} select a single view per face and blend in frequency space views close to texture patch borders. To correct geometric inaccuracies, in~\cite{zhou2014color} camera poses are jointly optimized with the photometric consistency. Following the success of patch-based synthesis methods, Bi~\etal propose a single view-independent texture mapping method that account for geometric misalignment~\cite{bi2017patch}. Generally these methods do not exploit efficiently viewpoint visual redundancy.

\textbf{Multi-view texture SR methods.}
To retrieve fine appearance details, a handful of texture SR methods have leveraged the SR principle in the multi-view case and compute texture maps  with a resolution higher than the input images~\cite{koch1998multi,nakamura2000generation}. Goldl\"ucke~\etal 
introduce an image formation model  to super-resolve texture maps~\cite{goldlucke2009superresolution} and  to refine the geometry and camera calibration~\cite{goldlucke2014super}. Tsiminaki~\etal\cite{tsiminaki2014high} further improve SR texture quality by exploiting additional temporal redundancy and by uniformly correcting calibration and geometry errors with optical flow. These methods are however computationally expensive.

We alleviate the limitations of these model-based SR by introducing the deep learning-based approaches that have been proven to outperform in the 2D case.

\subsection{Super-resolution benchmark}
\label{subsec:SR_benchmark}

In order to be able to use deep learning-based techniques for super-resolving the texture of 3D objects, datasets need to be available. For 2D image SR there are several benchmarking datasets Set5~\cite{bevilacqua2012low}, Set14~\cite{zeyde2010single}, Urban100~\cite{Huang-CVPR-2015}, BSD100~\cite{MartinFTM01} and works~\cite{yang2014single,kohler2017benchmarking}.
ImageNet~\cite{deng2009imagenet} has been also used as training dataset in several example based approaches~\cite{dong2014learning,dong2016image}. More recently, DIV2K dataset was introduced to provide higher quality images~\cite{Agustsson_2017_CVPR_Workshops}.

Such data are however not available in the multi-view case. We propose in this work a methodology to compute textures of several resolution and we provide a 3D texture dataset, 3DASR, that contains pairs of HR and LR textures of 3D objects.

\section{Texture Retrieval}
\label{sec:texture_retrieval_model}

\subsection{Image formation model}
\label{subsec:image_formation}

\begin{figure}[t]
\begin{center}
  \includegraphics[width=1\linewidth]{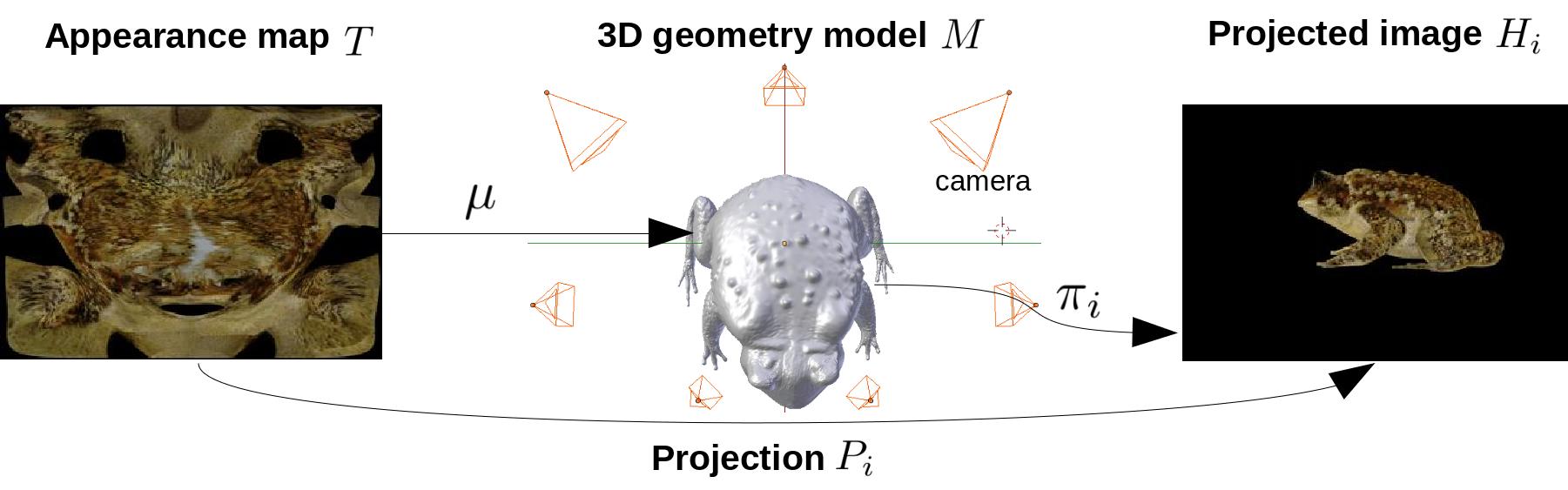}
\end{center}
\vspace{-0.4cm}
\caption{Image formation model.}
\vspace{-0.6cm}
\label{fig:image_formation_model}
\end{figure}

The image formation model simulates the generation of the image from the unknown texture map. In Fig.~\ref{fig:image_formation_model}, we can distinguish two steps~\ie, texture mapping and projection to image space.
\vspace{-0.4cm}

\paragraph{Texture mapping} The texture mapping function
$\mu$ assigns each entity of the texture map (texel) to a 3D point of the geometry. In order to be able to define  the texture map and the mapping, we first need to parameterize the geometry in a common space.  We assume that the 3D model $M$ is a known triangulated mesh and thus we can define any UV parameterization. In
~\cite{balmelli2002space} advanced algorithms that result in space-optimized texture maps are discussed. In this work we use a fixed UV parameterization, described in Subsec.~\ref{subsec:real_subsets}. Through this mapping function $\mu$, a texel $x$ is mapped to a point $\mu(x)$ of the 3D mesh model $M$.
\vspace{-0.4cm}

\paragraph{Projection to image space}We assume that we know the camera poses and the intrinsic camera parameters. The textured 3D object is then projected into the image space given the known projection matrices.
Let $\pi_{i}$ be the camera projection matrix at the view point $i$ and $H_{i}$ the corresponding image of resolution $h \times w$. The geometric point $\mu(x)$ is projected to the pixel location $(\pi_{i} \circ \mu)(x)$ in the image plane. Let $T ^{h \cdot w}$ and  $H_i^{h \cdot w}$ be the vectorized version of the texture map and the projected image. The image is then expressed as a linear combination of the texture map $H_i^{h \cdot w} = P_{i} T ^{h \cdot w}$ where $P$ is a matrix of dimension $h\cdot w\times h\cdot w$. To estimate this projection operator several issues need to be addressed. First, two geometric points of the surface might be projected into the same location due the convexity of the geometry and then only the visible color value needs to be selected. Second, this projection step can lead to non-integer locations~\cite{li2016multiview}. Third, the distribution of the projected points in the image space is non-uniform, which means that the points may be sparse for some areas.
To combine the contributions of all the projected 
the projected points falling into the neighborhood of a pixel $q$ we introduce the Gaussian function as the weighting function. This function takes the location proximity into account, encouraging pixels near the center of $q$ while penalizing those far way from $q$. By combining the contributions of of all projected  points falling into the neighborhood of a pixel with this Gaussian function we solve for the sparse areas in the image space that can originate due to high curvature regions of the surface.

We retrieve the texture maps by inverting the image formation model. We examine several scaling factors including the ground truth high resolution and down-scaling factor $\times 2, \times3, \times 4$. Given the  projection matrices with the  multi-view images we compute the corresponding texture maps.

\begin{figure*}[hbt]
\begin{center}
  \includegraphics[width=0.95\linewidth]{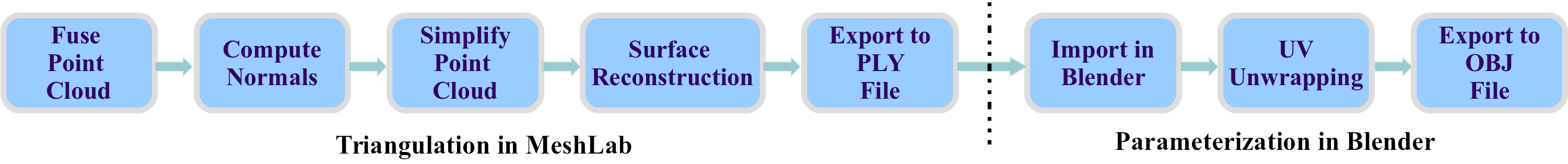}
\end{center}
\vspace{-0.4cm}
\caption{Conversion from point clouds to mesh models and unified parameterization.}
\vspace{-0.4cm}
\label{fig:triangulation_parameterization}
\end{figure*}

\section{The Dataset: 3DASR}
\label{sec:dataset}

The 3DASR dataset we provide is based on four existing subsets; one synthetic subset SyB3R~\cite{ley2016syb3r} and three real subsets EHT3D~\cite{schops2017multi}, MiddleBury~\cite{seitz2006comparison}, and Collection of \emph{Bird}, \emph{Beethoven} and \emph{Bunny} from the multi-view dataset of TUM \cite{goldlucke2014super}, \emph{Fountain}  \cite{Zhou-Koltun-TOG-2014} and \emph{Relief} \cite{Zollhofer-et-al-TOG-2015}. We follow a generic pipeline to preprocess all  subsets. We compute the triangulated 3D mesh with texture coordinates and vertex normals. We use the images provided by the original dataset  as the HR images and we downscale them using scale factors  $\times 2, \times3, \times 4$ to compute the corresponding LR images. The projection matrices for the corresponding LR images are derived by RQ matrix decomposition of the original projection matrix and then scaling down the intrinsic parameters. 

\subsection{The real subsets}
\label{subsec:real_subsets}

ETH3D, Collection, and MiddleBury correspond to real scenes. Regarding ETH3D, we use the training 
set of the HR multi-view subset that contains 13 scenes.
Every scene is provided with multi-view images captured by DSLR cameras, the camera intrinsic and extrinsic parameters, and the ground truth point clouds captured by laser scanners. Collection is a collection of 6 3D scenes. We use the \emph{TempleRing} and \emph{DinoRing} of MiddleBury.

\vspace{-0.4cm}
\paragraph{Mesh: triangulation and UV mapping.} We first compute the triangulated mesh and then unwrap it to define the texture map. Through the UV unwrapping we assign to each vertex a UV coordinate.

For MiddleBury, we use the Multi-View Stereo (MVS) pipeline \cite{schoenberger2016mvs} to reconstruct the meshes. For \emph{Bird}, \emph{Beethoven} and \emph{Bunny} we use the same meshes as in the paper \cite{tsiminaki2014high} and for \emph{Fountain} \emph{Relief} the meshes are refined in the work of Maier~\etal~\cite{Maier-et-al-ICCV-2017}. 

{To ensure low appearance distortion, we use conformal parameterization similar to ~\cite{goldlucke2009superresolution,tsiminaki2014high}. We compute a conformal atlas by selecting the algorithm of LSCM~\cite{levy2002least} that is implemented in Blender.}

For ETH3D subset, the provided 3D model is just a point cloud. Therefore, both of the processing steps are needed. Fig.~\ref{fig:triangulation_parameterization} shows the workflow. Note that triangulation is implemented in MeshLab while parameterization is done in Blender. First of all, for most of the scenes, there are multiple point clouds and each of them captures the scene geometry from different viewpoints. Thus, these point clouds are fused to create a fully-fledged scene geometry followed by the computation of normals. The merged point cloud contains tens of millions of points which may become a computation bottleneck for the post-processing. Thus, the point cloud is simplified using Poisson disk sampling~\cite{corsini2012efficient} which reduces the number of points while maintains the geometric details of the scene. Then the mesh is reconstructed using ball pivoting algorithm~\cite{bernardini1999ball}.

The reconstruction result is exported to a PLY file which is imported into Blender. Blender's UV unwrapping procedure is used for UV parameterization. At last, the triangulated mesh with UV texture coordinates is exported to an OBJ file.
\vspace{-0.4cm}
\paragraph{Images and projection matrices.}We consider the provided images by the original dataset as the HR images and we derive 
the LR images by down-sampling the HR. The intrinsic and extrinsic parameters are given for ETH3D and MiddleBury. Thus computing the projection matrices is straightforward. For the Collection subset, we use RQ decomposition to compute intrinsic and extrinsic parameters. For all of the three subsets, the projection matrices corresponding to the LR images are derived by down-scaling the intrinsic parameters with $\times 2$, $\times 3$, and $\times 4$ scaling factors.

\subsection{The synthetic subset: SyB3R}
\label{subsec:syb3r}

\begin{figure}[t]
\begin{center}
  \includegraphics[width=0.9\linewidth]{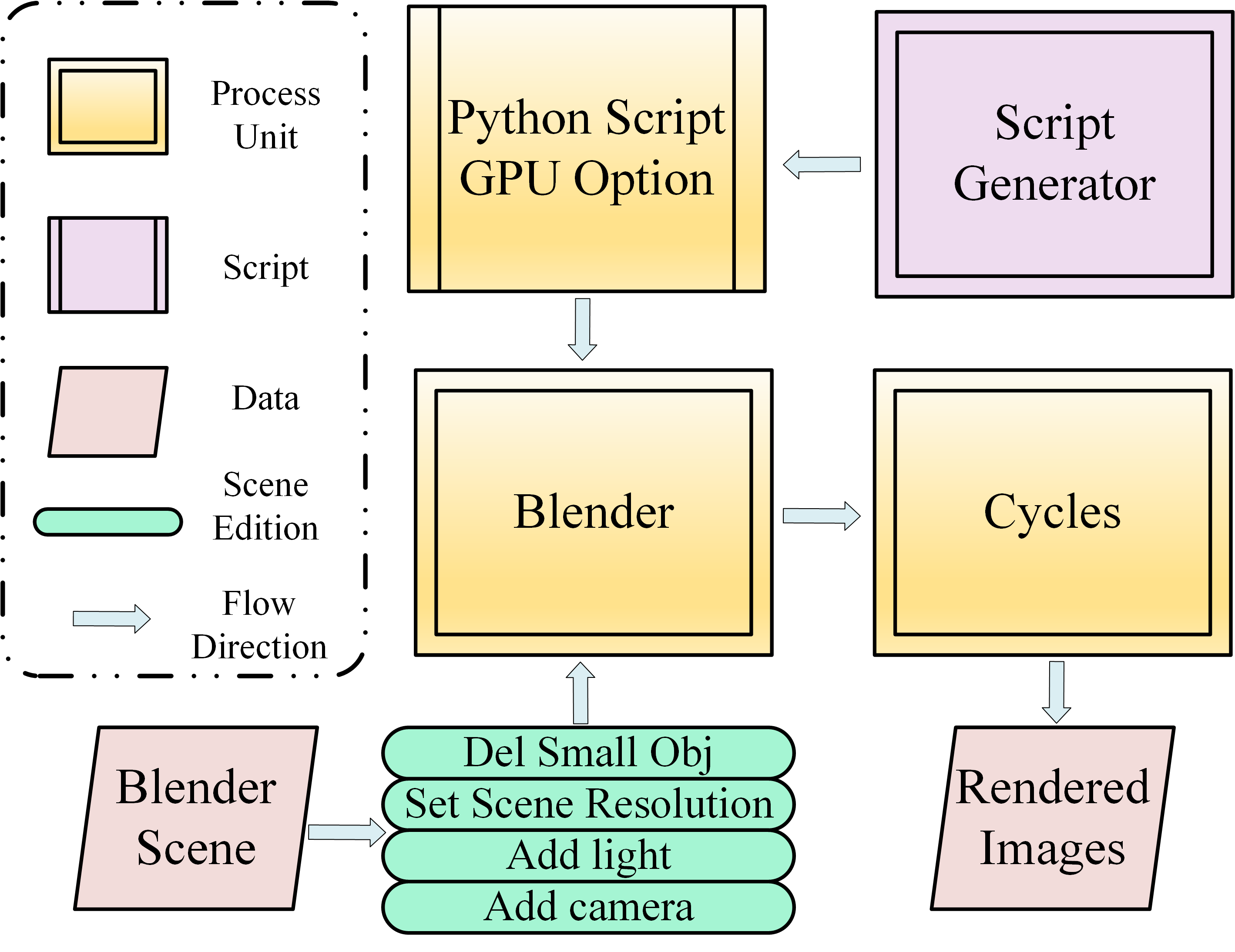}
\end{center}
\vspace{-0.4cm}
\caption{SyB3R image rendering pipeline.}
\vspace{-0.4cm}
\label{fig:image_rendering_pipeline}
\end{figure}

SyB3R is a synthetic dataset containing four scenes. Each scene contains an accurate geometry mesh model with optimal UV parameterization. The image rendering pipeline is shown in Fig.~\ref{fig:image_rendering_pipeline}. To speed up the rendering, we add GPU option to the Python script. We edit the synthetic scene by keeping the major object, setting image resolutions, adding lights and cameras. The generated script and altered scene are passed to Blender and Cycles, resulting in the rendered images. The original mesh model of SyB3R contains several separated objects whose texture maps may overlap with each other in the texture space. To address this problem, we only keep the major part of the scene,~\ie, the body of \emph{Toad}, the skull of \emph{Skull}, and the single rock of \emph{Geological Sample}. We do not use \emph{Lego Bulldozer} because it consists of many small pieces without meaningful texture.

\begin{figure}[t]
    \centering
    \begin{minipage}[t]{0.31\linewidth}
    \centering
    \includegraphics[width=1\textwidth]{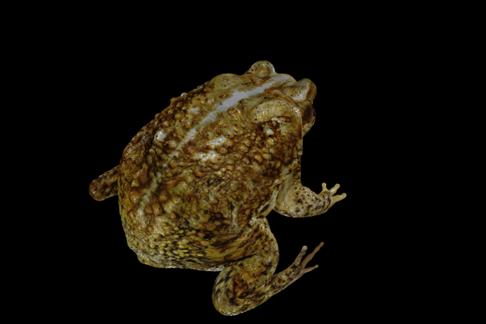}
    \end{minipage}
    ~
    \begin{minipage}[t]{0.31\linewidth}
    \centering
    \includegraphics[width=1\textwidth]{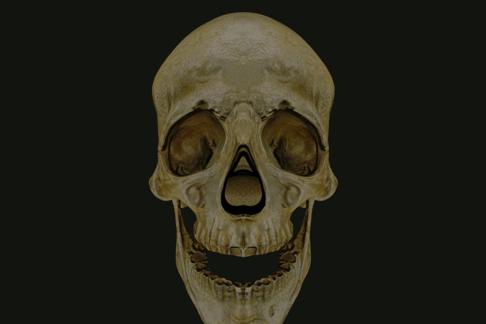}
    \end{minipage}
    ~
    \begin{minipage}[t]{0.31\linewidth}
    \centering
    \includegraphics[width=1\textwidth]{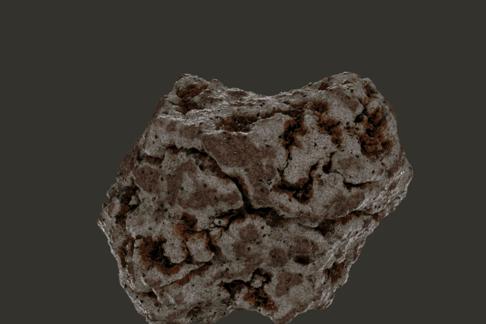}
    \end{minipage}
\vspace{-0.2cm}
\caption{Rendered images of SyB3R.}
\vspace{-0.4cm}
\label{fig:rendered_image}
\end{figure}
\vspace{-0.4cm}

\paragraph{Camera and light.} 
To capture every surface of the object, 14 cameras are uniformly aligned on the sphere surrounding the object. The focal length of the cameras is 25 mm. {The size of the sensor is $32 \times 18$ mm.} To ensure uniform background across the rendered images, 6 lights are added in the  scene lighting from the 6 directions of the object. 
\vspace{-0.4cm}
\paragraph{Rendering images} The resolution of HR images is $3888 \times 2592$ while the resolution of the LR images is calculated by dividing the HR width and height with respective scaling factors.
Knowing the focal length, image resolution, principal point, rotation matrix and translation vector, the $3 \times 4$ camera projection matrix is computed. As stated by the authors~\cite{ley2016syb3r}, the rendering time can be multiple hours per image due to the high computational load of the image synthesis process. Thus, we use GPU to render the images. Examples of rendered images are shown in Fig.~\ref{fig:rendered_image}.

\begin{figure*}
\setlength{\tabcolsep}{1pt}
\renewcommand{\arraystretch}{0.3} 
\resizebox{\linewidth}{!}
{
\begin{tabular}{cccccccc}
\includegraphics[width=.2\linewidth]{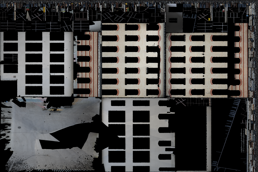}&
\includegraphics[width=.2\linewidth]{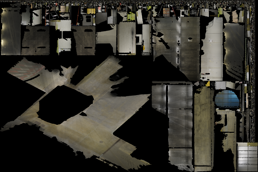}&
\includegraphics[width=.2\linewidth]{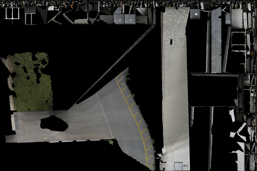}&
\includegraphics[width=.2\linewidth]{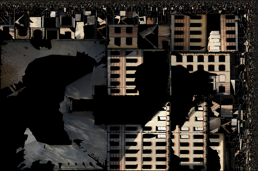}&
\includegraphics[width=.2\linewidth]{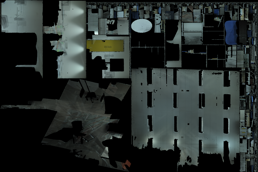}&
\includegraphics[width=.2\linewidth]{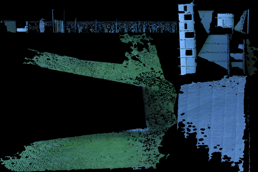}&
\includegraphics[width=.2\linewidth]{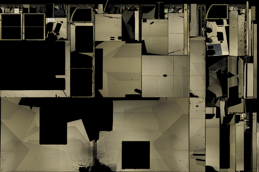}&
\includegraphics[width=.2\linewidth]{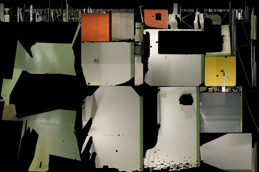}\\
\includegraphics[width=.2\linewidth]{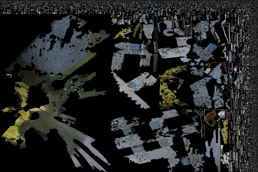}&
\includegraphics[width=.2\linewidth]{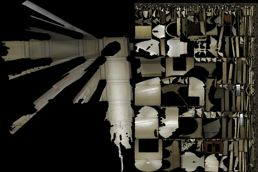}&
\includegraphics[width=.2\linewidth]{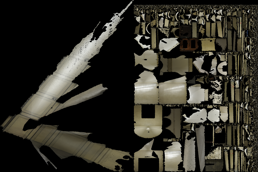}&
\includegraphics[width=.2\linewidth]{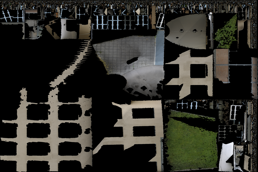}&
\includegraphics[width=.2\linewidth]{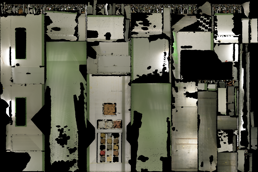}&
\includegraphics[width=.2\linewidth]{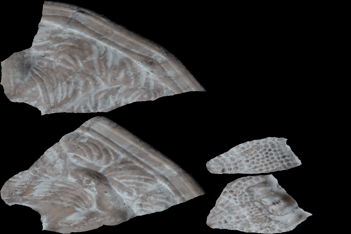}&
\includegraphics[width=.2\linewidth]{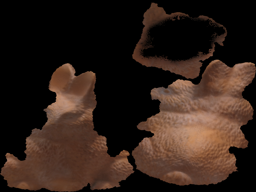}&
\includegraphics[width=.2\linewidth]{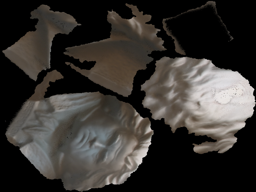} \\
\includegraphics[width=.2\linewidth]{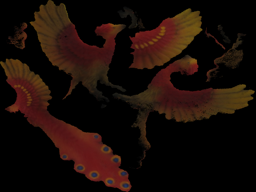}&
\includegraphics[width=.2\linewidth]{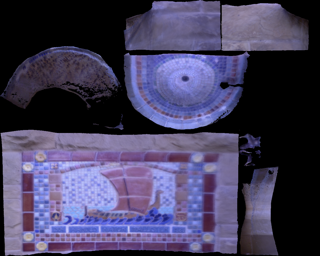}&
\includegraphics[width=.2\linewidth]{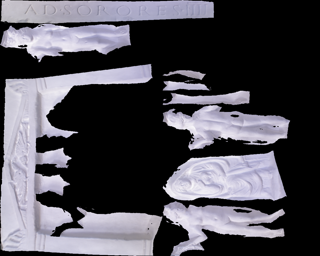}&
\includegraphics[width=.2\linewidth]{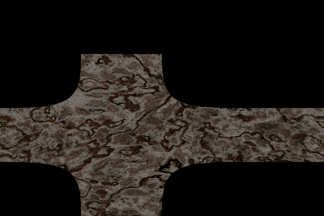}&
\includegraphics[width=.2\linewidth]{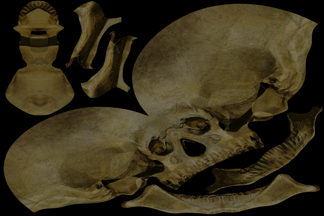}&
\includegraphics[width=.2\linewidth]{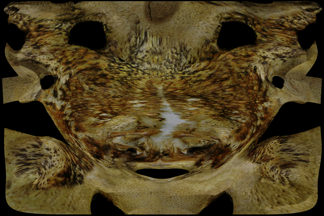}&
\includegraphics[width=.2\linewidth]{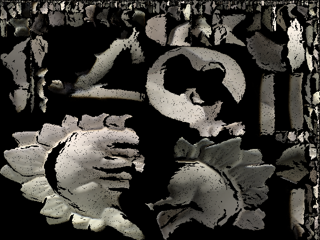}&
\includegraphics[width=.2\linewidth]{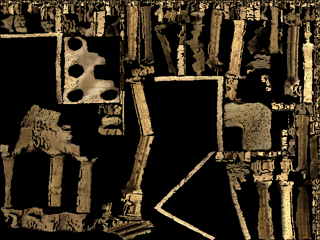}\\
\end{tabular}
}
\vspace{-0.4cm}
\caption{The 24 texture maps from our dataset. From left to right row-wise there are 13 textures derived from ETH3D, 6 from Collection, 3 from SyB3R, and 2 from MiddleBury.}
\vspace{-0.4cm}
\label{fig:texture_map}
\end{figure*}

\subsection{Texture maps}
\label{subsec:texture_maps}
After generating these data, we can now use the texture retrieval algorithm and compute the texture maps of $4$ different resolutions. Fig.~\ref{fig:texture_map} shows the texture maps of the $24$ different scenes.

\section{Learning-Based Methods}
\label{sec:learning_methods}

Our 3DASR dataset contains pairs of HR and LR texture maps which resemble two dimensional images. This allows us to make use of state-of-the-art 2D deep learning-based image SR methods. Such an integration is however not without its own source of difficulties. Being in the multi-view setting, the geometric information needs also to be encoded.
The texture domain has its own characteristics compare to natural images. It is thus important to adapt the 2D SR deep learning-based method to this new domain. We incorporate the 3D geometric information through the normals and we show how to guide the learning process.

\subsection{Normal information}
\label{subsec:normal}

\begin{figure*}
    \centering
    \subfloat[NLR\label{fig:nlr}]{\includegraphics[width=0.48\textwidth]{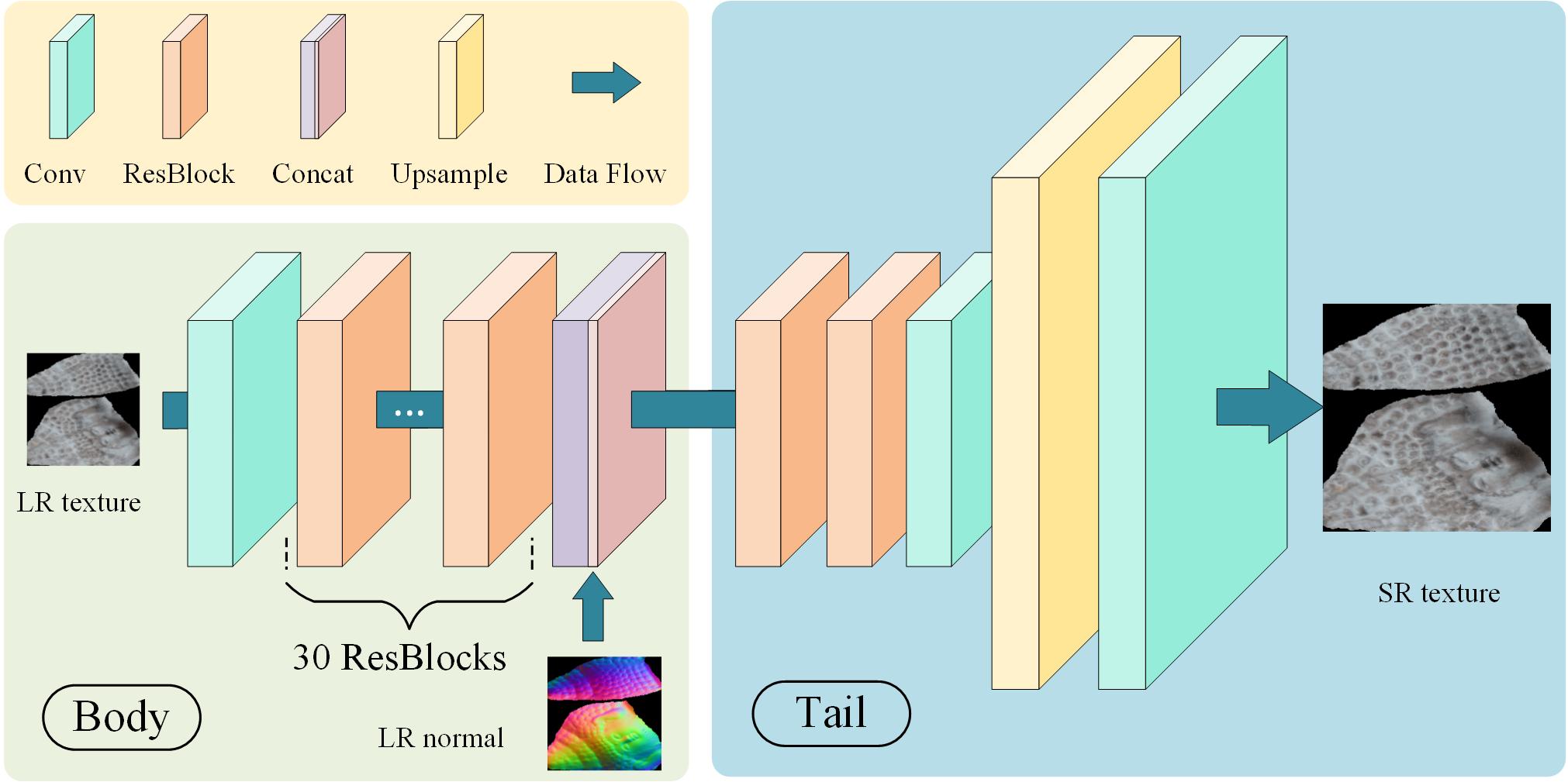}}
    ~
    \subfloat[NHR\label{fig:nhr}]{\includegraphics[width=0.48\textwidth]{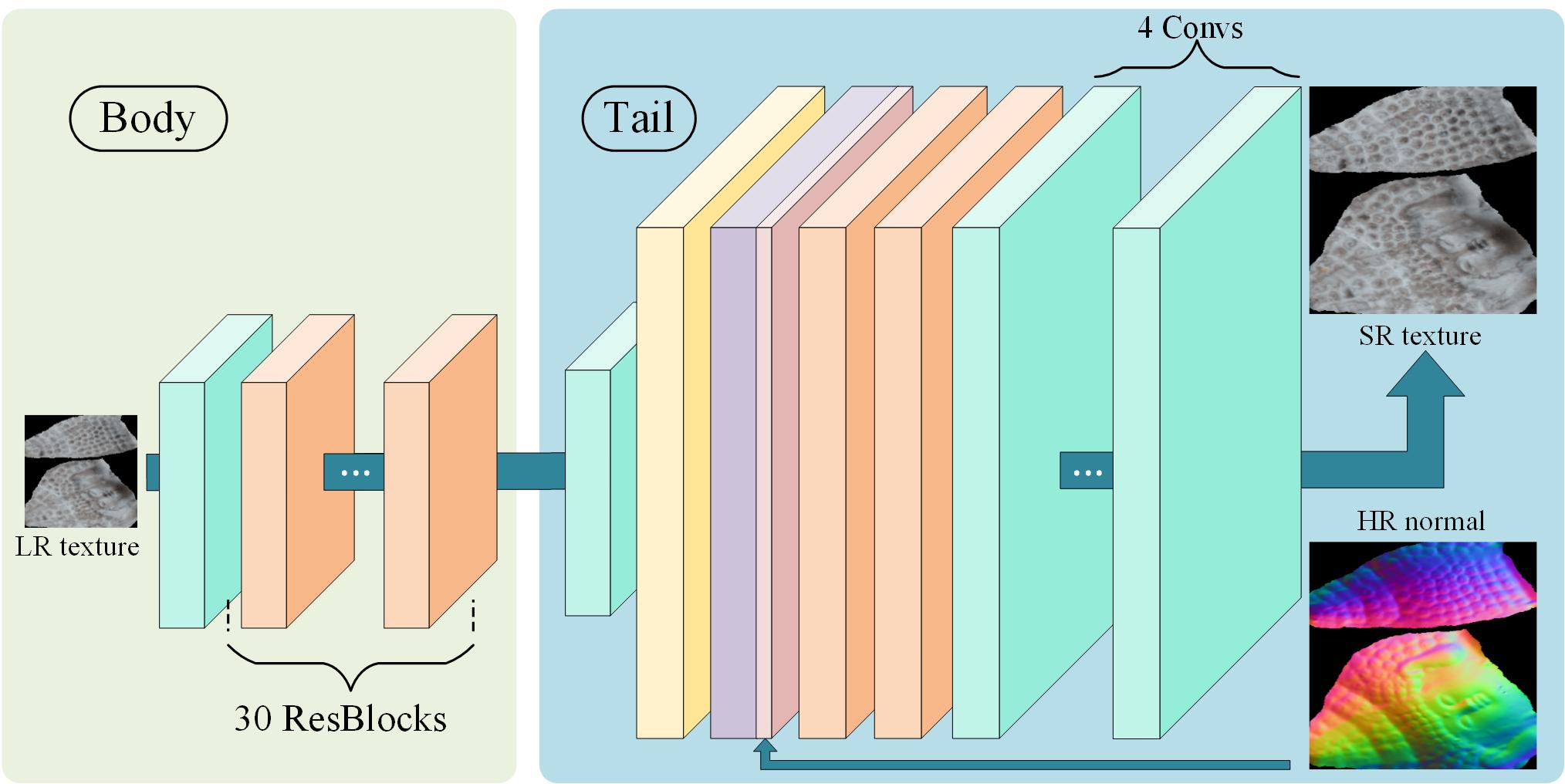}}
    \vspace{-0.2cm}
    \caption{Network structure of (a) NLR and (b) NHR based on the the EDSR~\cite{lim2017enhanced}. The change in the dimension of the blocks indicates the resolution change of the feature maps. In (a) normal maps are computed in the input low resolution space and are concatenated with the feature map before the upscaling layer. In (b) normal maps are computed in the high resolution space and concatenated after the upscaling layer.}
    \vspace{-0.4cm}
    \label{fig:network_architecture}
\end{figure*}

\begin{figure}
    \centering
    \subfloat[\emph{Relief}\label{fig:normal_relief}]{
    \begin{tabular}{*1c}
    \includegraphics[width=0.2\textwidth]{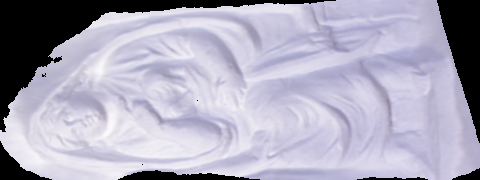} \\
    \includegraphics[width=0.2\textwidth]{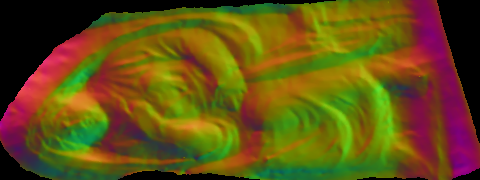}
    \end{tabular}
    }
    ~
    \subfloat[\emph{courtyard}\label{fig:courtyard}]{
    \begin{tabular}{*1c}
    \includegraphics[width=0.2\textwidth]{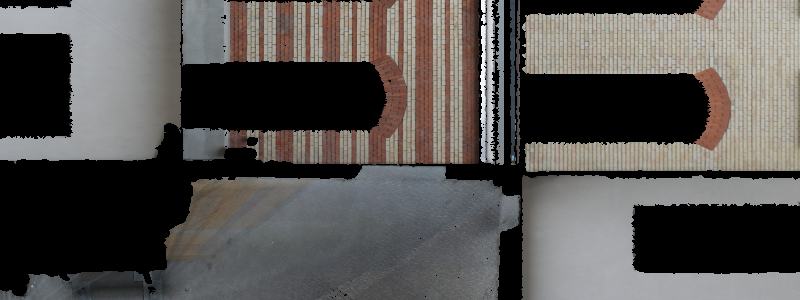} \\
    \includegraphics[width=0.2\textwidth]{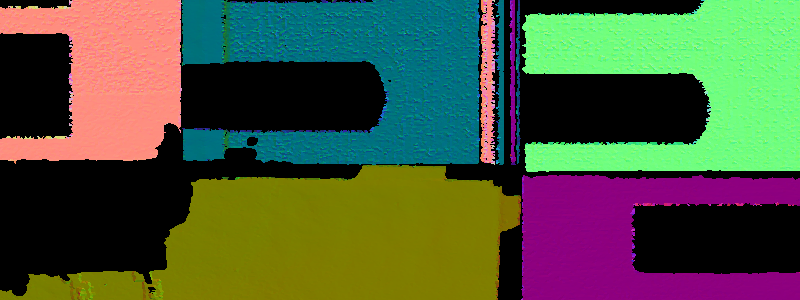}
    \end{tabular}
    }
    \vspace{-0.2cm}
    \caption{Normal maps capture the local structure of the surface.}
    \vspace{-0.4cm}
    \label{fig:normal_visualization}
\end{figure}

Normal coordinates can be normalized and stored as pixel colors in normal maps (Fig.~\ref{fig:normal_visualization}) which have the same support as the texture maps. These normal maps capture the local structure of the surface. We thus use them into the network to  introduce the 3D geometric information. We store them as PNG images with 4 channels. The first 3 channels store the normalized normal coordinates and the fourth alpha channel is a mask that shows the support of the texture map, namely, where texel information is available.

\subsection{Network architecture}
\label{subsec:network}

The next essential step is to incorporate the normal maps and adjust the neural network to the multi-view setting. There are two main approaches. The first is to use them directly as input information by concatenating them with the texture maps. The second approach is to interpret them as high-level features and concatenate them with feature maps computed at specific layers of the network. We follow the second approach due to the following two considerations. First, the normal maps encode 3D geometric information and can indeed be seen as high-level feature maps. Second, in the case where the normal maps were used as input, the whole network should be trained from scratch. Given the small size of our 3DASR dataset this would lead to over-fitting. Thus, by introducing them at higher layers we train only the few last layers of the network, fine-tune the lower ones and avoid this way over-fitting.

In order to examine the importance of the geometric information in the performance of the training, we compute the normals in both spaces of the low and high resolution texture maps. We call them LR
and HR normal maps accordingly. We use EDSR~\cite{lim2017enhanced} as a case study network to show the adaption of the network. We thus provide two difference versions, one where the
LR normal maps are added before the upsampling layer and a second where the HR normal maps are added after the upsampling layer.

\begin{table*}[ht]
  \vspace{-0.4cm}
  \caption{The PSNR results of different methods for scaling factor $\times 2$, $\times 3$, and $\times 4$.}
  \vspace{-0.4cm}
  \label{tbl:benchmark}
  \centering
  \footnotesize
  \begin{tabular}{c||c|c|c||c|c|c||c|c|c||c|c|c||c|c|c}
    \hline  
    \multirow{2}{*}{Method} & \multicolumn{3}{c||}{ETH3D} & \multicolumn{3}{c||}{Collection} & \multicolumn{3}{c||}{MiddleBury} & \multicolumn{3}{c||}{SyB3R} & \multicolumn{3}{c}{Average} \\ \cline{2-16}
    & $\times 2$ & $\times 3$ & $\times 4$ & $\times 2$ & $\times 3$ & $\times 4$ & $\times 2$ & $\times 3$ & $\times 4$ & $\times 2$ & $\times 3$ & $\times 4$ & $\times 2$ & $\times 3$ & $\times 4$  \\ \hline
    Nearest 	& 19.06 	& 16.71 	& 14.68 	& 24.22 	& 19.7 	& 16.92 	& 10.08 	& 7.93 	& 7.08 	& 30.84 	& 27.88 	& 25.82 	& 21.07 	& 18.12 	& 16.0 \\ \hline
    Bilinear 	& 20.61 	& 18.24 	& 16.32 	& 26.2 	& 21.48 	& 18.84 	& 11.87 	& 8.88 	& 7.77 	& 31.75 	& 28.83 	& 26.9 	& 22.67 	& 19.6 	& 17.56 \\ \hline
    Bicubic 	& 20.21 	& 17.96 	& 15.88 	& 25.67 	& 21.12 	& 18.29 	& 11.32 	& 8.81 	& 7.73 	& 31.77 	& 28.78 	& 26.73 	& 22.28 	& 19.34 	& 17.16 \\ \hline
    Lanczos 	& 20.01 	& 17.74 	& 15.69 	& 25.42 	& 20.86 	& 18.07 	& 11.14 	& 8.81 	& 7.81 	& 31.71 	& 28.7 	& 26.63 	& 22.09 	& 19.15 	& 17.0 \\ \hline
    HRST 	& 16.18 	& -- 	& 16.12 	& 32.29 	& -- 	& 29.63 	& 22.13 	& -- 	& 20.88 	& 27.9 	& -- 	& 26.34 	& 22.17 	& -- 	& 21.17 \\ \hline
    HRST-CNN 	& -- 	& -- 	& -- 	& 32.24 	& -- 	& 29.9 	& 22.76 	& -- 	& 21.55 	& -- 	& -- 	& -- 	& -- 	& -- 	& -- \\ \hline
    EDSR 	& 16.75 	& 14.08 	& 12.03 	& 21.77 	& 17.2 	& 14.24 	& 8.49 	& 7.13 	& 6.61 	& 29.31 	& 26.18 	& 23.81 	& 18.89 	& 15.79 	& 13.61 \\ \hline
    EDSR-FT 	& 21.13 	& 19.75 	& 18.44 	& 28.25 	& 25.53 	& 24.19 	& 12.73 	& 11.21 	& 9.9 	& 32.78 	& 29.9 	& 28.31 	& 23.66 	& 21.75 	& 20.4 \\ \hline
    NLR-Sub 	& 21.21 	& 20.11 	& 19.2 	& 28.08 	& 25.0 	& 23.27 	& 14.68 	& 12.37 	& 11.11 	& 32.18 	& 28.84 	& 26.64 	& 23.75 	& 21.78 	& 20.47 \\ \hline
    NLR 	& 21.31 	& 20.27 	& 19.18 	& 28.38 	& 25.85 	& 24.84 	& 13.67 	& 12.92 	& 12.29 	& 32.57 	& 29.57 	& 27.67 	& 23.85 	& 22.22 	& 21.08 \\ \hline
    NHR 	& 25.19 	& 23.95 	& 22.7 	& 30.25 	& 28.41 	& 26.27 	& 17.16 	& 17.21 	& 15.63 	& 30.57 	& 27.42 	& 24.39 	& 26.46 	& 24.94 	& 23.22 \\ \hline
  \end{tabular}
  \vspace{-0.1cm}
\end{table*}

\subsection{Implementation details}
\label{subsec:implementation}

The architecture of the two adapted networks is shown in Fig.~\ref{fig:nlr} and Fig.~\ref{fig:nhr}, which we name as NLR and NHR, representing the utilization of LR and HR normal maps. In Fig.~\ref{fig:nlr}, LR normal maps are concatenated with the feature maps after the 30th ResBlock. The following two ResBlocks and the upsampling layer learn representation from the combined feature map. In Fig.~\ref{fig:nhr}, upsampling layer is moved before the two fine-tuning ResBlocks and the HR normal maps are added directly after the upsampling layer. Four additional convolutional layers follow the two ResBlocks. The number of feature maps after the concatenation becomes 260 which is the sum of the original 256 channels and the additional 4 channels of the normal map. 

We name the layers from the starting convolutional layer to the 30th ResBlock as the body part of the network. The remaining layers are referred to as the tail part. The parameters of the body part are loaded from pretrained EDSR model and fine-tuned to adapt to the texture domain while those of the tail part are randomly initialized and trained from scratch. Thus, a larger learning rate ${10} ^ {-4}$ is used to train the tail parameters while a smaller one ${10} ^ {-5}$ is used to fine-tune the body parameters. We also directly fine-tune the EDSR model without any architecture modification. An in-between learning rate ${2.5} \times {10} ^ {-5}$ is used. To train the CNN, the mask is used to identify the active areas of the texture maps. We crop the texture maps into patches of size $48 \times 48$ and feed them into the network for training by excluding these ones that have black areas larger than a predefined threshold $50$. During inference the CNN is applied on the whole LR texture map.

The provided dataset contains 4 subsets and 24 texture maps in total. Cross-validation is used to get the evaluation result on the whole dataset. That is, we divide the 24 texture map into 2 splits, one for training and one for testing. The texture maps of the 4 subsets are equally distributed to the two splits, thus each with 12 texture maps. In addition, we also try cross-validation within the subset. That is, the training and testing texture maps are from the same subset. The 4 subsets are captured under different conditions and they may have different characteristics. In the case of 
cross validation within the subset, the training and testing data are from the same subset and they have the same characteristics. In the case of cross-validation on the whole dataset, there are more training data but with different characteristics. A  comparison of these two cases can indicate whether subset characteristics or large training set is more important in our problem setting. The networks are trained for 50 epochs for subset cross-validation and 100 epochs for all of the other experiments.

\begin{figure*}
\setlength{\tabcolsep}{1pt}
\renewcommand{\arraystretch}{0.3}
\resizebox{\linewidth}{!}
{
\begin{tabular}{cccccc}
Groud Truth & Bilinear & EDSR & EDSR-FT & NLR & NHR \\
    \includegraphics[width=0.16\textwidth]{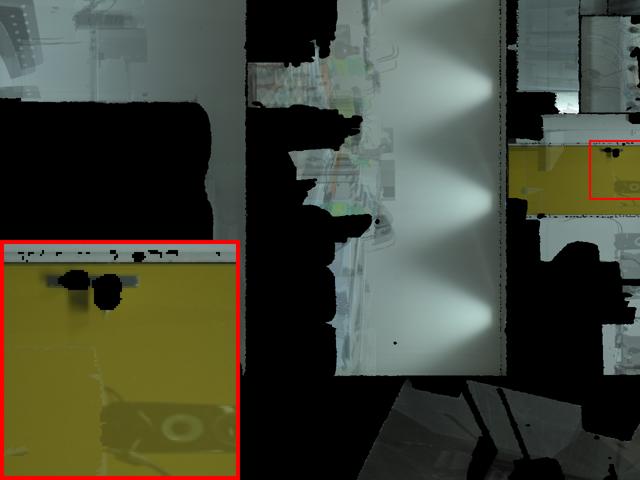} 
    & \includegraphics[width=0.16\textwidth]{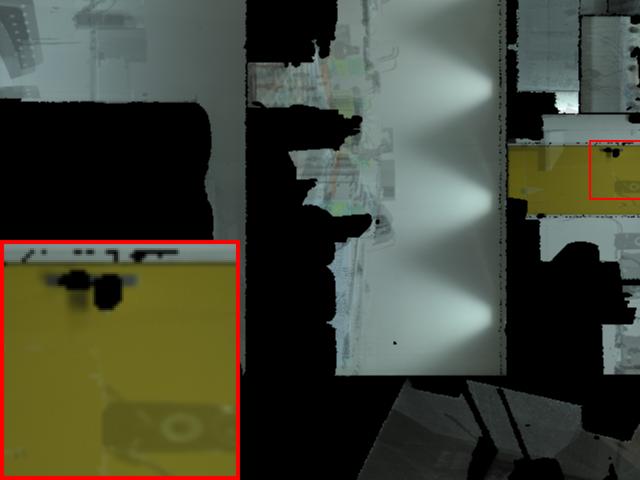}     
    & \includegraphics[width=0.16\textwidth]{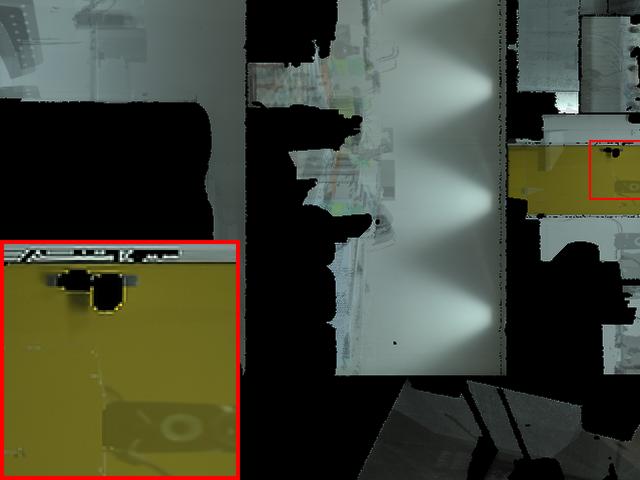} 
    & \includegraphics[width=0.16\textwidth]{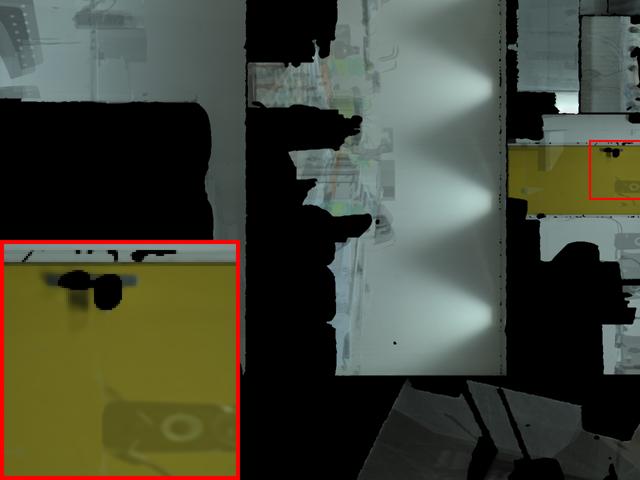} 
    & \includegraphics[width=0.16\textwidth]{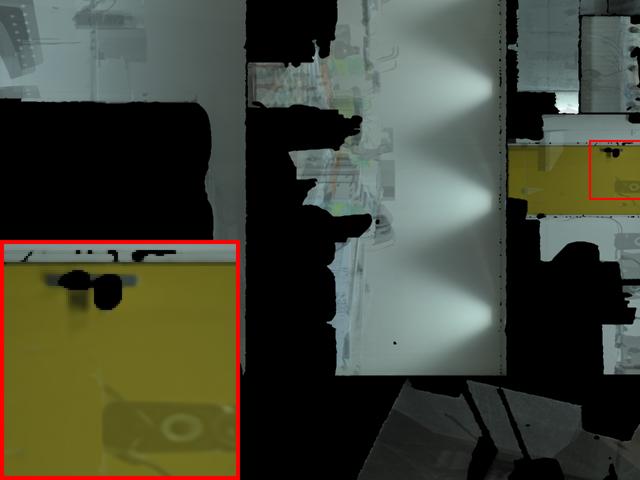} 
    & \includegraphics[width=0.16\textwidth]{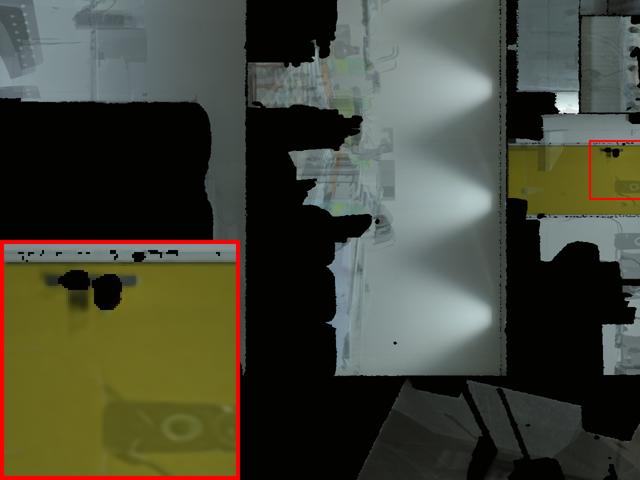} \\
    \includegraphics[width=0.16\textwidth]{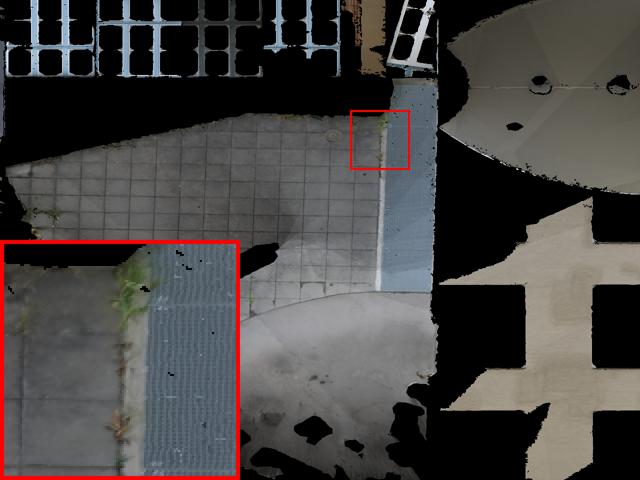} 
    & \includegraphics[width=0.16\textwidth]{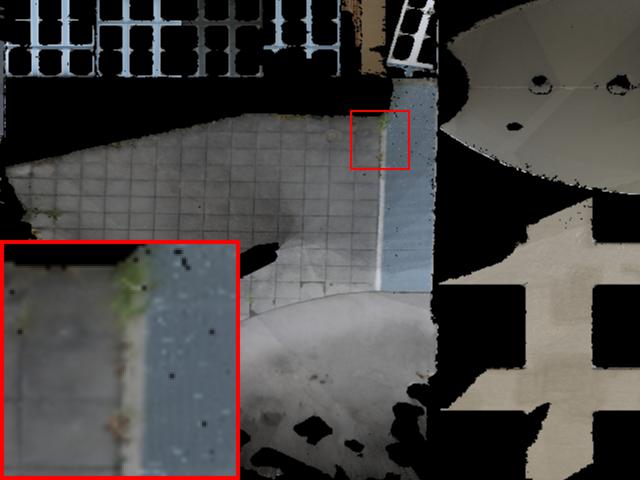}     
    & \includegraphics[width=0.16\textwidth]{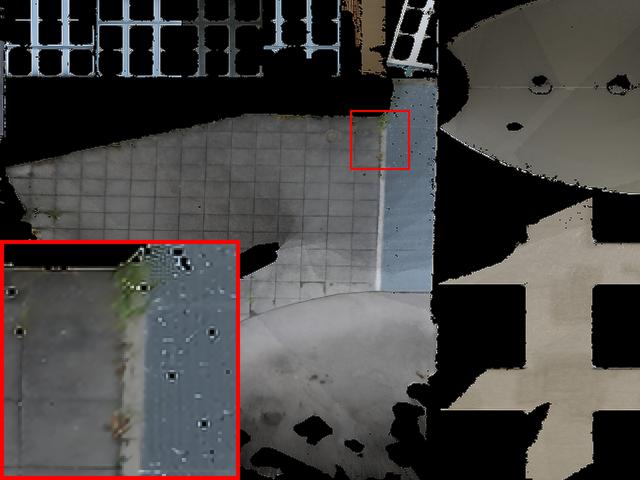} 
    & \includegraphics[width=0.16\textwidth]{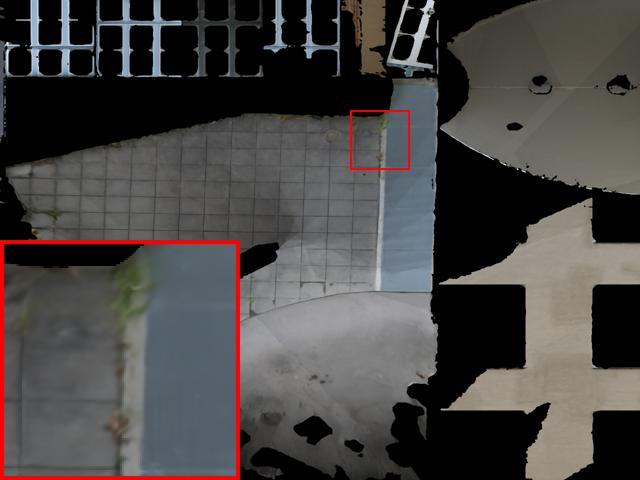} 
    & \includegraphics[width=0.16\textwidth]{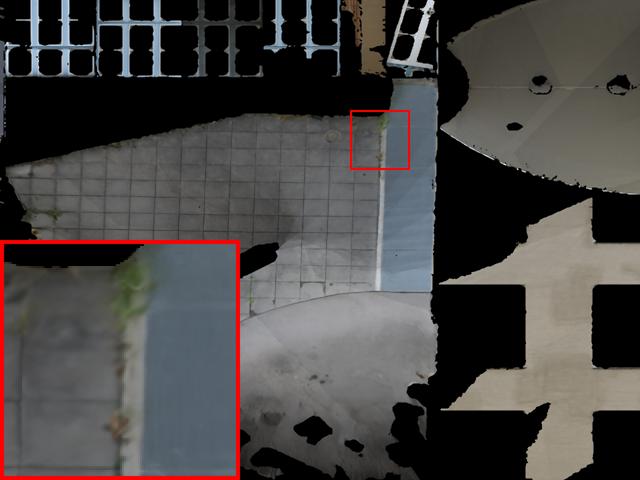} 
    & \includegraphics[width=0.16\textwidth]{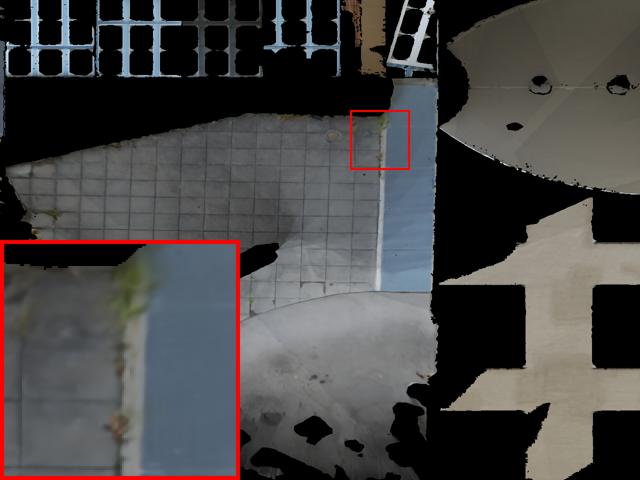} \\
    \includegraphics[width=0.16\textwidth]{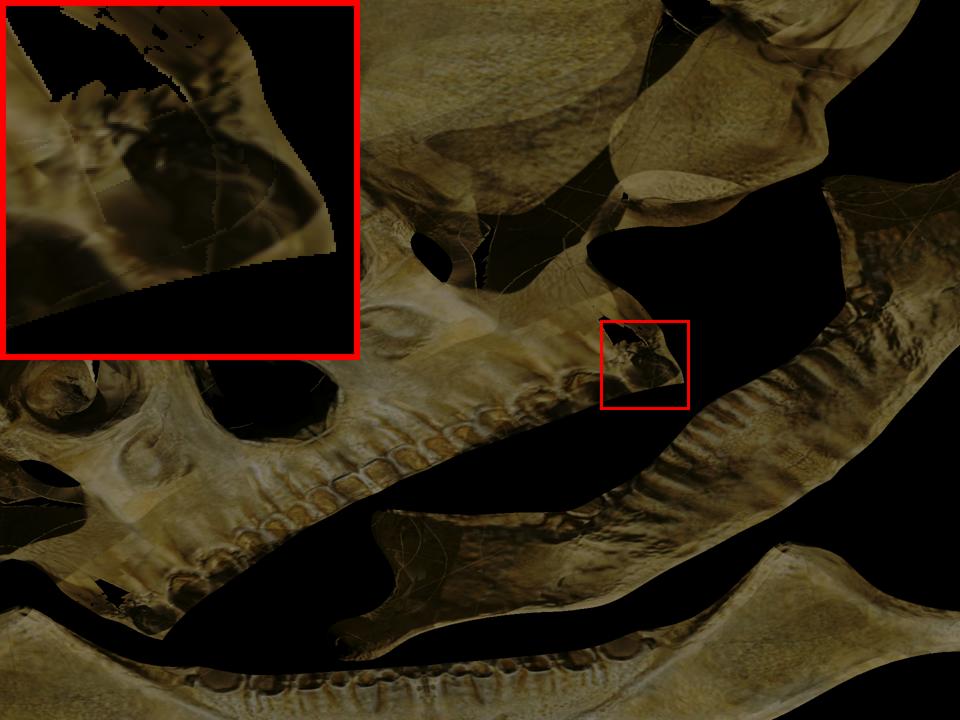} 
    & \includegraphics[width=0.16\textwidth]{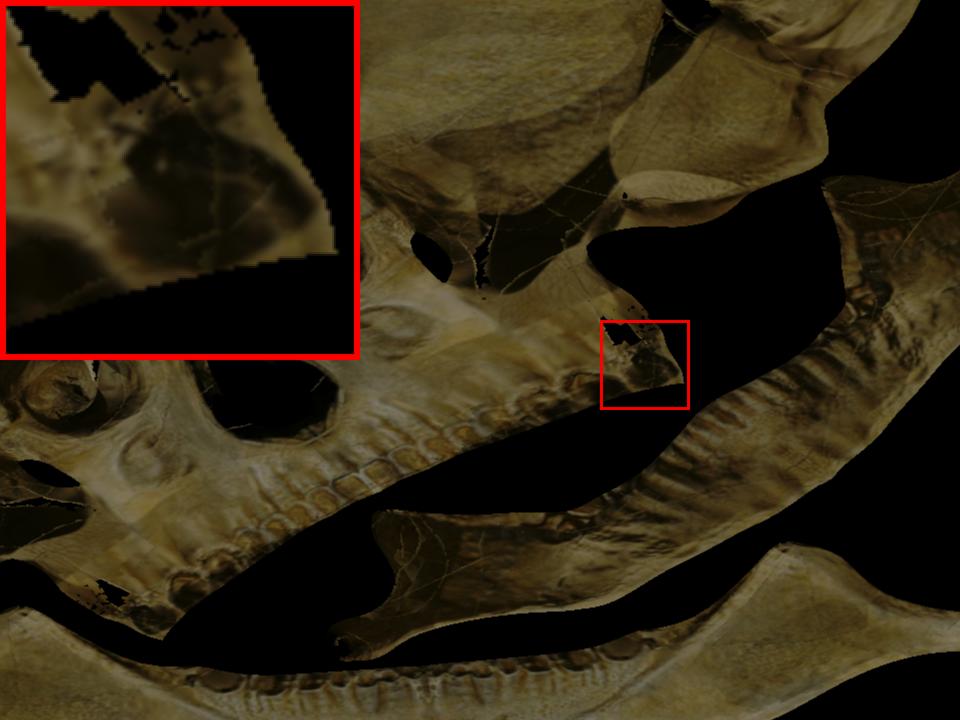}     
    & \includegraphics[width=0.16\textwidth]{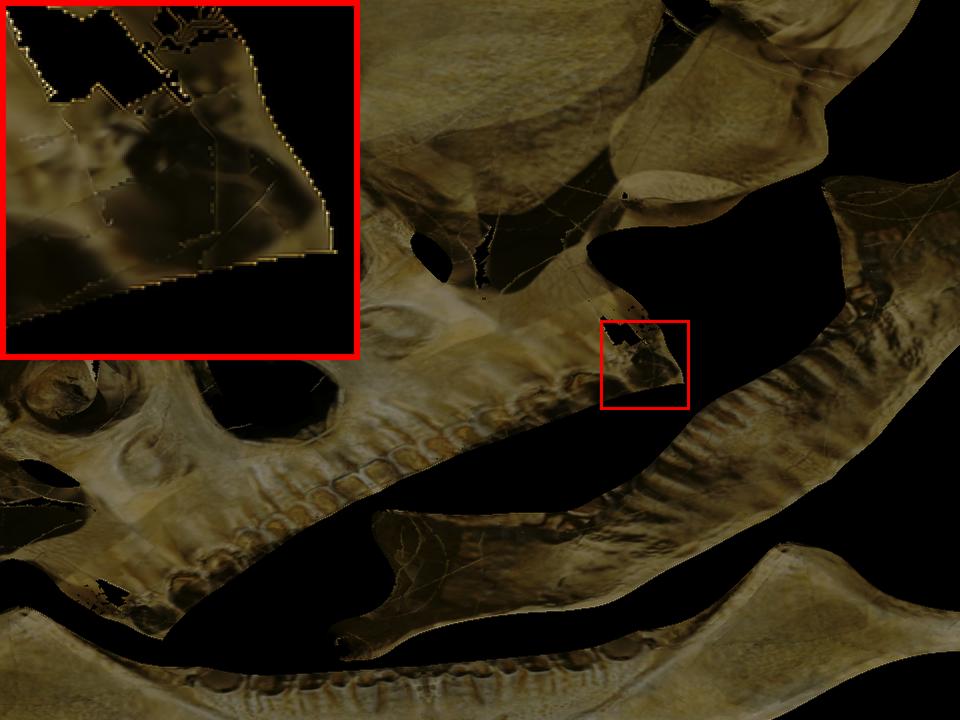} 
    & \includegraphics[width=0.16\textwidth]{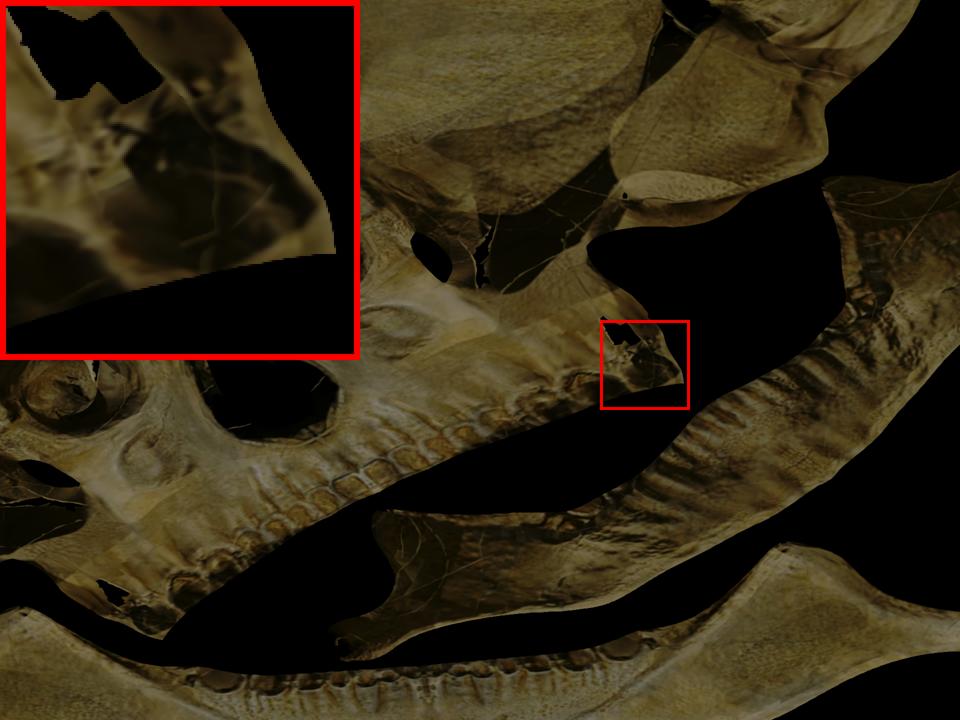} 
    & \includegraphics[width=0.16\textwidth]{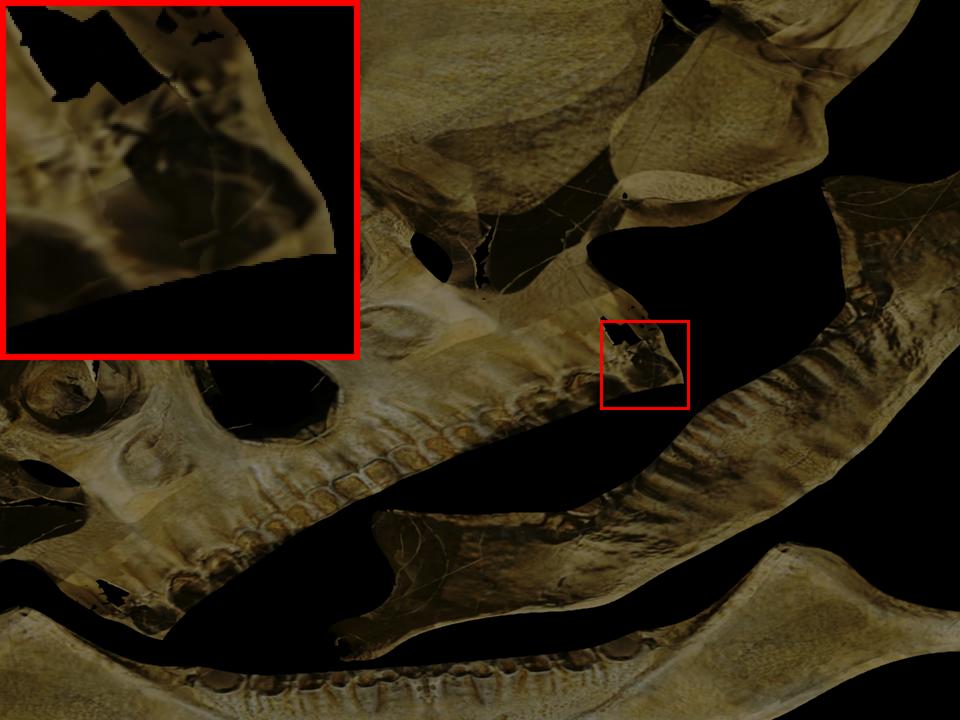} 
    & \includegraphics[width=0.16\textwidth]{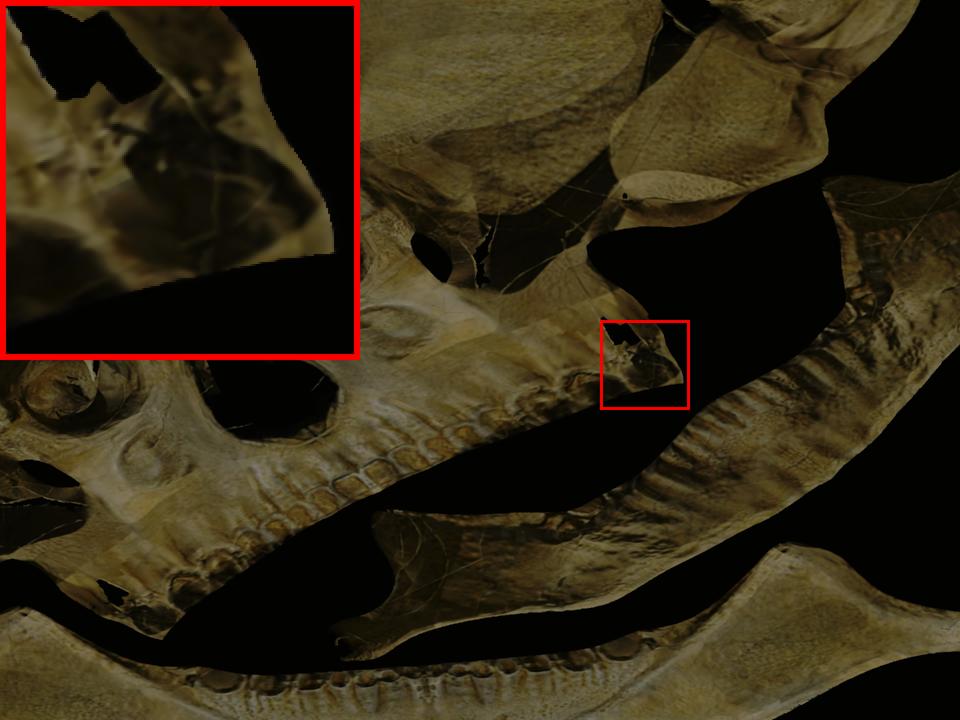} \\
\end{tabular}
}
\vspace{-0.4cm}
\caption{The visual results of \emph{pipes}, \emph{terrace}, and \emph{Skull} for scaling factor $\times 2$.}
\vspace{-0.4cm}
\label{fig:visual_results1}
\end{figure*}

\begin{figure*}
\setlength{\tabcolsep}{1pt}
\renewcommand{\arraystretch}{0.3}
\resizebox{\linewidth}{!}
{
\begin{tabular}{ccccc}
EDSR 21.77dB & EDSR-FT 28.25dB & NLR 28.38 dB & NHR 30.25dB & HRST 32.29dB\\
    \includegraphics[width=0.19\textwidth]{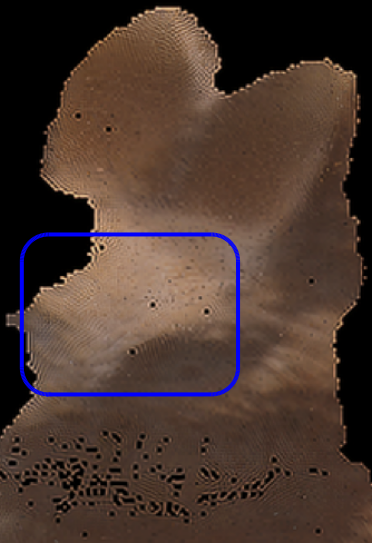}
    & \includegraphics[width=0.19\textwidth]{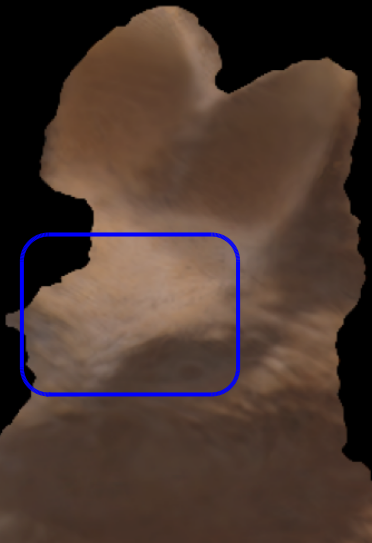}
    & \includegraphics[width=0.19\textwidth]{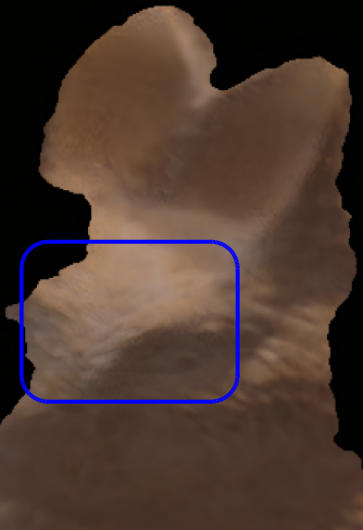}
    & \includegraphics[width=0.19\textwidth]{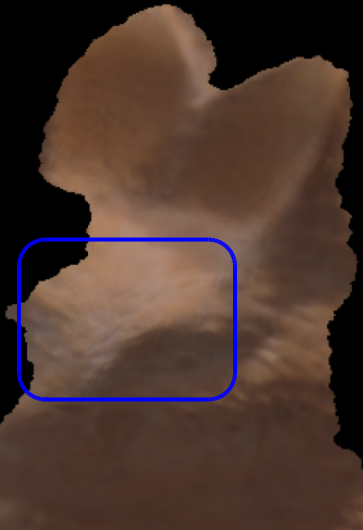}
    & \includegraphics[width=0.19\textwidth]{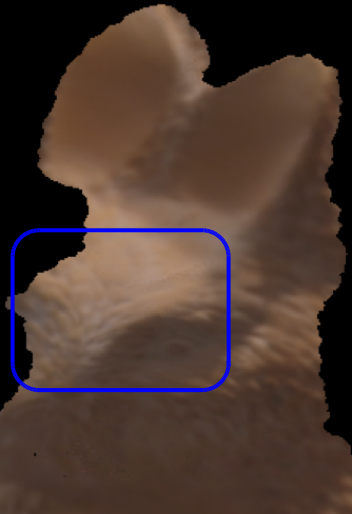}\\
\end{tabular}
}
\vspace{-0.4cm}
\caption{PSNR (dB) and close-ups of the super-resolved texture map of \emph{Bunny} for scaling factor $\times 2$. While adding gradually the characteristics of the domain more details are recovered. The NHR achieves the highest PSNR value among the deep learning-based approaches while it stays still below the model-based HRST. Note that with NHR, texture SR  is a \textbf{feed forward step}, while with HRST is an \textbf{iterative approach}.}
\vspace{-0.4cm}
\label{fig:visual_results2}
\end{figure*}

\section{Results}
\label{sec:results}

Using our 3DASR dataset, we compare three main categories; interpolation-based, model-based and learning-based methods for super-resolving the appearance of 3D objects. The interpolation-based methods include nearest, bilinear, bicubic, and Lanczos~\cite{duchon1979lanczos} interpolation. We use the method of Tsiminaki~\etal \cite{tsiminaki2014high} as a representative of the model-based category, denoted as HRST. Using the EDSR network as a base model, we introduce several modifications of it. There are in total 6 different cases. EDSR: We use the pretrained network EDSR and directly test it on our data. EDSR-FT: We fine-tune the pretrained EDSR on our 3DASR dataset without architecture modification and using whole set cross-validation. 
NLR-Sub: We incorporate LR normal map into EDSR and use subset cross-validation. NLR: We incorporate LR normal map into EDSR as in Fig.~\ref{fig:nlr} and use whole set cross-validation. NHR: We incorporate HR normal map into EDSR as in Fig.~\ref{fig:nhr} and use whole set cross-validation. HRST-CNN: We use EDSR as a post-processing step of the super-resolved texture maps of HRST. In this scenario, the upsampling layer of EDSR is replaced with ordinary convolutional layers.

\subsection{Objective metrics}
\label{subsec:objective_metric}

We compute PSNR metrics in the active regions of the texture domains, that is, on the set of texels in the texture domain that is actually mapped to the 3D model. For the purpose of benchmarking, these metrics can also be computed in the image domain by reprojecting the texture maps into the image space. According to the PSNR values of Table~\ref{tbl:benchmark}, we can draw the following conlcusions.
\vspace{-0.4cm}
\paragraph{Interpolation based methods.} Among the interpolation-based methods, bilinear interpolation achieves better results than bicubic and Lanczos interpolation, which contradicts the 2D image interpolation. This can be probably explained by the fact that the texture and the ordinary image domains have different characteristics.  In the 2D image SR,  LR image is modeled as bicubic down-sampled verison of the HR image, which favors advanced interpoaltion methods. In the multi-view setting, due to the several sources of variability, the LR and HR texture maps might be not strictly aligned. 
\vspace{-0.4cm}
\paragraph{Fine-tuning learning-based methods.} The texture domain knowledge is
different than the image domain. The fine-tuning of EDSR-FT incorporates the characteristics of the texture compare to the pretrained EDSR model. Thus, algorithms need to be adpated to the spesific domain.
\vspace{-0.4cm}
\paragraph{LR vs. HR normal maps.}
We incorporate the 3D geometric information of the multi-view setting through the normal maps
and we  compare to the simple case of fine-tuned EDSR-FT. According to the PSNR values, the geometric information imrpoves the quality of the reconstructed texture maps. We then validate its importance
by comparing the two cases of NLR and NHR. The  PSNR values increase even more when we express this geometric information with higher precision. NHR case, where HR normal maps are used outperforms NLR. Thus, HR normal maps capture more geometric details and  improve the performance. 
\vspace{-0.4cm}
\paragraph{Subset characteristics vs. training data size.} 
NLR-Sub uses cross-validation on the subset  while NLR on the whole set.
In the case of NLR-Sub, the subset characteristics are respected while in the case of NLR not. The main advantage of NLR is that more data are used for training (12 HR texture maps). The high PSNR values of the NLR compared to NLR-Sub indicate that the training data size is more important than subset characteristics to this task. Furthermore, the PSNR gap between NLR and NLR-Sub on ETH3D is larger than that on MiddleBury and Collection. This is because ETH3D is a relatively larger dataset than MiddleBury and Collection. Thus, even if subset cross-validation is used, NLR-Sub does not diverge a lot from NLR on ETH3D dataset. Therefore, we conclude that although each subset may have its own characteristics, training data size stands out as a major factor.
\vspace{-0.4cm}
\paragraph{Model based vs. learning based methods.} 
The model-based method HRST formulates the texture retrieval problem as an optimization problem.  It is  a two-stage iterative algorithm and its computational cost increases even more with an increase of geometric complexity. This explains the unstable behaviour of HRST method across the datasets. HRST outperforms NHR 
on MiddleBury and Collection whereas on ETH3D and SyB3R not. In most of the cases,  HRST-CNN enhances the super-resolved texture maps. It is important to note that even in the cases where the model-based method outperforms the deep learning-based approach, the PSNR values are relatively close. More importantly, the deep learning-based approach is a feed-forward step that can be executed in seconds while the model-based is a heavy iterative process.

\subsection{Visual results}
\label{subsec:visual_results}

The visual results are shown in Fig.~\ref{fig:visual_results1} and Fig.~\ref{fig:visual_results2}. Directly upsampling the LR texture maps creates blurring images. EDSR leads to some white texels along the boundaries between the black region and the texture region. While we introduce gradually the characteristics of the domain through the EDSR-FT, NLR, and NHR methods, we successfully recover more visual details.

\section{Conclusion}
\label{sec:conclusion}
We provided 3DASR, a 3D appearance SR dataset \footnote{The dataset, the evaluation codes, and the baseline models is available at \url{https://github.com/ofsoundof/3D_Appearance_SR}.} that captures both synthetic and real scenes with a large variety of texture characteristics. It is based on four datasets, ETH3D, Collection, MiddleBury, and SyB3R. The dataset contains ground truth HR  texture maps and  LR texture maps of  scaling factors $\times 2$, $\times 3$, and $\times 4$. The 3D mesh, multi-view images, projection matrices, and normal maps are also provided. We introduced
a deep learning-based SR framework in the multi-view setting. We showed that 2D deep learning-based SR techniques can successfully  be adapted to the new texture domain by introducing the geometric information via normal maps and achieve relatively similar performance to the model-based methods. This work opens up a novel direction of deep learning-based texture SR methods for the multi-view setting.  A necessary next step is to
enlarge our dataset either through common augmentation techniques or by following our proposed texture retrieval pipeline to introduce new datasets. The fact that the performance of our  deep learning-based SR framework is in some cases (MiddleBury and Collection) below the model-based one indicates that there is still space for more elaborate methods that unify the concepts of model-based SR techniques and the 2D deep learning-based approaches.

{\small
\bibliographystyle{ieee}
\bibliography{ms}
}
\appendix
\onecolumn
\clearpage









\section{Details of the Provided Dataset}

\begin{table*}[b]
  \caption{Details of the each of the scenes in the provided dataset including the size (MB) of the mesh, number of vertices and faces (k) in the mesh, the resolution of HR images, and the number of views in each scene.}
  \label{tbl:scene_details}
  \centering
  \begin{tabular}{c||c||c|c|c||c|c}
    \hline  
    \multirow{2}{*}{Dataset} & \multirow{2}{*}{Scene} & \multicolumn{3}{c||}{Mesh} & \multicolumn{2}{c}{Image} \\ \cline{3-7} 
    &&Mesh size & No. vertices & No. Faces & Resolution & No. Views\\ \hline
    \multirow{13}{*}{ETH3D} & \emph{courtyard} & 136.0 & 646 & 1,168 & 3096 $\times$ 2064 & 38 \\ \cline{2-7}
    & \emph{delivery$\_$area} & 100.1 & 511 & 911 & 3096 $\times$ 2064 & 44\\ \cline{2-7}
    & \emph{electro}  & 32.7 & 164 & 293  & 3084 $\times$ 2052 & 45\\ \cline{2-7}
    & \emph{facade}  & 121.5 & 583 & 1,026 & 3096 $\times$ 2052 & 46\\ \cline{2-7}
    & \emph{kicker}  & 116.1 & 571 & 1,004 & 3096 $\times$ 2064 & 31\\ \cline{2-7}
    & \emph{meadow}  & 32.0 & 156 & 282 & 3096 $\times$ 2064 & 15\\ \cline{2-7}
    & \emph{office}  & 194.1 & 882 & 1,663 & 3108 $\times$ 2064 & 26\\ \cline{2-7}
    & \emph{pipes}  & 171.3 & 850 & 1,502 & 3108 $\times$ 2064 & 14\\ \cline{2-7}
    & \emph{playground}  & 44.8 & 233 & 367 & 3096 $\times$ 2064 & 38\\ \cline{2-7}
    & \emph{relief}  & 36.7 & 186 & 324 & 3096 $\times$ 2064 & 31\\ \cline{2-7}
    & \emph{relief$\_$2}  & 58.5 & 299 & 519  & 3096 $\times$ 2064 & 31\\ \cline{2-7}
    & \emph{terrace}  & 52.0 & 327 & 421 & 3096 $\times$ 2064 & 23\\ \cline{2-7}
    & \emph{terrains}  & 55.5 & 265 & 495 & 3096 $\times$ 2064 & 42\\ \hline
    \multirow{6}{*}{Collection} & \emph{Beethoven} & 16.0 & 74 & 144 & 1024 $\times$ 768 & 33\\ \cline{2-7}
    & \emph{Bird}  & 16.7 & 78 & 150 & 1024 $\times$ 768 & 20\\ \cline{2-7}
    & \emph{Buddha}  & 16.5 & 75 & 150 & 1404 $\times$ 936 & 91\\ \cline{2-7}
    & \emph{Bunny}  & 12.3 & 57 & 111 & 1024 $\times$ 768 & 36\\ \cline{2-7}
    & \emph{Fountain}  & 40.4 & 200 & 399 & 1280 $\times$ 1024 & 55\\ \cline{2-7}
    & \emph{Relief}  & 47.6 & 233 & 464 & 1280 $\times$ 1024 & 40\\ \hline
    \multirow{2}{*}{Middlebury} & \emph{DinoRing} & 143.2 & 619 & 1,237 & 640 $\times$ 480 & 48\\ \cline{2-7}
    & \emph{TempleRing}  & 85.2 & 369 & 737 & 640 $\times$ 480 & 47\\ \hline
    \multirow{3}{*}{SyB3R} & \emph{GeologicalSample} & 14.7 & 98 & 197  & 3888 $\times$ 2592 & 14\\ \cline{2-7}
    & \emph{Skull}  & 2.8 & 19 & 38 & 3888 $\times$ 2592 & 14\\ \cline{2-7}
    & \emph{Toad}  & 77 & 481 & 962 & 3888 $\times$ 2592 & 14\\ \hline
  \end{tabular}
  \vspace{-0.4cm}
\end{table*}

\subsection{Mesh, images, and projection matrices}

The provided dataset has 24 different scenes in total including 13 from ETH3D, 6 from Collection, 3 from SyB3R, and 2 from Middlebury. Each scene contains a 3D mesh, multi-view images, and the corresponding projection matrices. The details of those scenes are provided in Table~\ref{tbl:scene_details} including the mesh size, the number of vertices and faces in the mesh, the resolution of the HR images, and the number of views in the scene. It is shown in Table~\ref{tbl:scene_details} that the scenes have different complexities, \ie, different mesh size and number of vertices and faces.   

\subsection{Texture maps}

\begin{figure*}
\setlength{\tabcolsep}{1pt}
\renewcommand{\arraystretch}{0.3}
\resizebox{\linewidth}{!}
{
\begin{tabular}{ccccc}
    &Groud Truth & $\times 2$ LR  & $\times 3$ LR & $\times 4$ LR \\
    \emph{courtyard} & \includegraphics[width=0.48\textwidth]{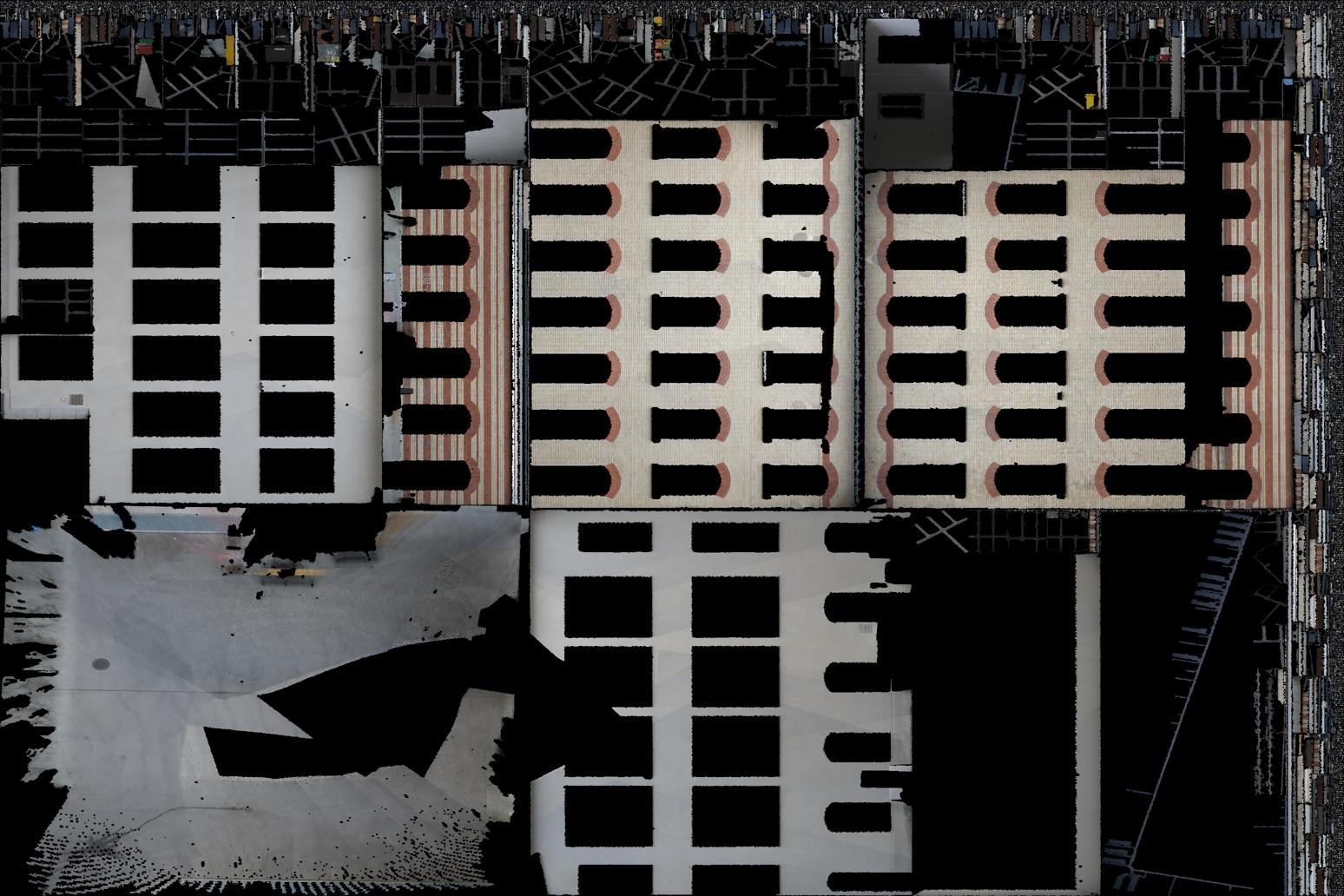} 
    & \includegraphics[width=0.24\textwidth]{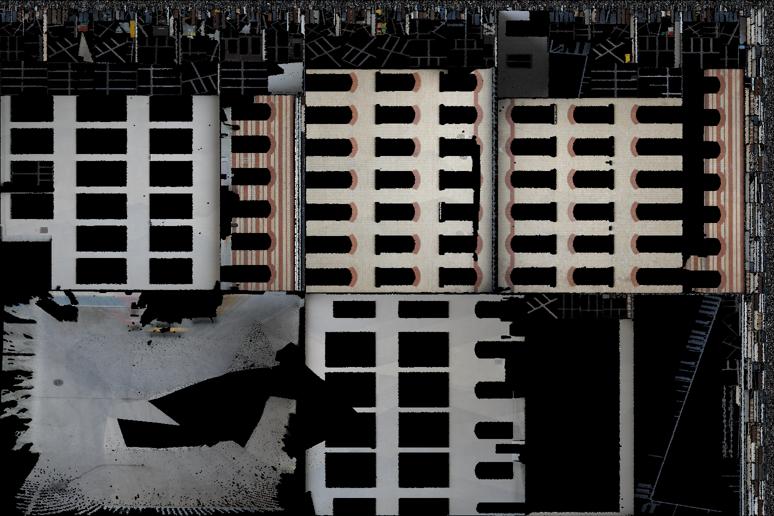}     
    & \includegraphics[width=0.16\textwidth]{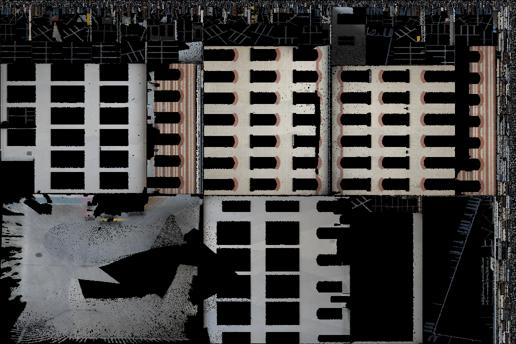}
    & \includegraphics[width=0.12\textwidth]{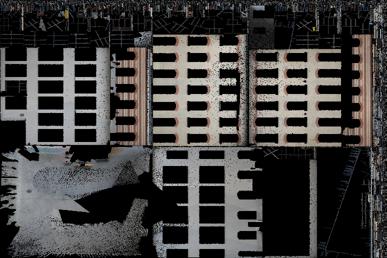} \\
    \emph{delivery$\_$area} & \includegraphics[width=0.48\textwidth]{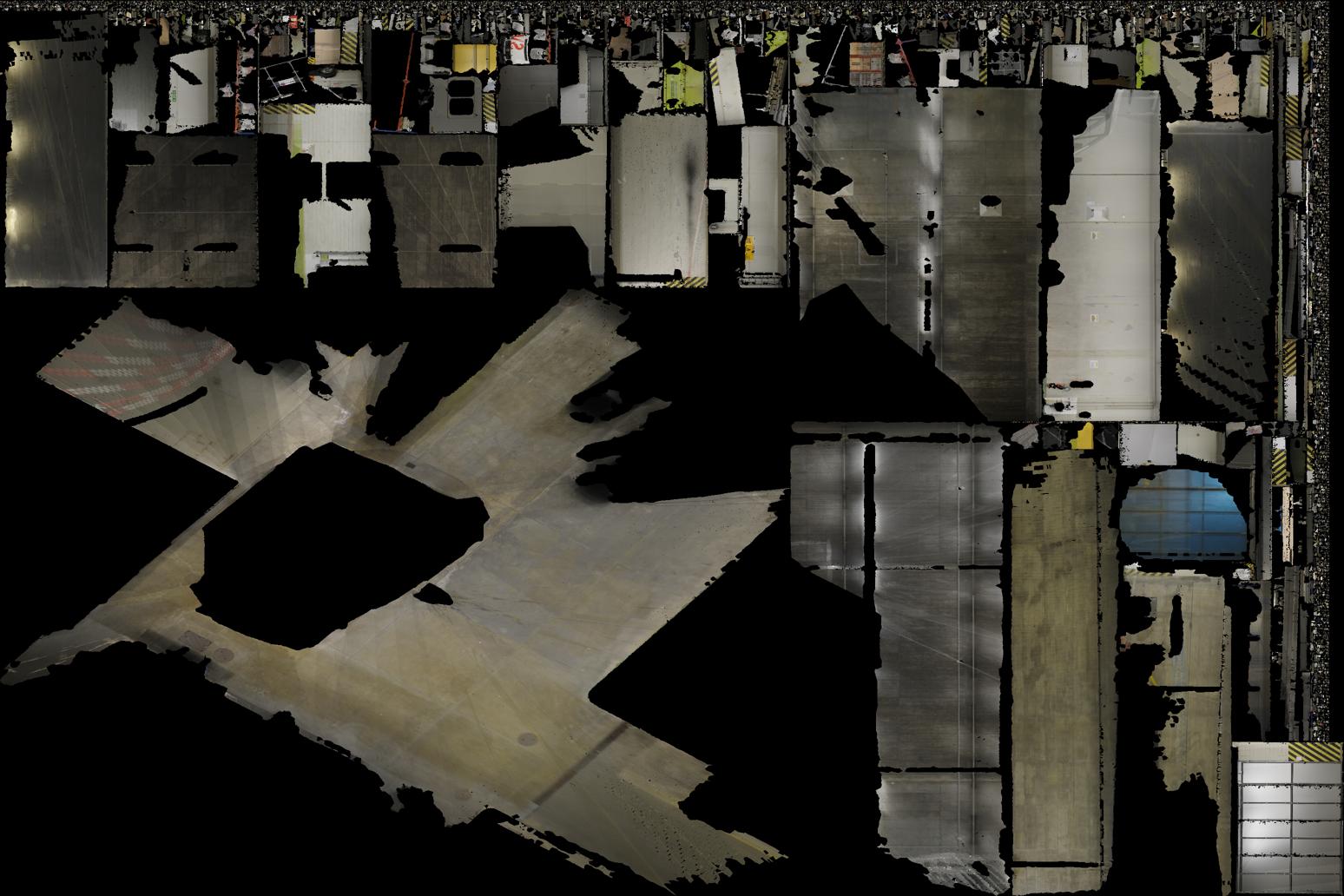} 
    & \includegraphics[width=0.24\textwidth]{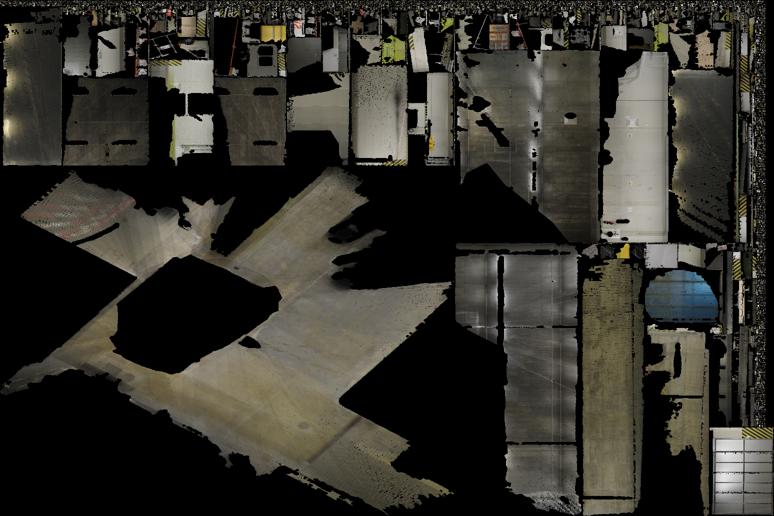}     
    & \includegraphics[width=0.16\textwidth]{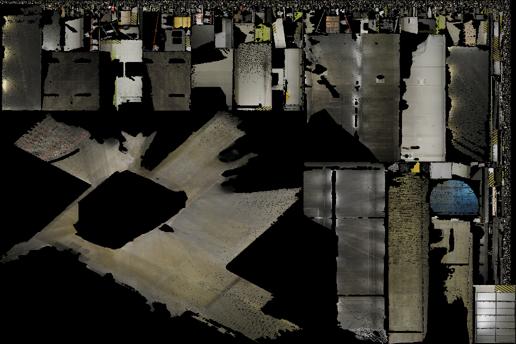}
    & \includegraphics[width=0.12\textwidth]{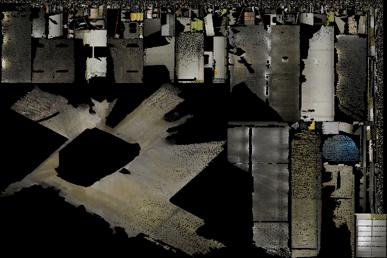} \\
    \emph{pipes} & \includegraphics[width=0.48\textwidth]{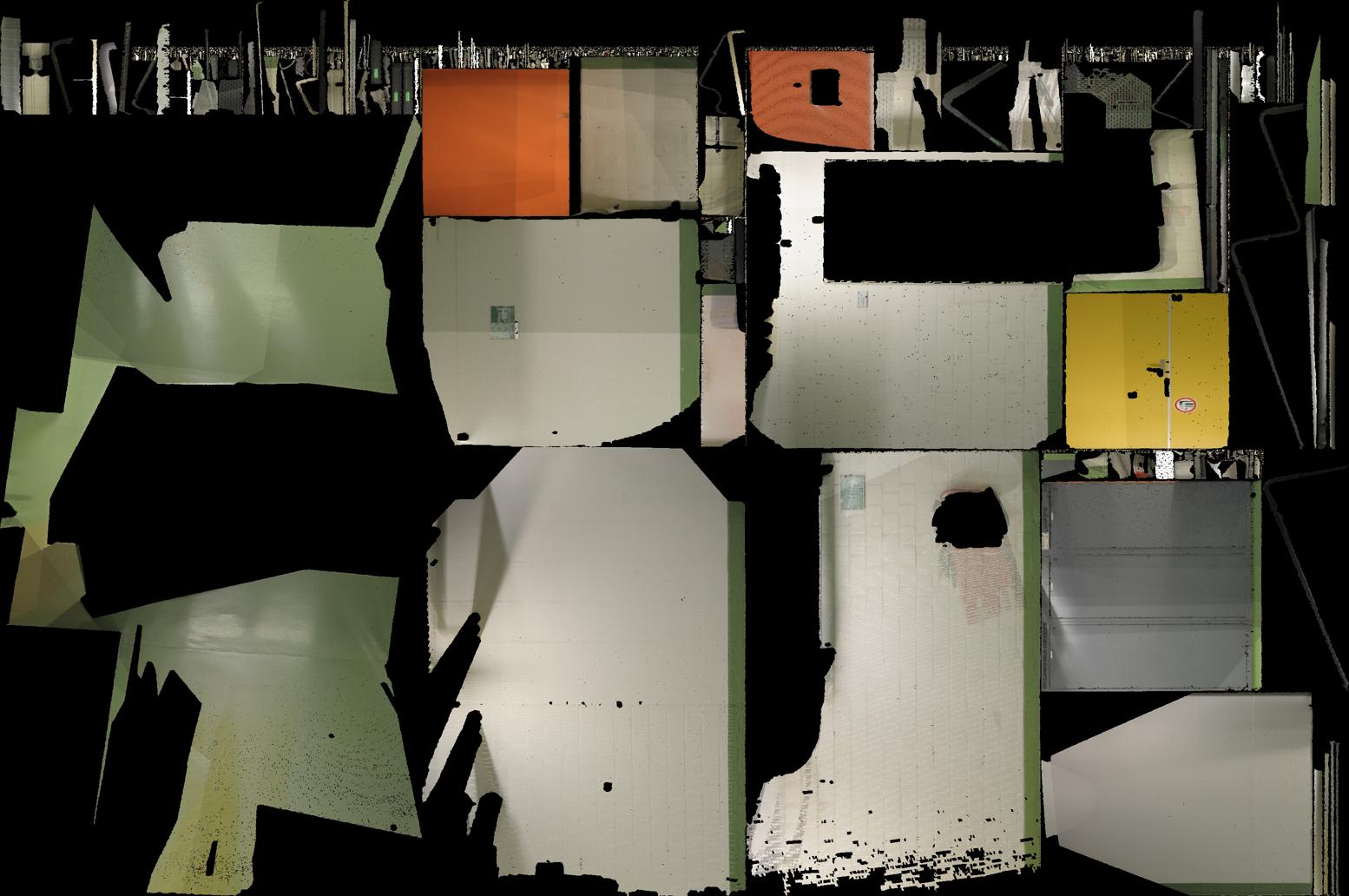} 
    & \includegraphics[width=0.24\textwidth]{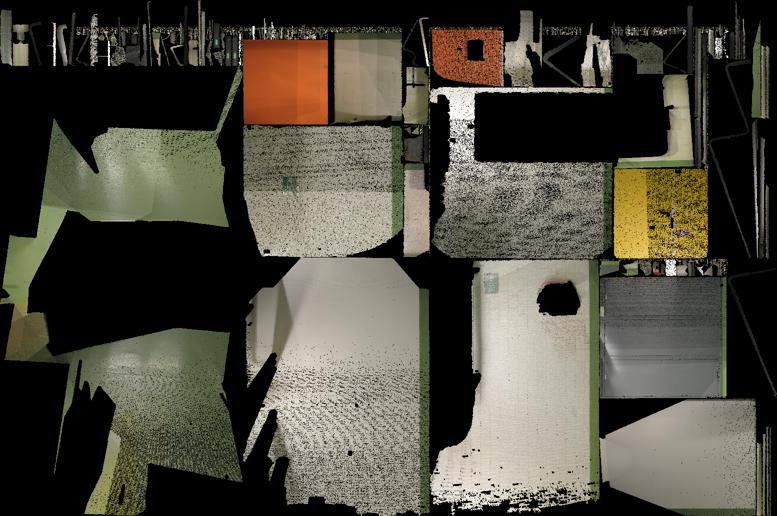}     
    & \includegraphics[width=0.16\textwidth]{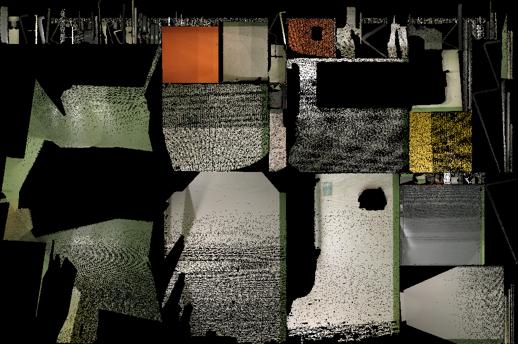}
    & \includegraphics[width=0.12\textwidth]{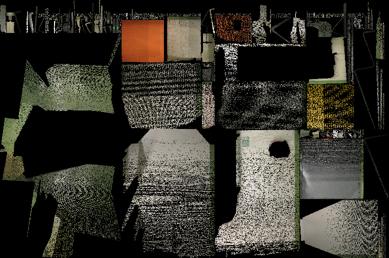} \\
    \emph{terrace} & \includegraphics[width=0.48\textwidth]{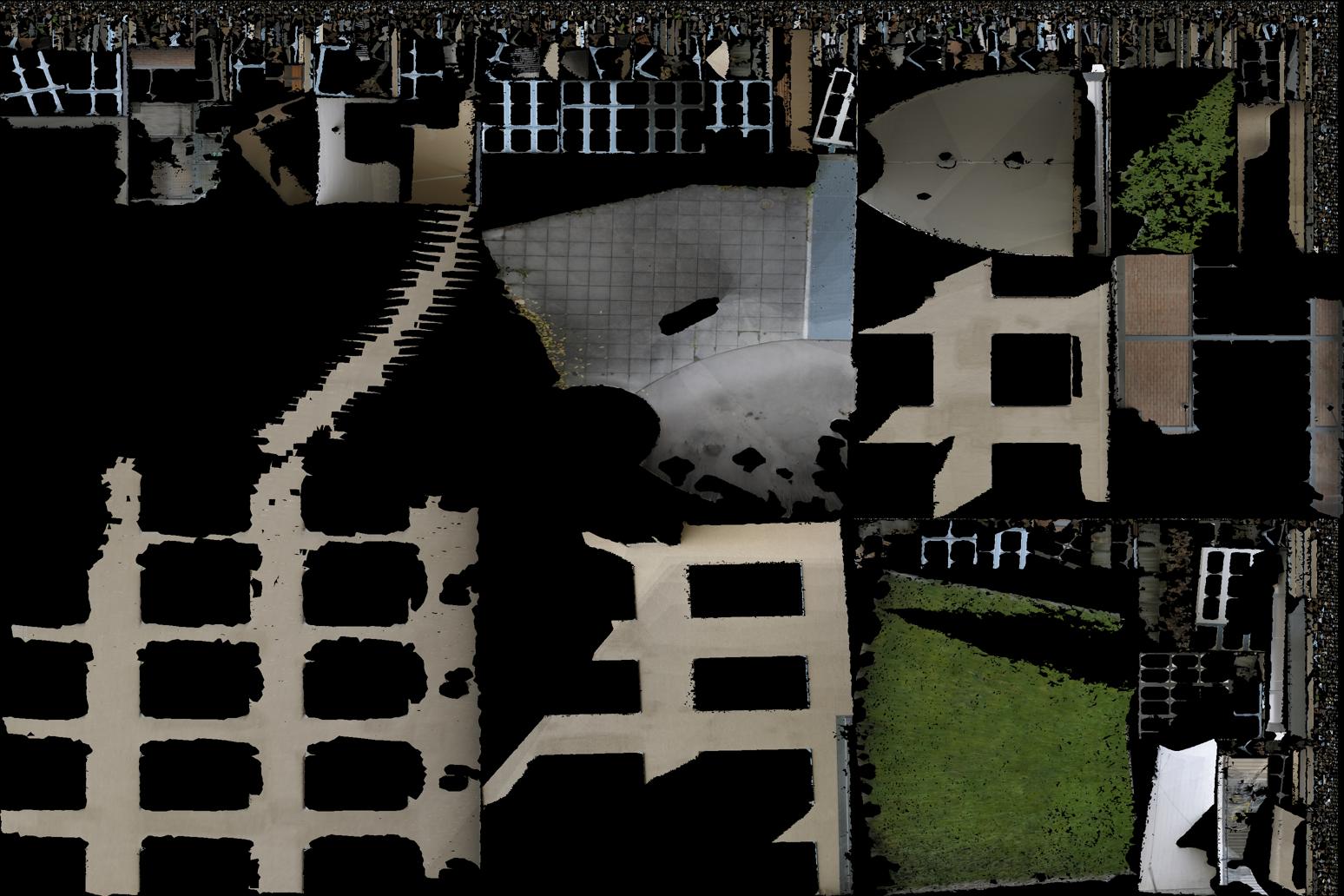} 
    & \includegraphics[width=0.24\textwidth]{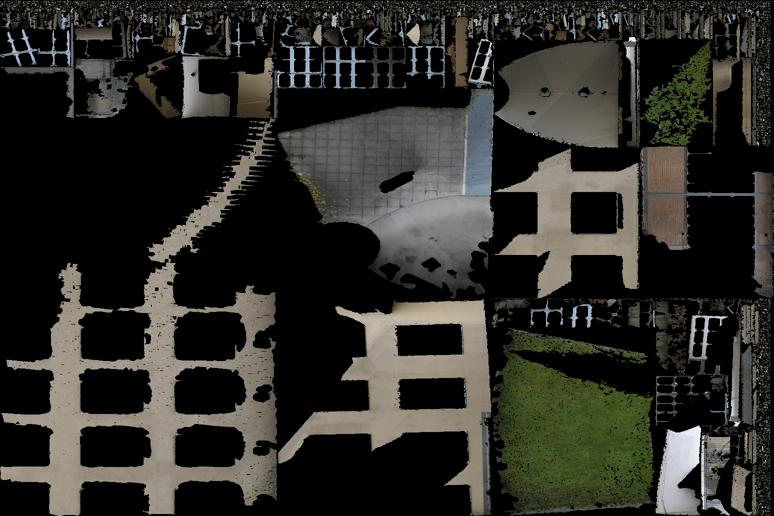}     
    & \includegraphics[width=0.16\textwidth]{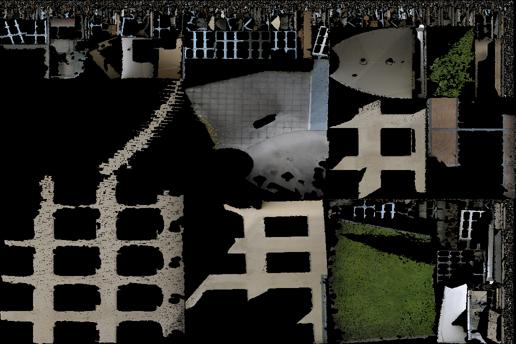}
    & \includegraphics[width=0.12\textwidth]{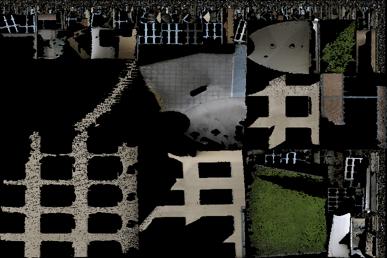} \\
\end{tabular}
}
\caption{The texture maps of \emph{courtyard}, \emph{delivery$\_$area}, \emph{pipes}, and \emph{terrace} for different resolutions.}
\label{fig:texture_maps1}
\end{figure*}

\begin{figure*}
\setlength{\tabcolsep}{1pt}
\renewcommand{\arraystretch}{0.3}
\resizebox{\linewidth}{!}
{
\begin{tabular}{ccccc}
    &Groud Truth & $\times 2$ LR  & $\times 3$ LR & $\times 4$ LR \\
    \emph{terrains} & \includegraphics[width=0.48\textwidth]{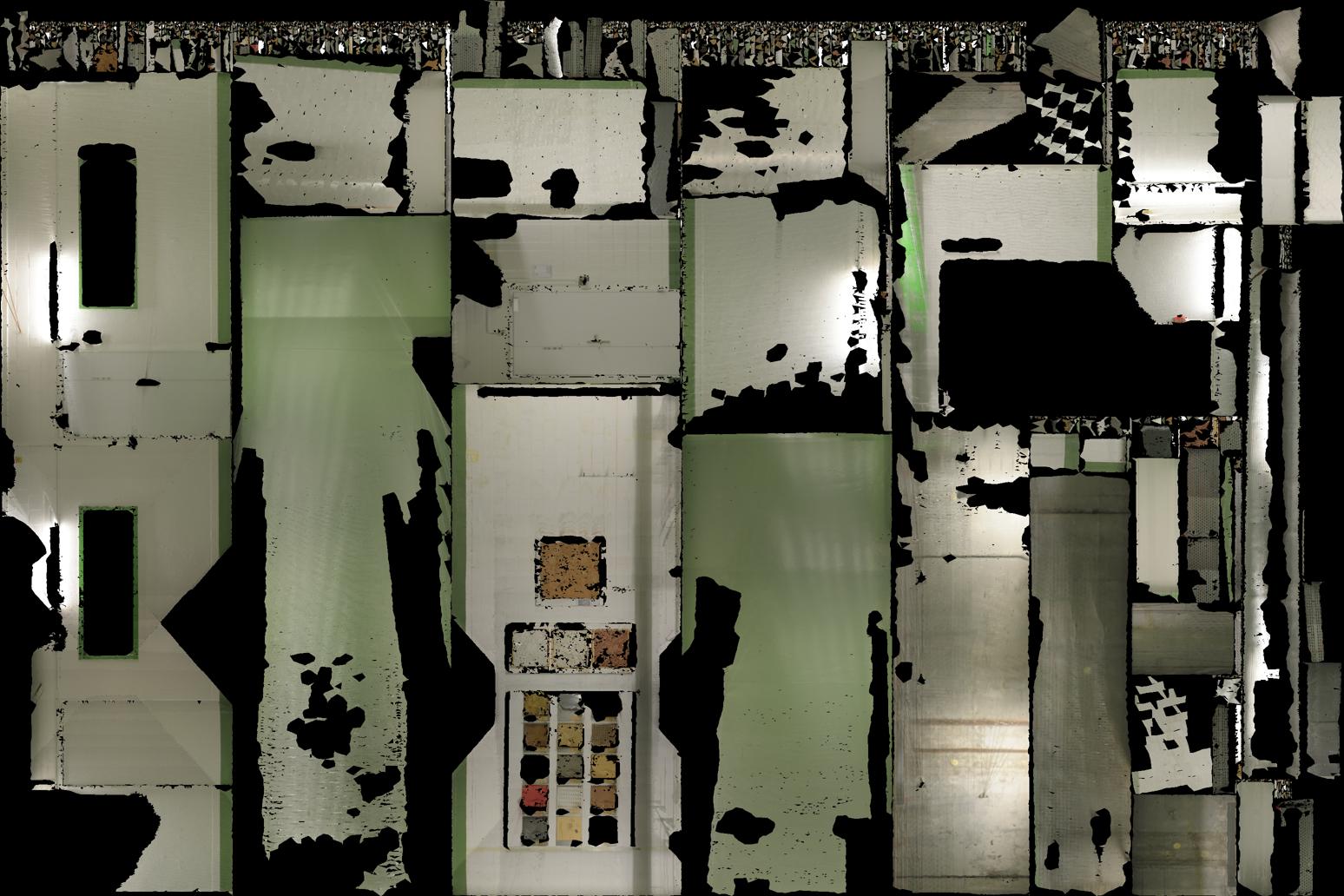} 
    & \includegraphics[width=0.24\textwidth]{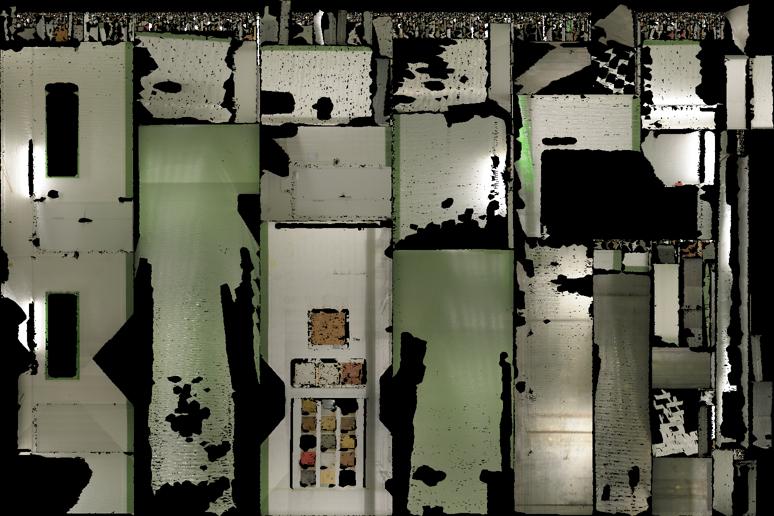}     
    & \includegraphics[width=0.16\textwidth]{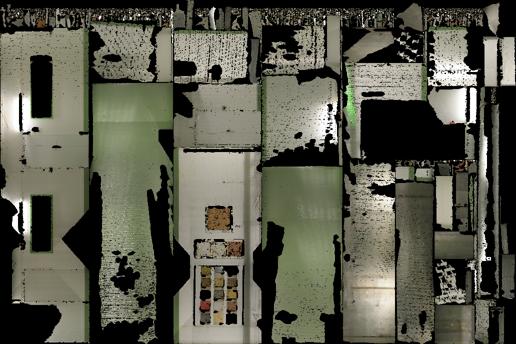}
    & \includegraphics[width=0.12\textwidth]{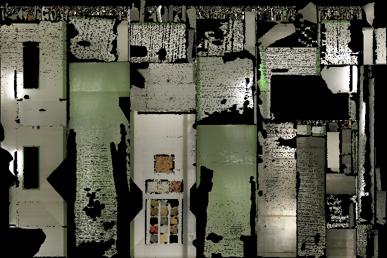} \\
    \emph{Skull} & \includegraphics[width=0.48\textwidth]{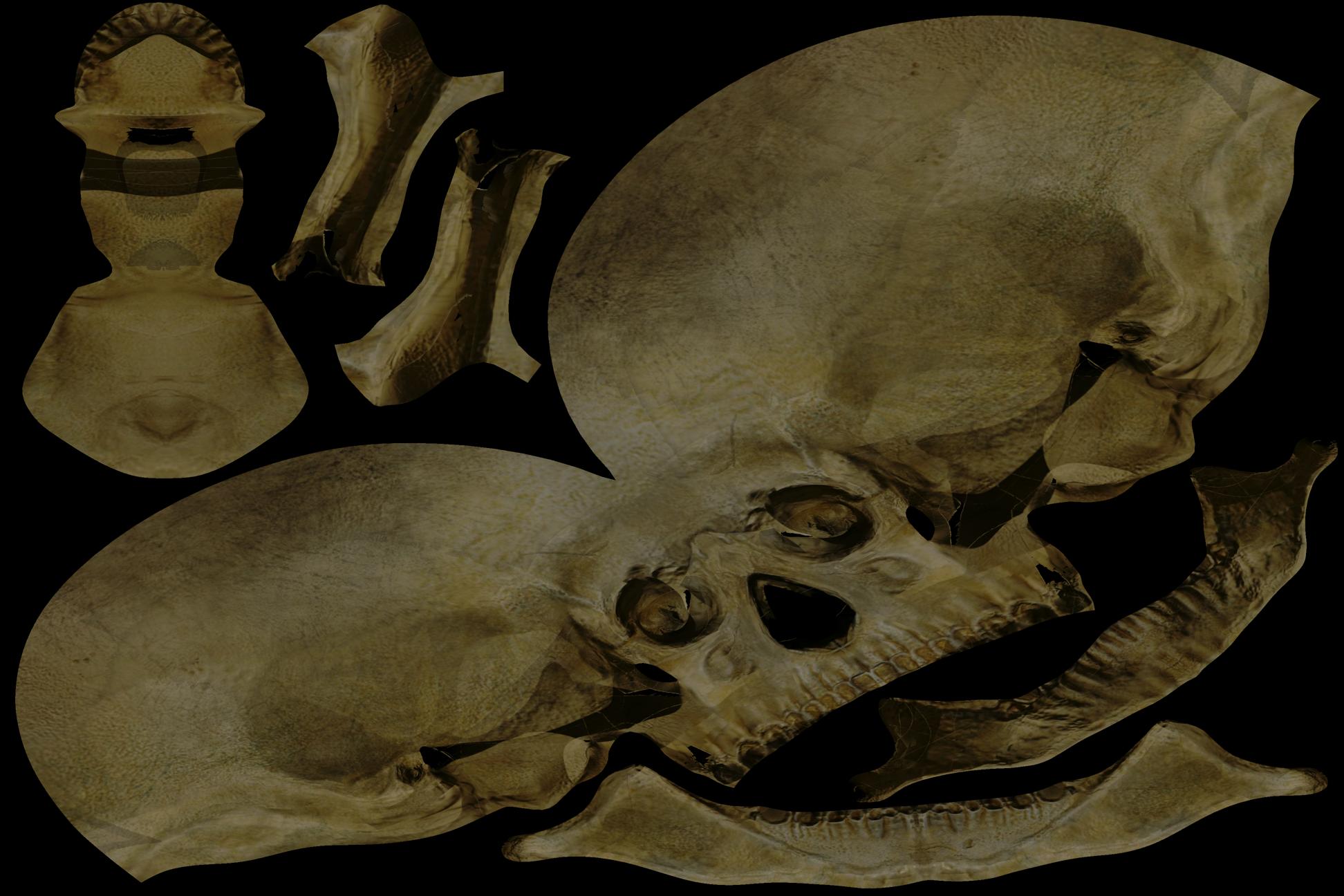} 
    & \includegraphics[width=0.24\textwidth]{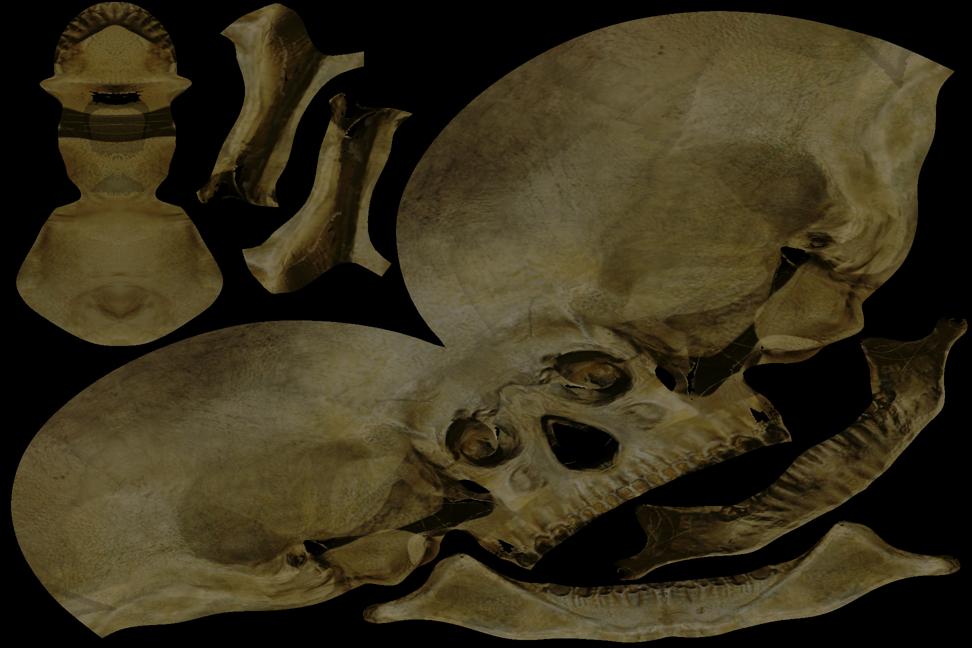}     
    & \includegraphics[width=0.16\textwidth]{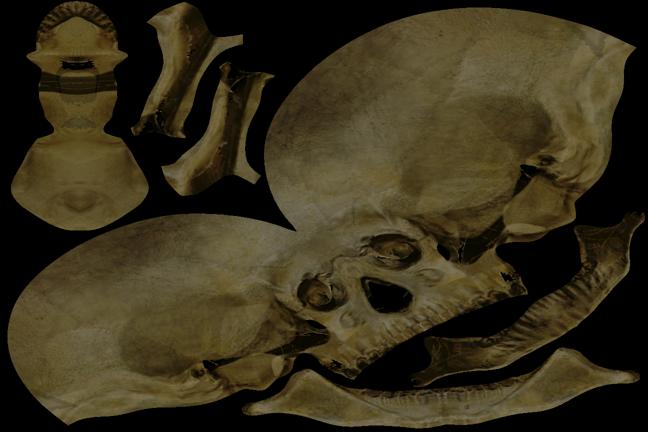}
    & \includegraphics[width=0.12\textwidth]{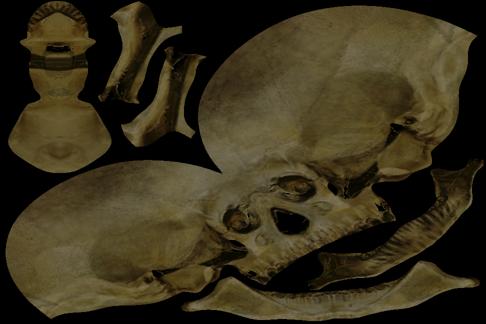} \\
    \emph{GeologicalSample} & \includegraphics[width=0.48\textwidth]{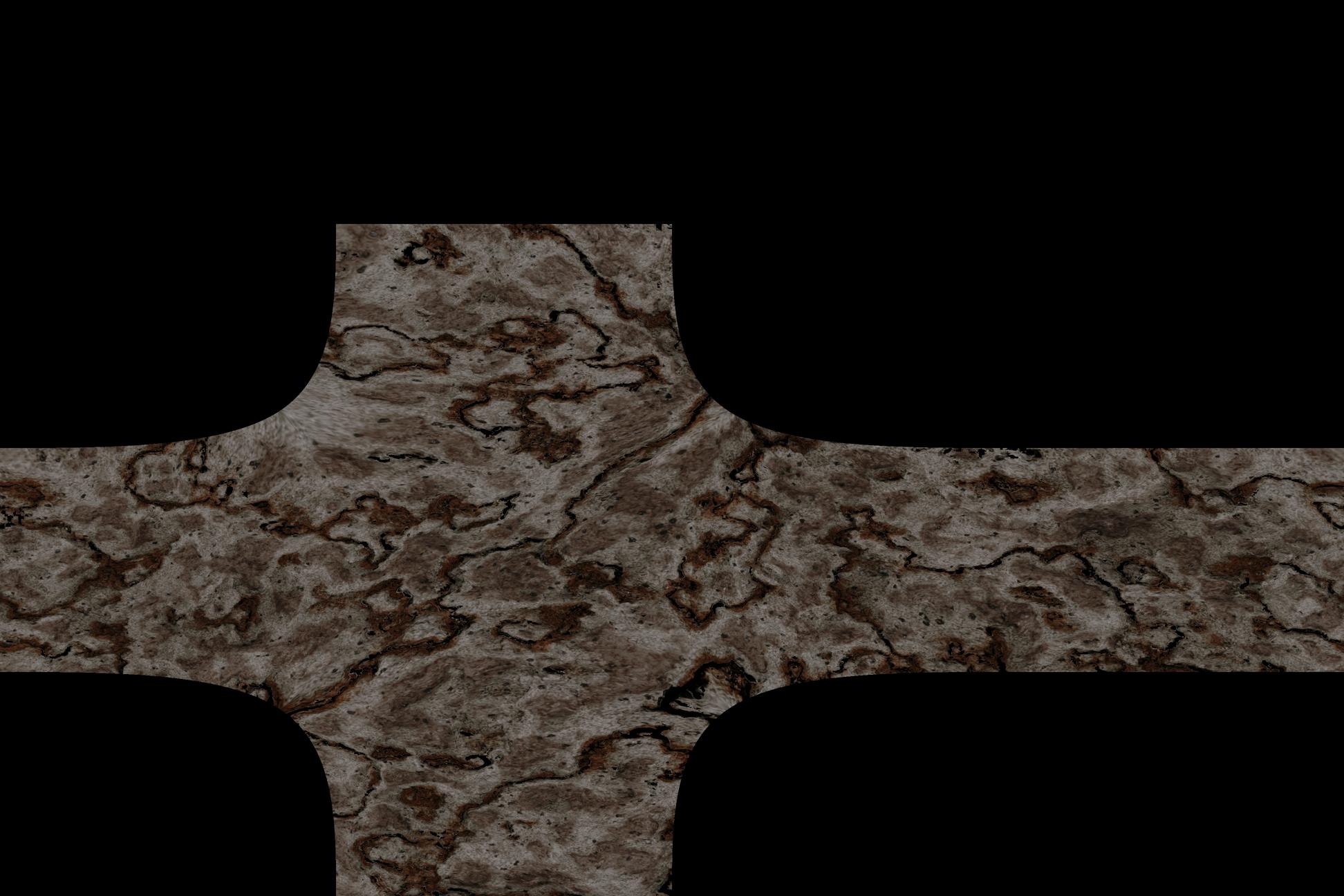} 
    & \includegraphics[width=0.24\textwidth]{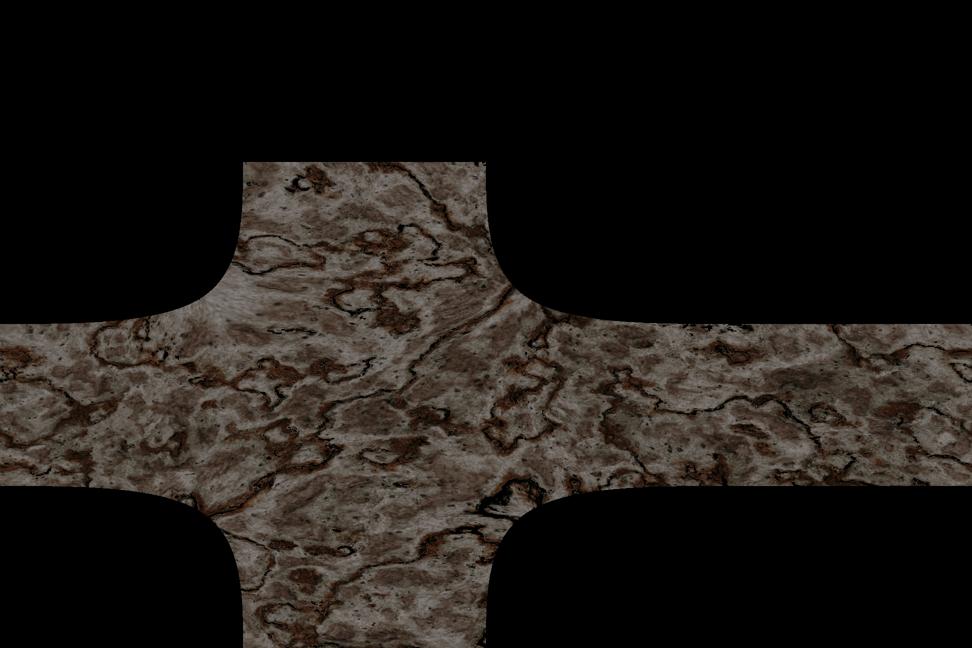}     
    & \includegraphics[width=0.16\textwidth]{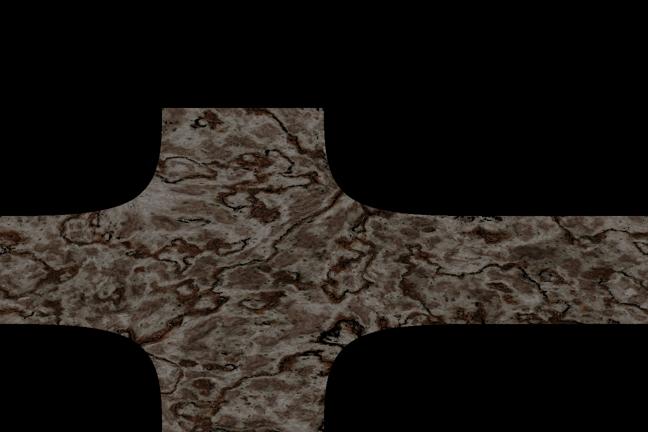}
    & \includegraphics[width=0.12\textwidth]{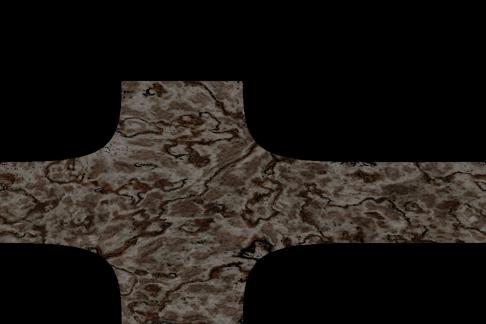} \\
    \emph{DinoRing} & \includegraphics[width=0.48\textwidth]{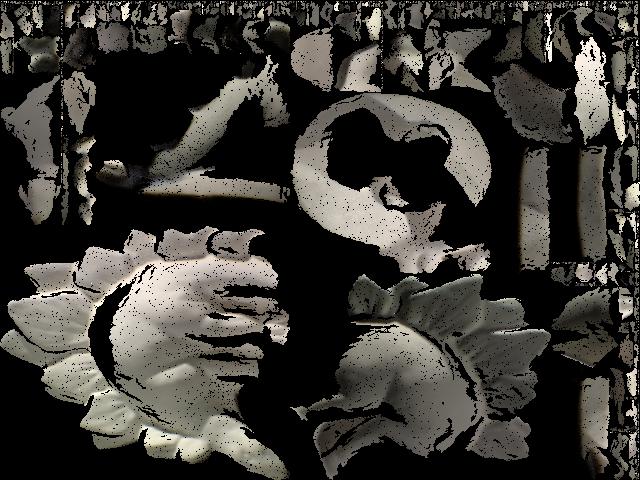} 
    & \includegraphics[width=0.24\textwidth]{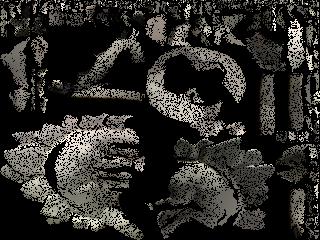}     
    & \includegraphics[width=0.16\textwidth]{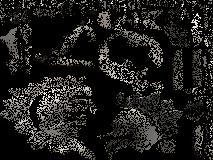}
    & \includegraphics[width=0.12\textwidth]{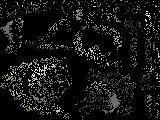} \\
\end{tabular}
}
\caption{The texture maps of \emph{terrains}, \emph{Skull}, \emph{GeologicalSample}, and \emph{DinoRing} for different resolutions.}
\label{fig:texture_maps2}
\end{figure*}

\begin{figure*}
\setlength{\tabcolsep}{1pt}
\renewcommand{\arraystretch}{0.3}
\resizebox{0.95\linewidth}{!}
{
\begin{tabular}{ccccc}
    &Groud Truth & $\times 2$ LR  & $\times 3$ LR & $\times 4$ LR \\
    \emph{Bird} & \includegraphics[width=0.48\textwidth]{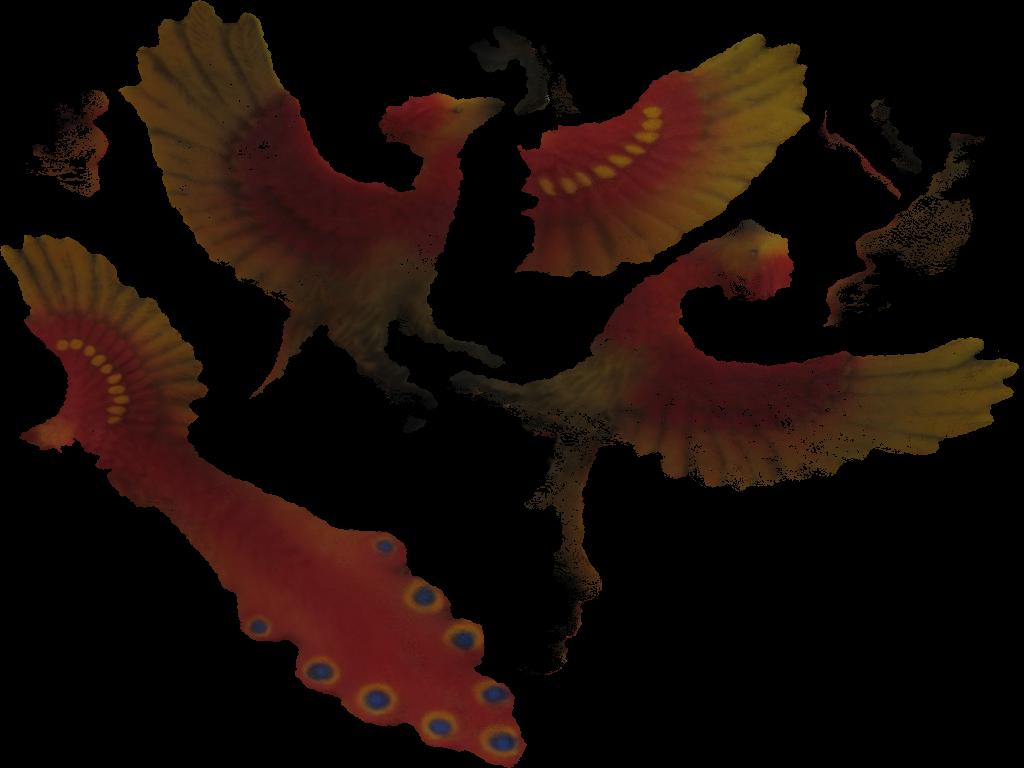} 
    & \includegraphics[width=0.24\textwidth]{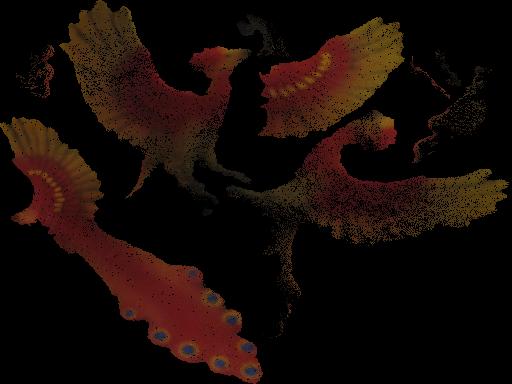}     
    & \includegraphics[width=0.16\textwidth]{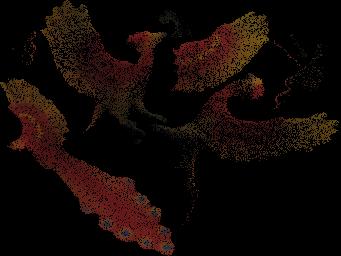}
    & \includegraphics[width=0.12\textwidth]{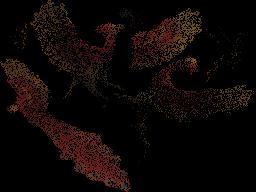} \\
    \emph{Buddha} & \includegraphics[width=0.48\textwidth]{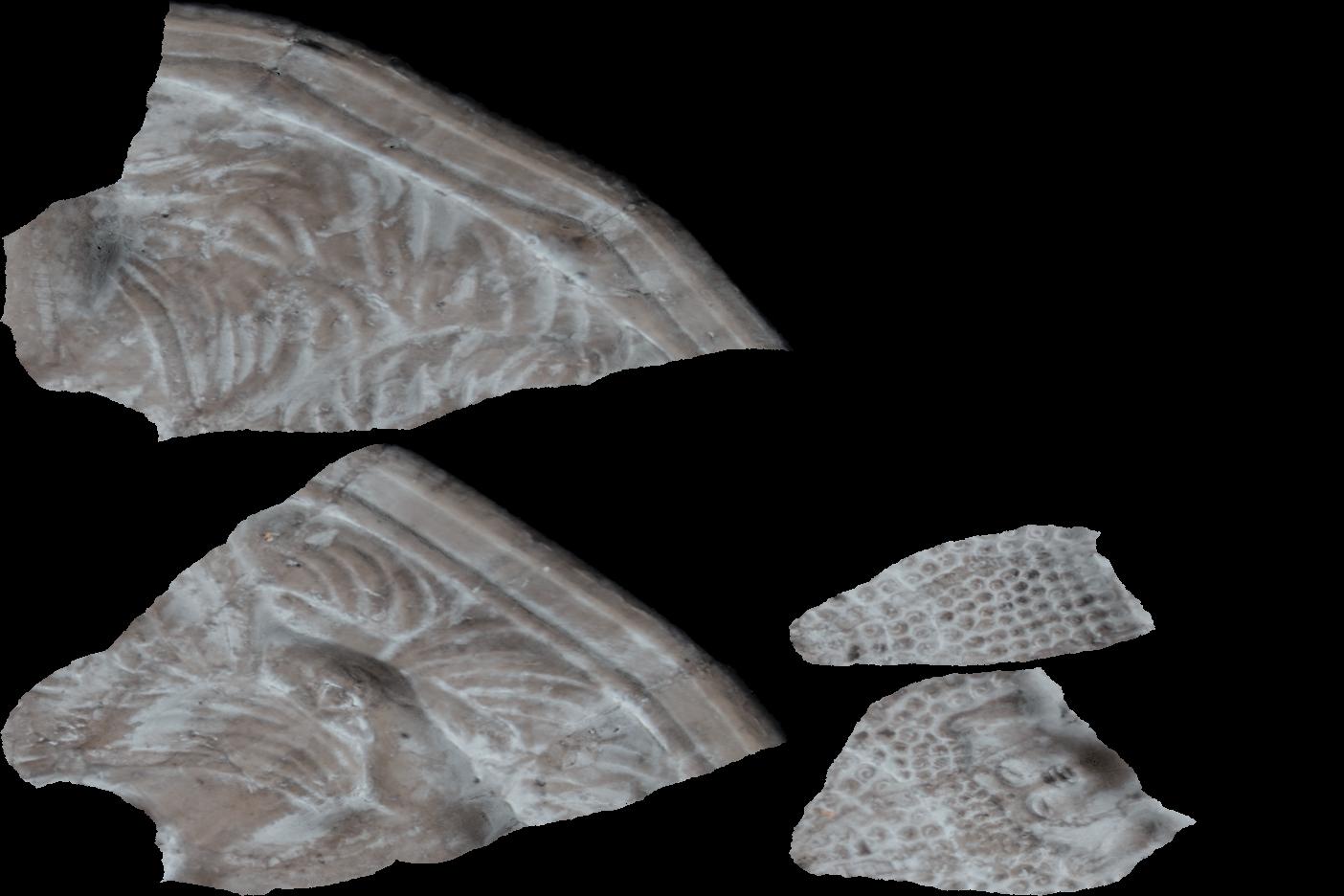} 
    & \includegraphics[width=0.24\textwidth]{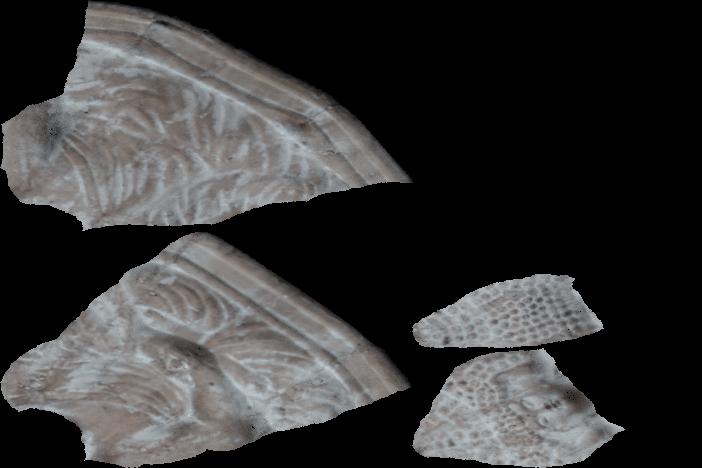}     
    & \includegraphics[width=0.16\textwidth]{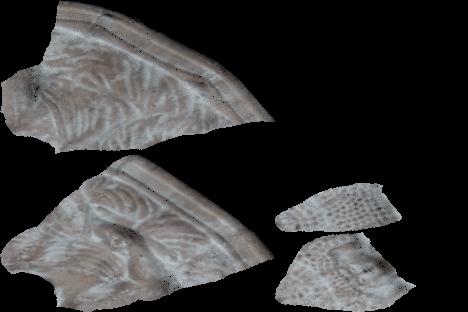}
    & \includegraphics[width=0.12\textwidth]{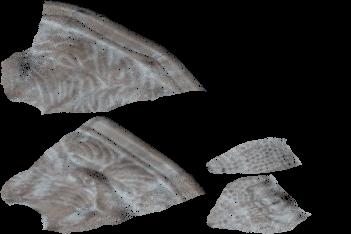} \\
    \emph{Bunny} & \includegraphics[width=0.48\textwidth]{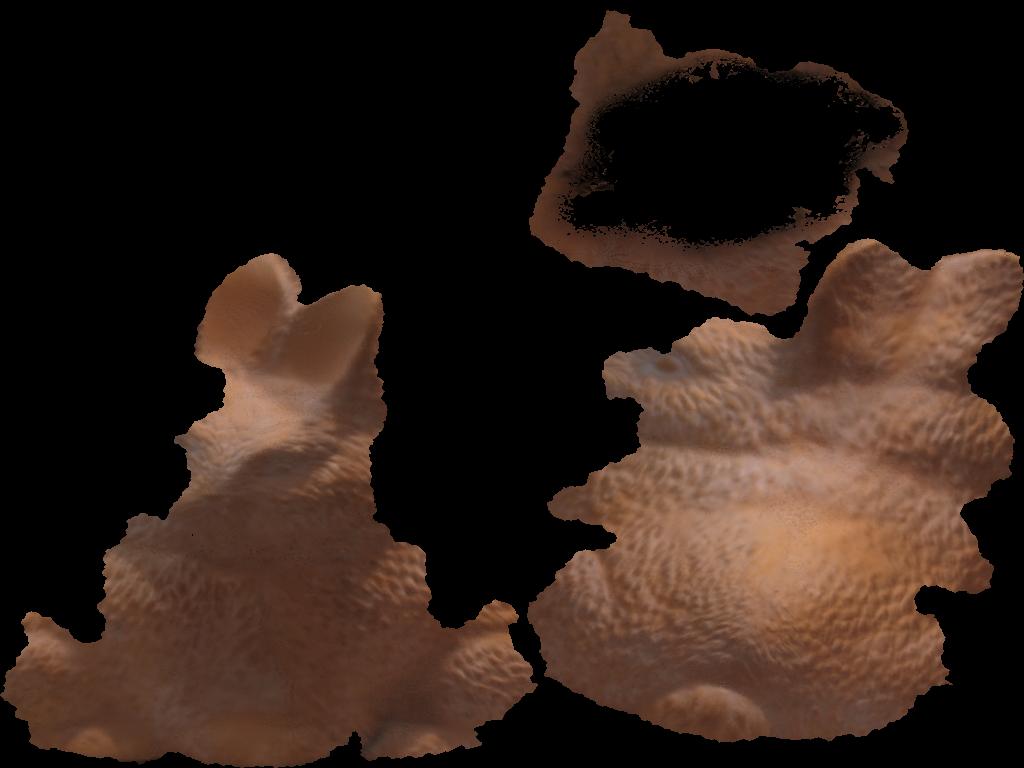} 
    & \includegraphics[width=0.24\textwidth]{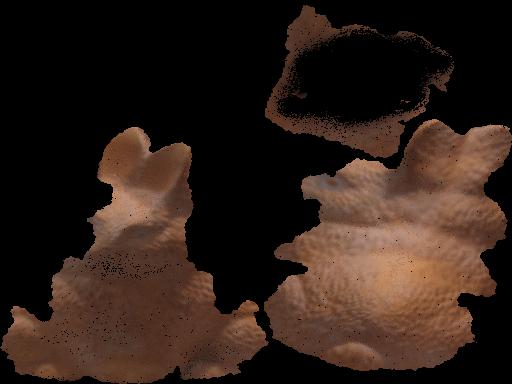}     
    & \includegraphics[width=0.16\textwidth]{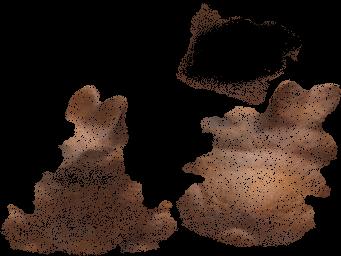}
    & \includegraphics[width=0.12\textwidth]{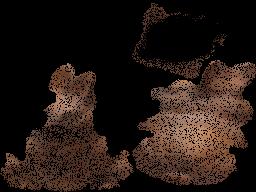} \\
    \emph{Fountain} & \includegraphics[width=0.48\textwidth]{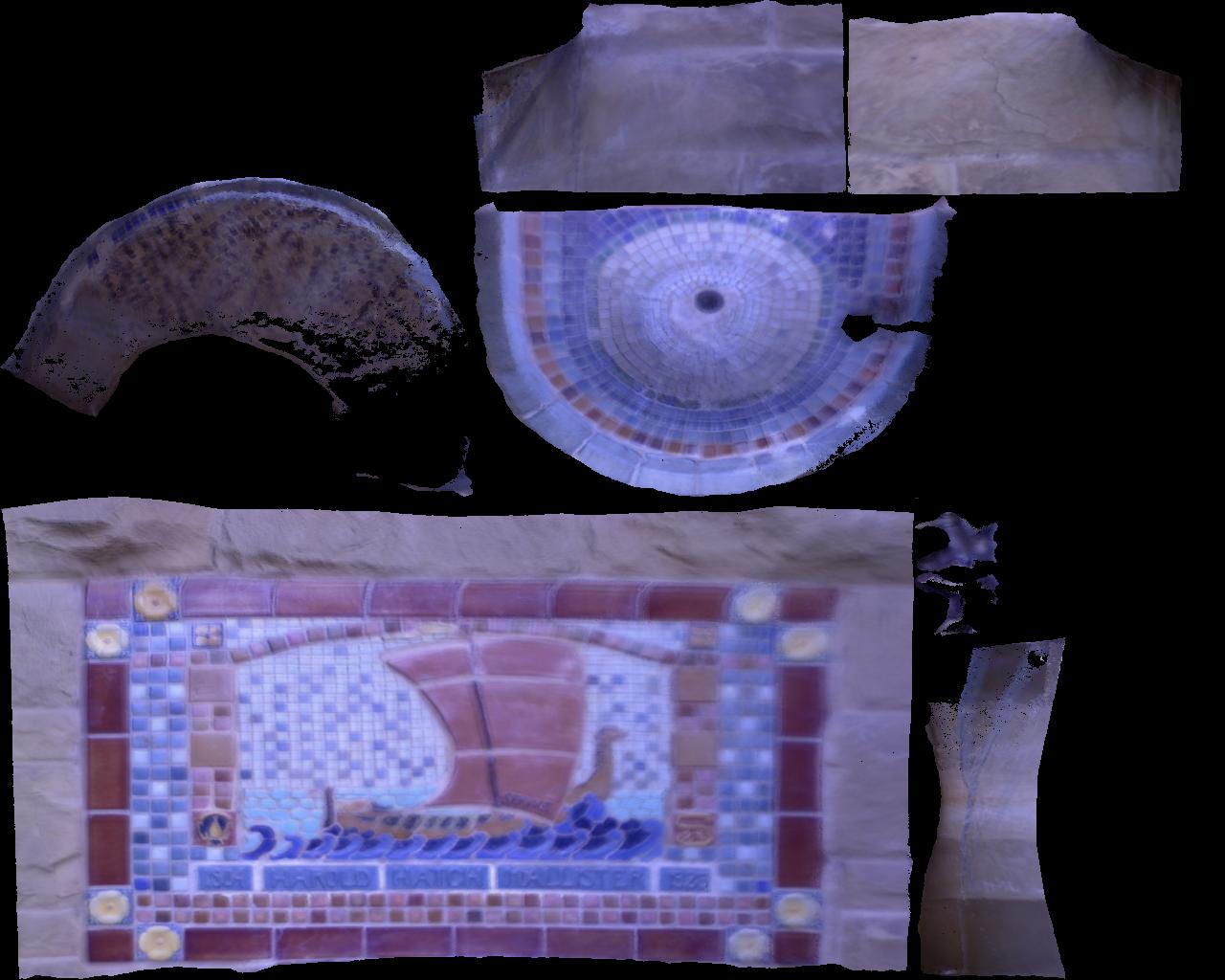} 
    & \includegraphics[width=0.24\textwidth]{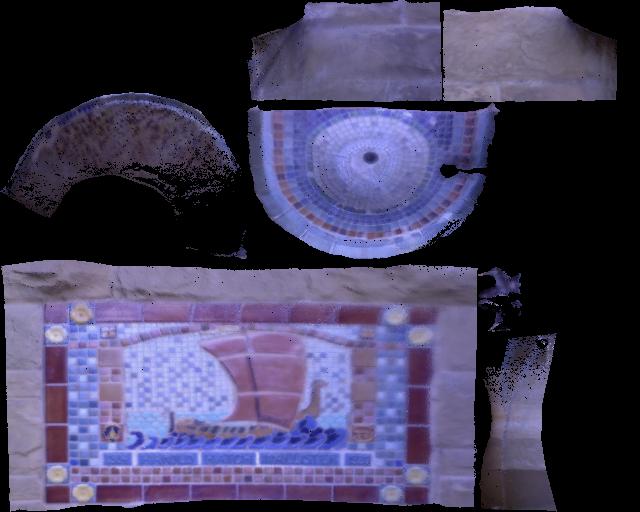}     
    & \includegraphics[width=0.16\textwidth]{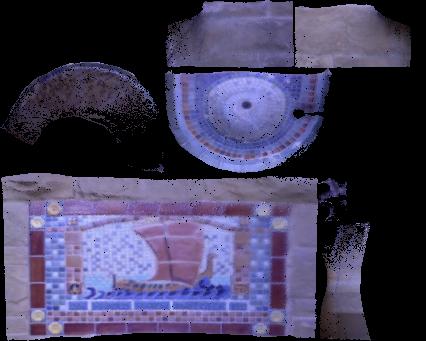}
    & \includegraphics[width=0.12\textwidth]{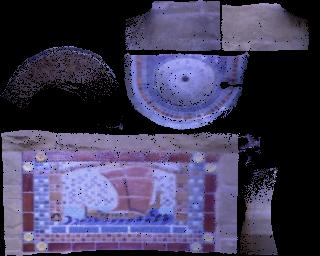} \\
\end{tabular}
}
\caption{The texture maps of \emph{Bird}, \emph{Buddha}, \emph{Bunny}, and \emph{Fountain} for different resolutions.}
\label{fig:texture_maps3}
\end{figure*}

Twelve of the 24 texture maps for different resolutions (HR, $\times 2$, $\times 3$, $\times 4$ down-sampling) are shown in Fig.~\ref{fig:texture_maps1}, Fig.~\ref{fig:texture_maps2}, and Fig.~\ref{fig:texture_maps3}, respectively. By comparing the texure maps with differnt resolutions, we find that the ground truth texture maps contain more details than the LR ones. In addition, the HR texture maps are denser than the LR ones.
Since optimal UV parameters exist for the synthetic scenes \emph{GeologicalSample}, \emph{Toad}, and \emph{Skull}, their texture maps have less disconnected support regions.

\section{SR Results}

\begin{figure*}[t]
\centering

\begin{minipage}[t]{0.32\linewidth}
\centering
\includegraphics[width=1\textwidth]{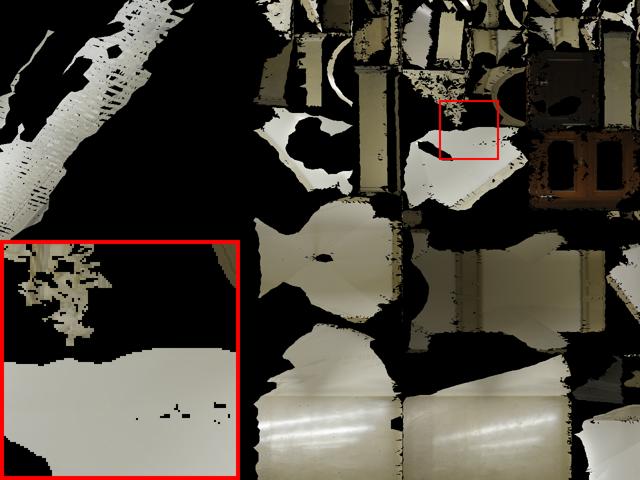}
{\footnotesize  Ground Truth}
\vspace{-2mm}
\end{minipage}
\begin{minipage}[t]{0.32\linewidth}
\centering
\includegraphics[width=1\textwidth]{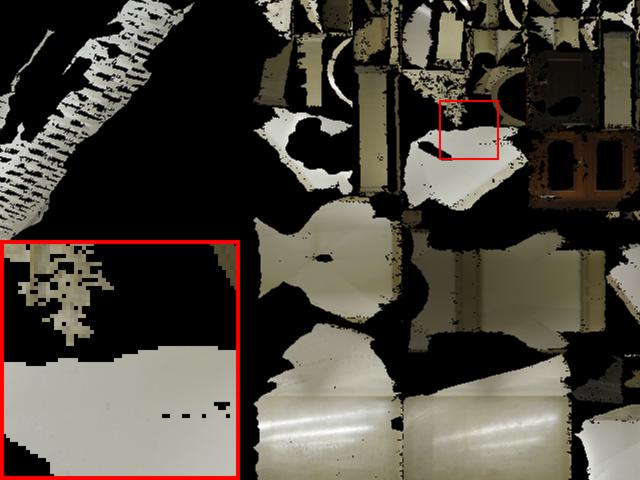}
{\footnotesize  Nearest}
\vspace{-2mm}
\end{minipage}
\begin{minipage}[t]{0.32\linewidth}
\centering
\includegraphics[width=1\textwidth]{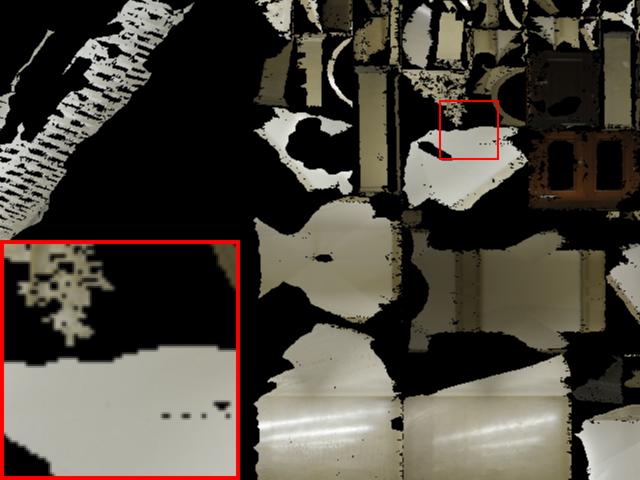}
{\footnotesize  Bilinear}
\vspace{-2mm}
\end{minipage}

\begin{minipage}[t]{0.32\linewidth}
\centering
\includegraphics[width=1\textwidth]{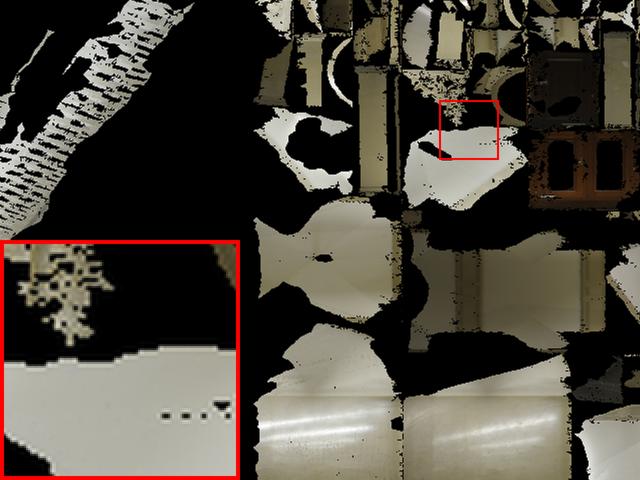}
{\footnotesize  Lanczos}
\end{minipage}
\begin{minipage}[t]{0.32\linewidth}
\centering
\includegraphics[width=1\textwidth]{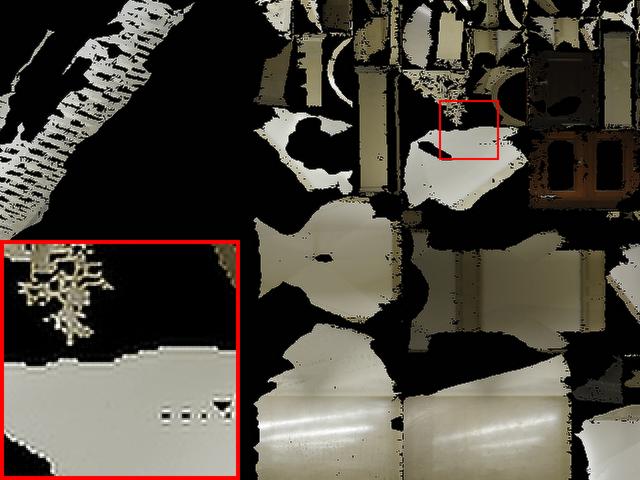}
{\footnotesize  FSRCNN}
\end{minipage}
\begin{minipage}[t]{0.32\linewidth}
\centering
\includegraphics[width=1\textwidth]{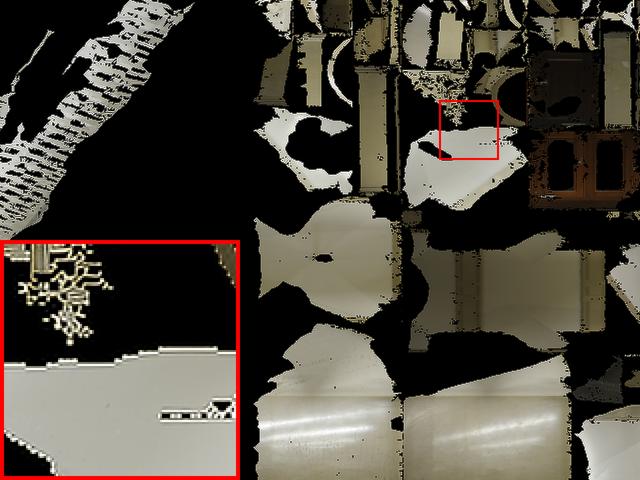}
{\footnotesize  SRResNet}
\end{minipage}

\begin{minipage}[t]{0.32\linewidth}
\centering
\includegraphics[width=1\textwidth]{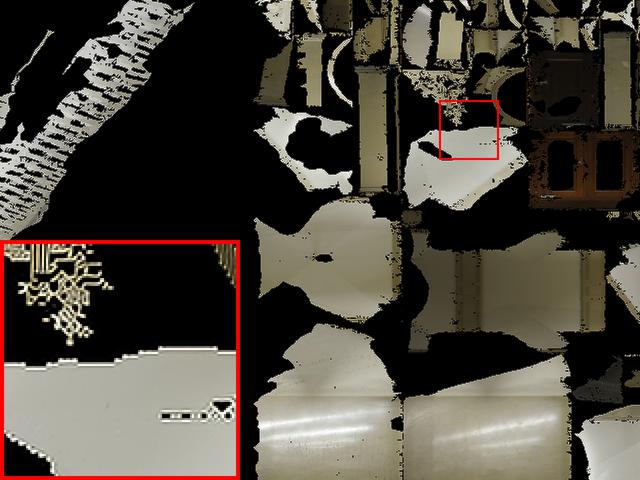}
{\footnotesize  EDSR}
\end{minipage}
\begin{minipage}[t]{0.32\linewidth}
\centering
\includegraphics[width=1\textwidth]{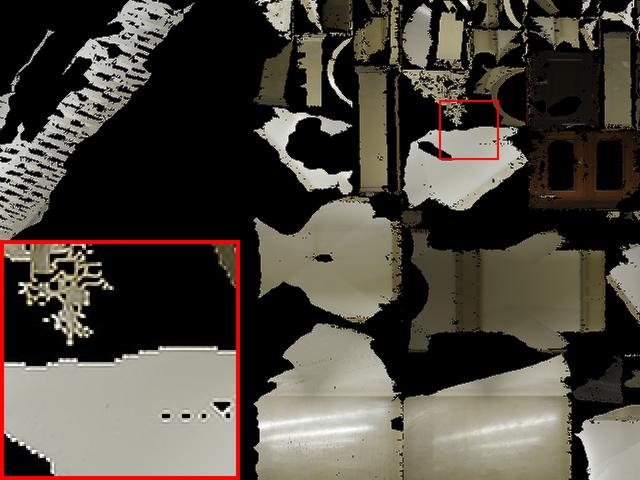}
{\footnotesize  RCAN}
\end{minipage}
\begin{minipage}[t]{0.32\linewidth}
\centering
\includegraphics[width=1\textwidth]{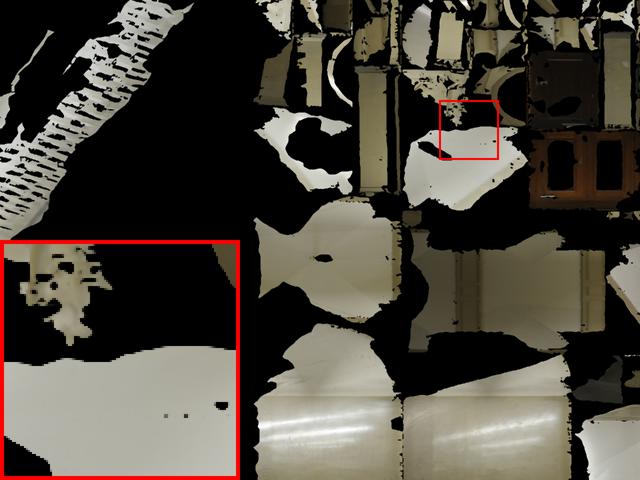}
{\footnotesize  EDSR-FT}
\end{minipage}

\begin{minipage}[t]{0.32\linewidth}
\centering
\includegraphics[width=1\textwidth]{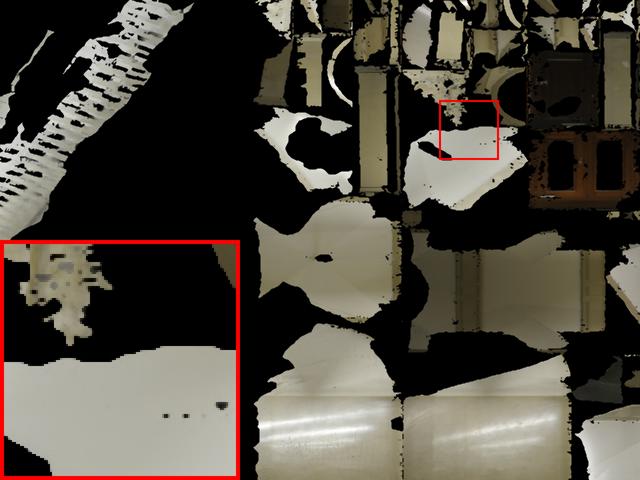}
{\footnotesize  NLR-Sub}
\end{minipage}
\begin{minipage}[t]{0.32\linewidth}
\centering
\includegraphics[width=1\textwidth]{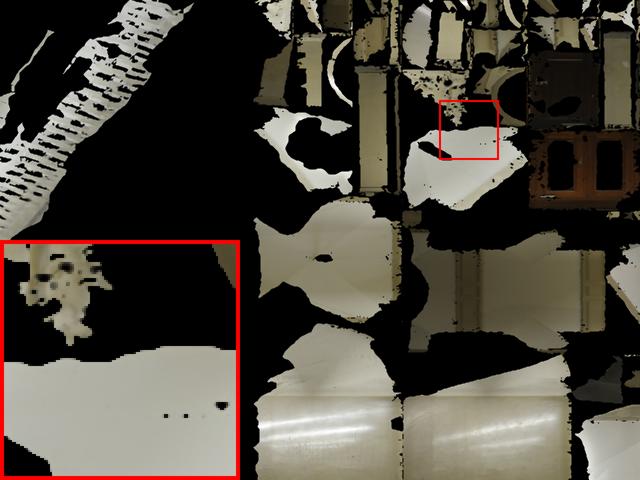}
{\footnotesize  NLR}
\end{minipage}
\begin{minipage}[t]{0.32\linewidth}
\centering
\includegraphics[width=1\textwidth]{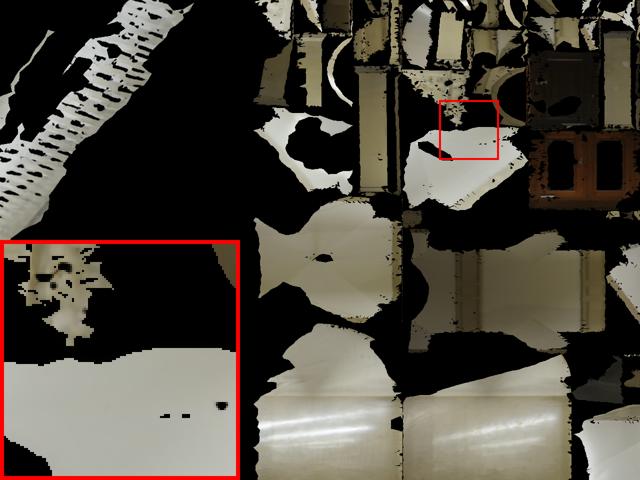}
{\footnotesize  NHR}
\end{minipage}
\caption{Texture map SR results of \emph{relief} by different methods.} 
\label{fig:sr_relief}
\end{figure*}

\begin{figure*}[t]
\centering

\begin{minipage}[t]{0.32\linewidth}
\centering
\includegraphics[width=1\textwidth]{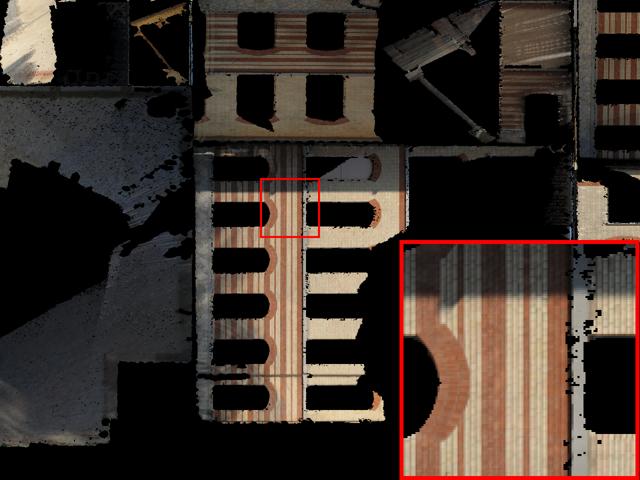}
{\footnotesize  Ground Truth}
\vspace{-2mm}
\end{minipage}
\begin{minipage}[t]{0.32\linewidth}
\centering
\includegraphics[width=1\textwidth]{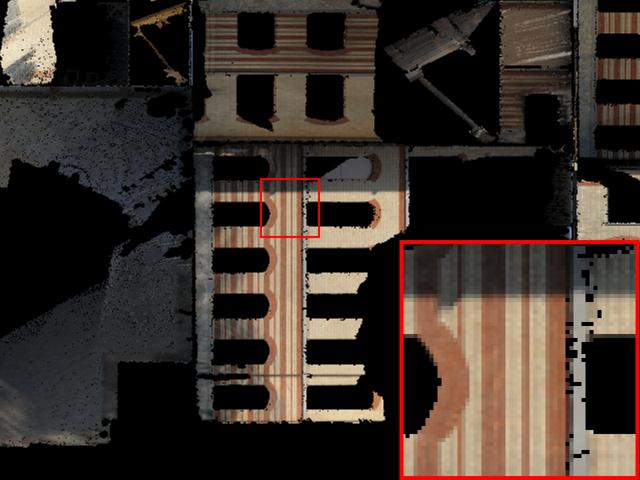}
{\footnotesize  Nearest}
\vspace{-2mm}
\end{minipage}
\begin{minipage}[t]{0.32\linewidth}
\centering
\includegraphics[width=1\textwidth]{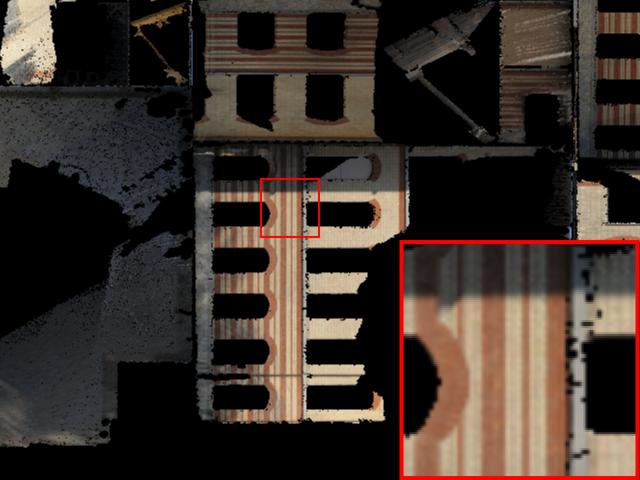}
{\footnotesize  Bilinear}
\vspace{-2mm}
\end{minipage}

\begin{minipage}[t]{0.32\linewidth}
\centering
\includegraphics[width=1\textwidth]{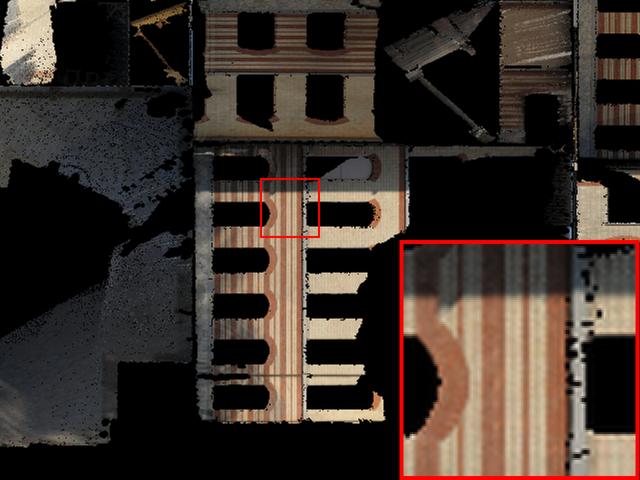}
{\footnotesize  Lanczos}
\end{minipage}
\begin{minipage}[t]{0.32\linewidth}
\centering
\includegraphics[width=1\textwidth]{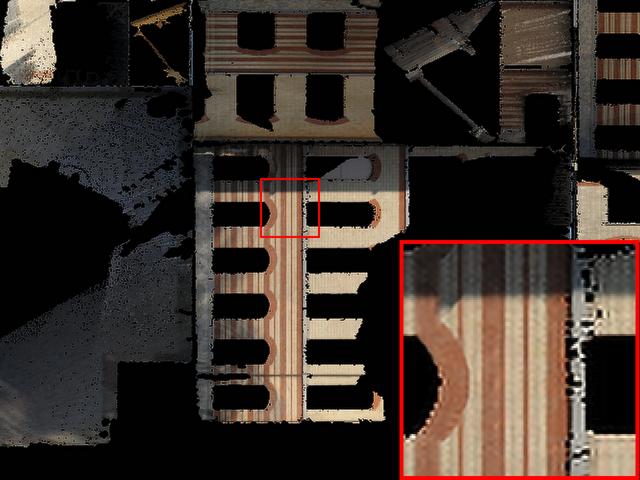}
{\footnotesize  FSRCNN}
\end{minipage}
\begin{minipage}[t]{0.32\linewidth}
\centering
\includegraphics[width=1\textwidth]{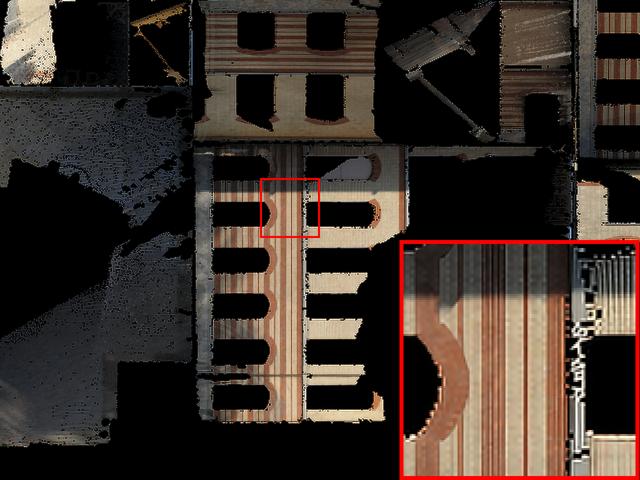}
{\footnotesize  SRResNet}
\end{minipage}

\begin{minipage}[t]{0.32\linewidth}
\centering
\includegraphics[width=1\textwidth]{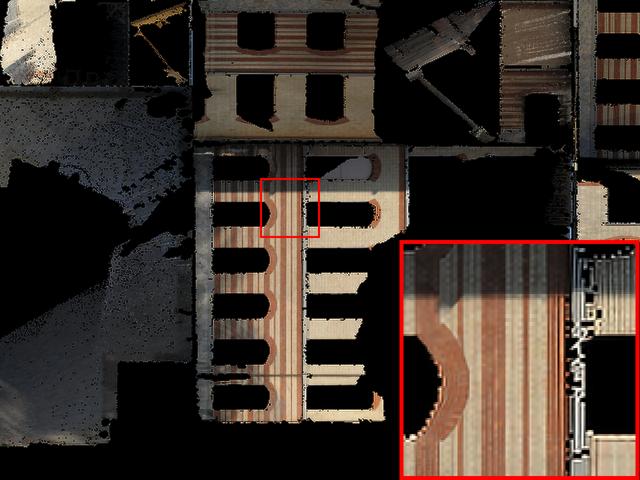}
{\footnotesize  EDSR}
\end{minipage}
\begin{minipage}[t]{0.32\linewidth}
\centering
\includegraphics[width=1\textwidth]{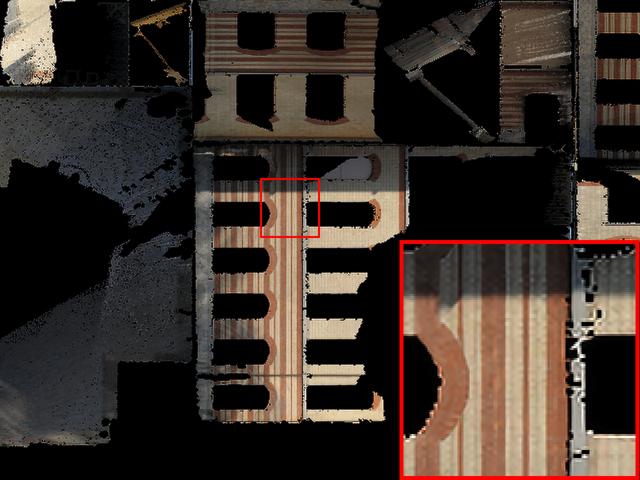}
{\footnotesize  RCAN}
\end{minipage}
\begin{minipage}[t]{0.32\linewidth}
\centering
\includegraphics[width=1\textwidth]{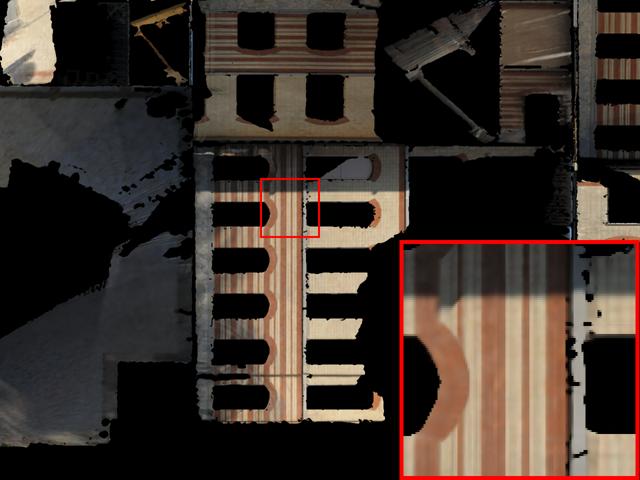}
{\footnotesize  EDSR-FT}
\end{minipage}

\begin{minipage}[t]{0.32\linewidth}
\centering
\includegraphics[width=1\textwidth]{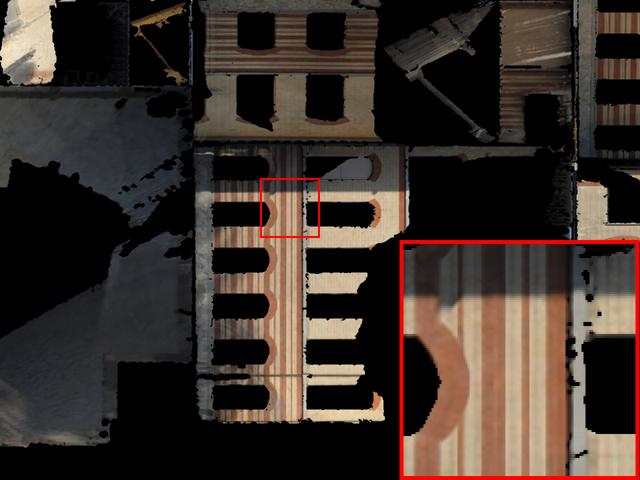}
{\footnotesize  NLR-Sub}
\end{minipage}
\begin{minipage}[t]{0.32\linewidth}
\centering
\includegraphics[width=1\textwidth]{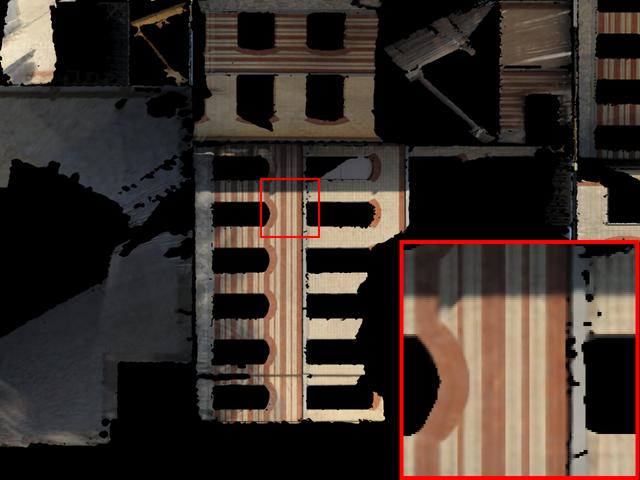}
{\footnotesize  NLR}
\end{minipage}
\begin{minipage}[t]{0.32\linewidth}
\centering
\includegraphics[width=1\textwidth]{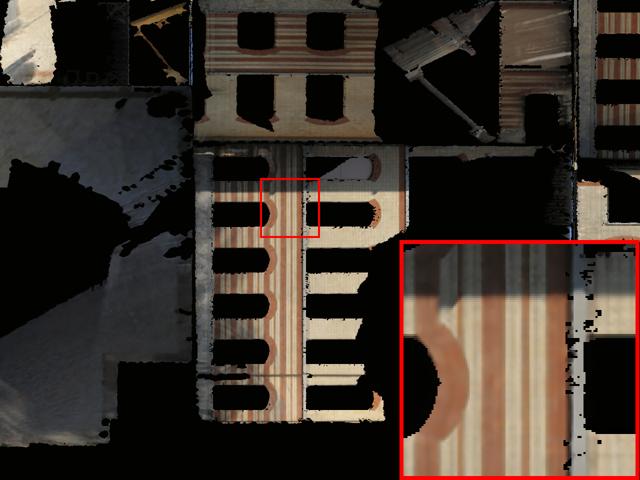}
{\footnotesize  NHR}
\end{minipage}

\caption{Texture map SR results of \emph{facade} by different methods.} 
\label{fig:sr_facade}
\end{figure*}

\begin{figure*}[t]
\centering

\begin{minipage}[t]{0.32\linewidth}
\centering
\includegraphics[width=1\textwidth]{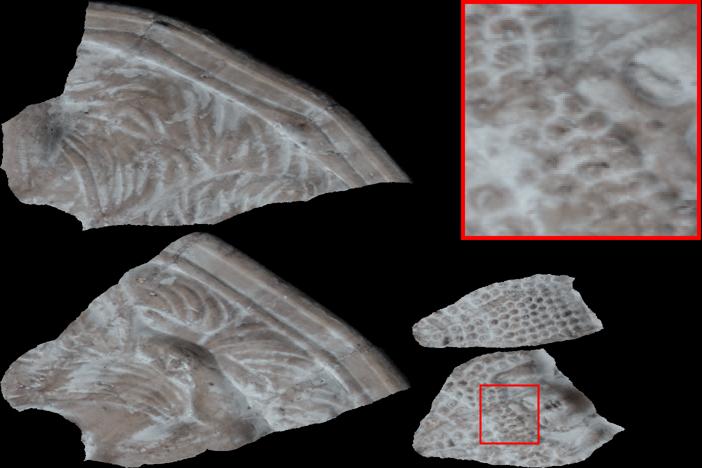}
{\footnotesize  Ground Truth}
\vspace{-2mm}
\end{minipage}
\begin{minipage}[t]{0.32\linewidth}
\centering
\includegraphics[width=1\textwidth]{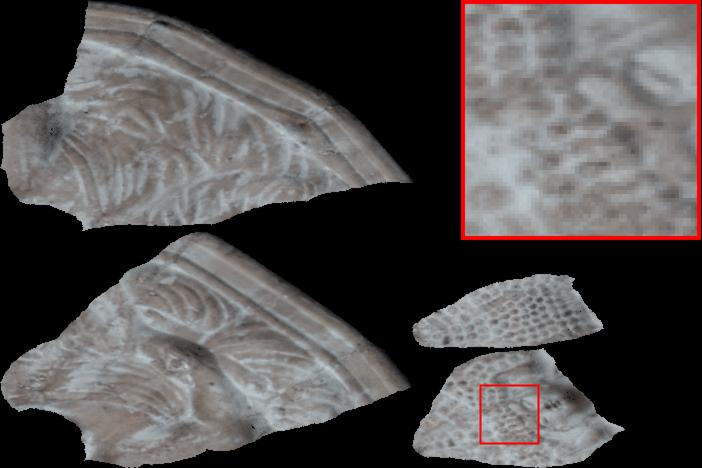}
{\footnotesize  Nearest}
\vspace{-2mm}
\end{minipage}
\begin{minipage}[t]{0.32\linewidth}
\centering
\includegraphics[width=1\textwidth]{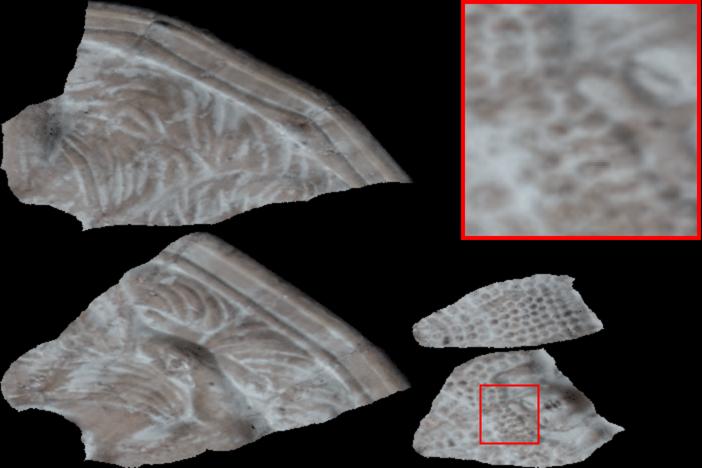}
{\footnotesize  Bilinear}
\vspace{-2mm}
\end{minipage}

\begin{minipage}[t]{0.32\linewidth}
\centering
\includegraphics[width=1\textwidth]{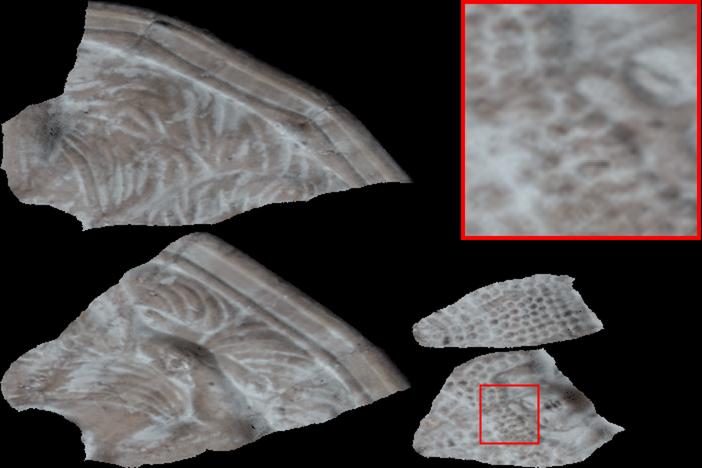}
{\footnotesize  Lanczos}
\end{minipage}
\begin{minipage}[t]{0.32\linewidth}
\centering
\includegraphics[width=1\textwidth]{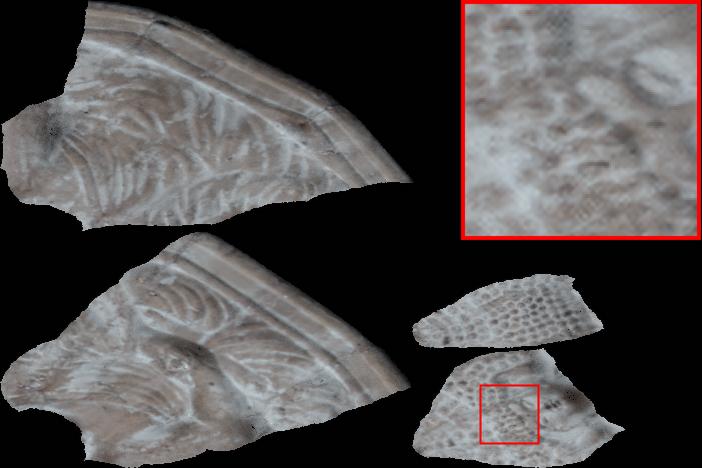}
{\footnotesize  FSRCNN}
\end{minipage}
\begin{minipage}[t]{0.32\linewidth}
\centering
\includegraphics[width=1\textwidth]{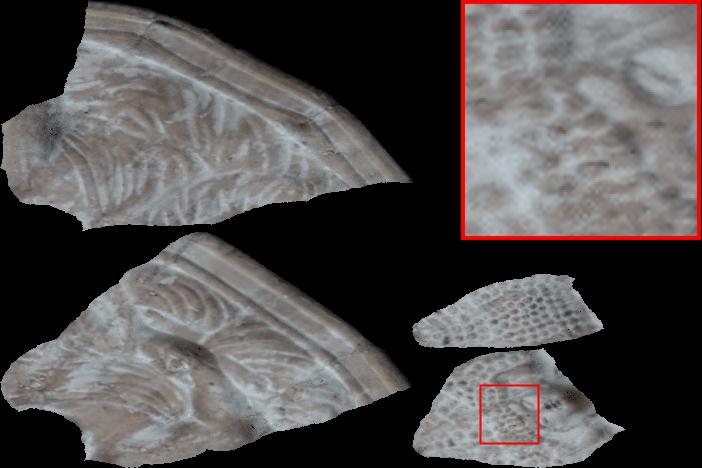}
{\footnotesize  SRResNet}
\end{minipage}

\begin{minipage}[t]{0.32\linewidth}
\centering
\includegraphics[width=1\textwidth]{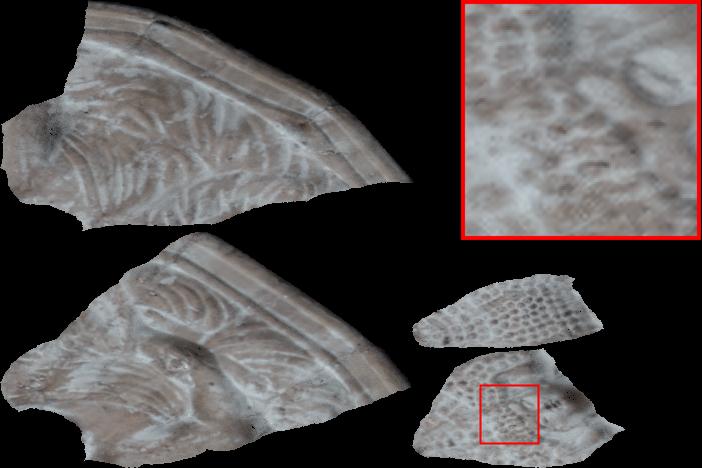}
{\footnotesize  EDSR}
\end{minipage}
\begin{minipage}[t]{0.32\linewidth}
\centering
\includegraphics[width=1\textwidth]{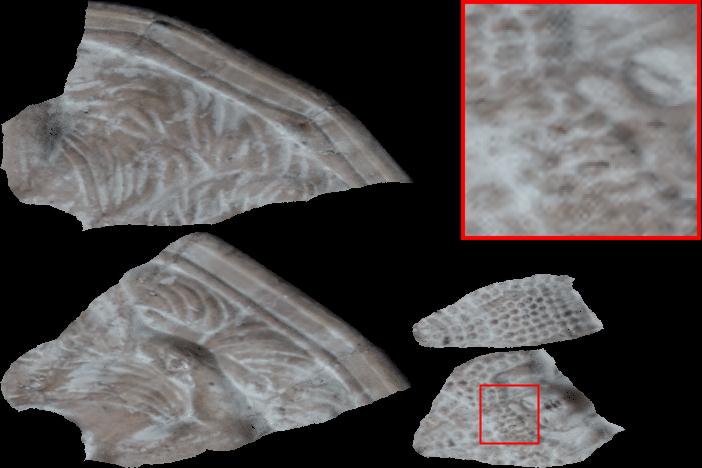}
{\footnotesize  RCAN}
\end{minipage}
\begin{minipage}[t]{0.32\linewidth}
\centering
\includegraphics[width=1\textwidth]{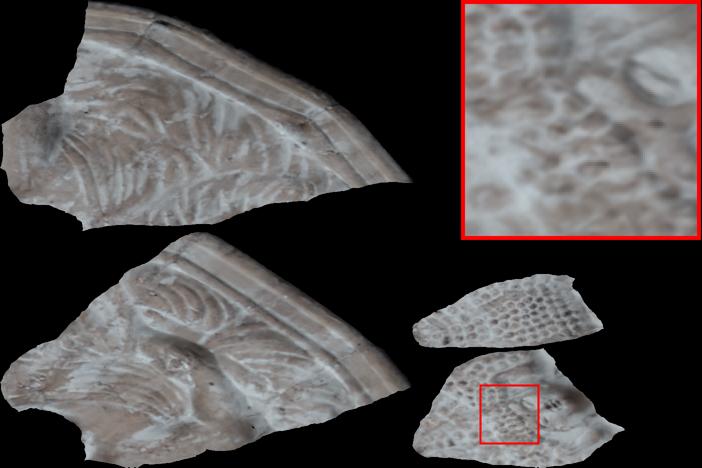}
{\footnotesize  EDSR-FT}
\end{minipage}

\begin{minipage}[t]{0.32\linewidth}
\centering
\includegraphics[width=1\textwidth]{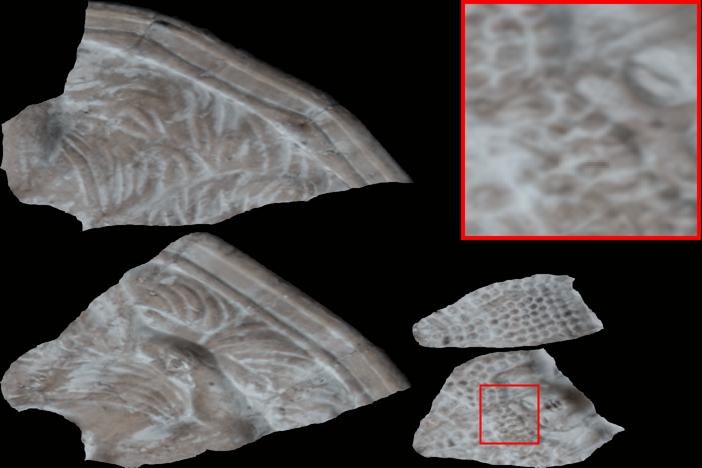}
{\footnotesize  NLR-Sub}
\end{minipage}
\begin{minipage}[t]{0.32\linewidth}
\centering
\includegraphics[width=1\textwidth]{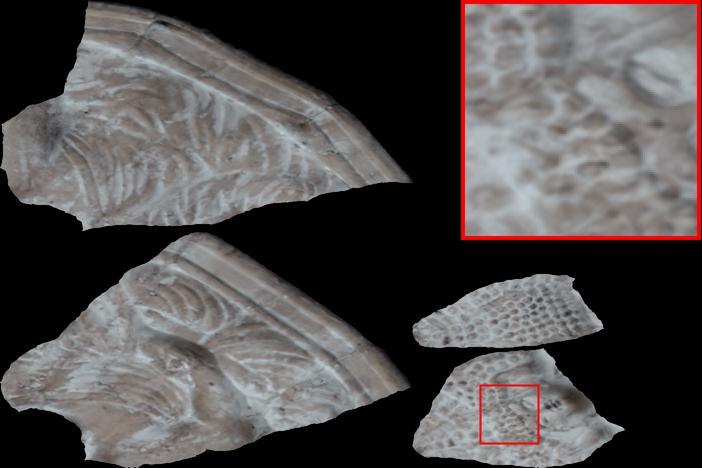}
{\footnotesize  NLR}
\end{minipage}
\begin{minipage}[t]{0.32\linewidth}
\centering
\includegraphics[width=1\textwidth]{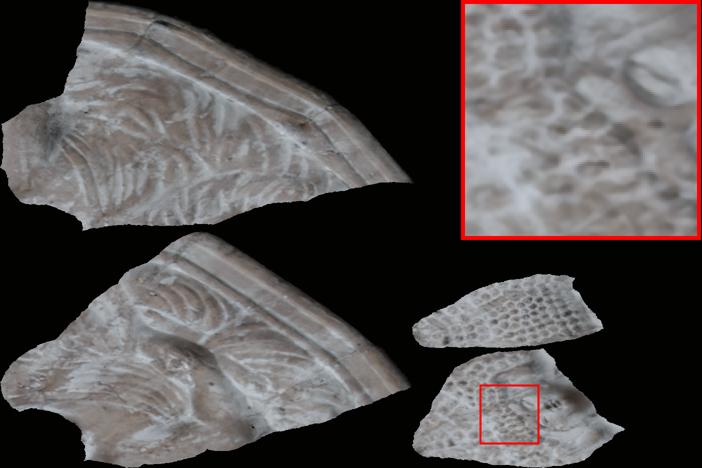}
{\footnotesize  NHR}
\end{minipage}
\caption{Texture map SR results of \emph{Buddha} by different methods.} 
\label{fig:sr_Buddha}
\end{figure*}

\begin{figure*}[t]
\centering
\begin{minipage}[t]{0.32\linewidth}
\centering
\includegraphics[width=1\textwidth]{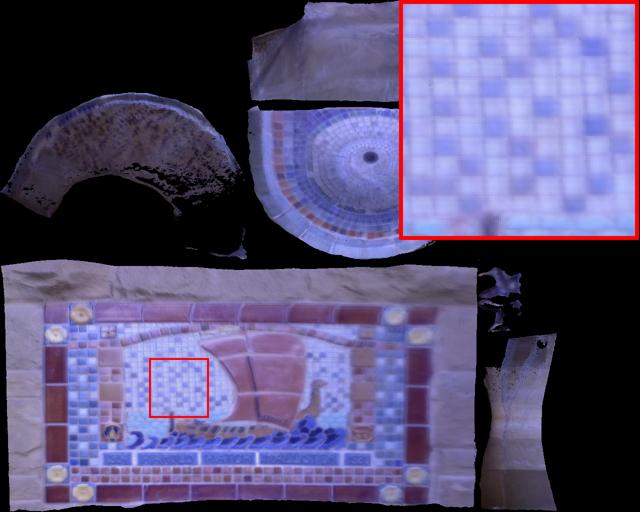}
{\footnotesize  Ground Truth}
\vspace{-2mm}
\end{minipage}
\begin{minipage}[t]{0.32\linewidth}
\centering
\includegraphics[width=1\textwidth]{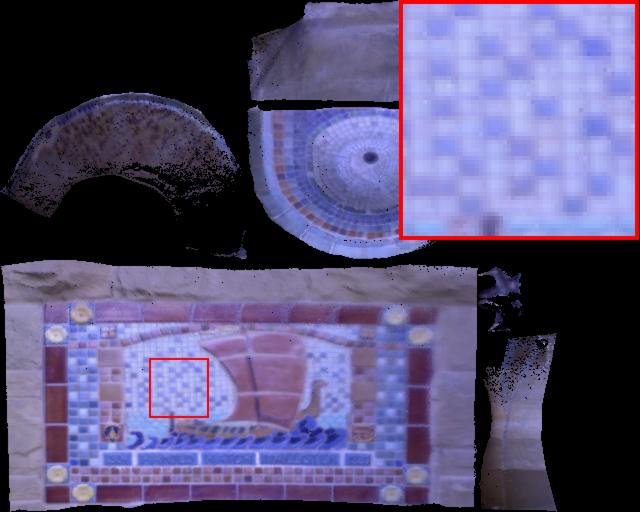}
{\footnotesize  Nearest}
\vspace{-2mm}
\end{minipage}
\begin{minipage}[t]{0.32\linewidth}
\centering
\includegraphics[width=1\textwidth]{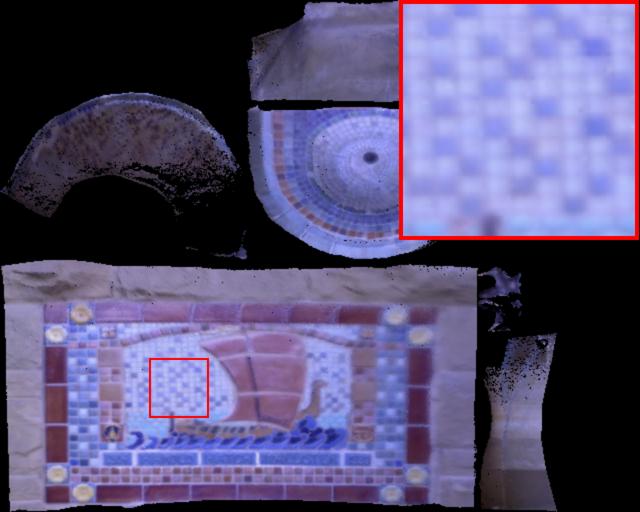}
{\footnotesize  Bilinear}
\vspace{-2mm}
\end{minipage}

\begin{minipage}[t]{0.32\linewidth}
\centering
\includegraphics[width=1\textwidth]{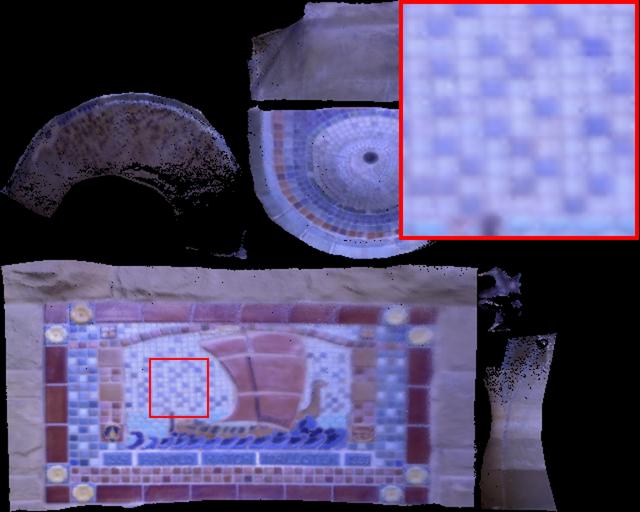}
{\footnotesize  Lanczos}
\end{minipage}
\begin{minipage}[t]{0.32\linewidth}
\centering
\includegraphics[width=1\textwidth]{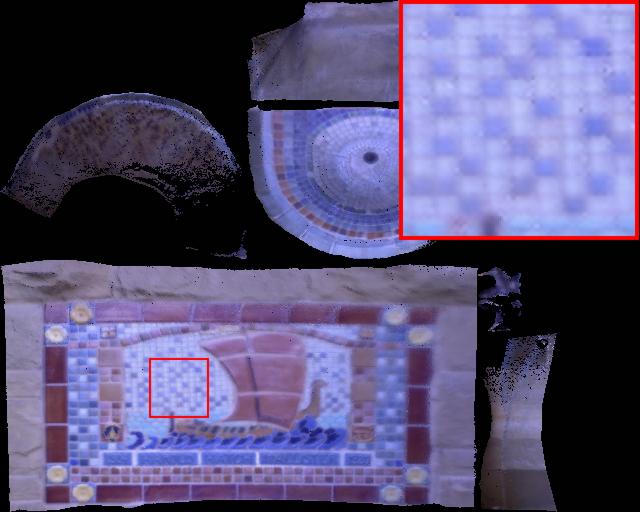}
{\footnotesize  FSRCNN}
\end{minipage}
\begin{minipage}[t]{0.32\linewidth}
\centering
\includegraphics[width=1\textwidth]{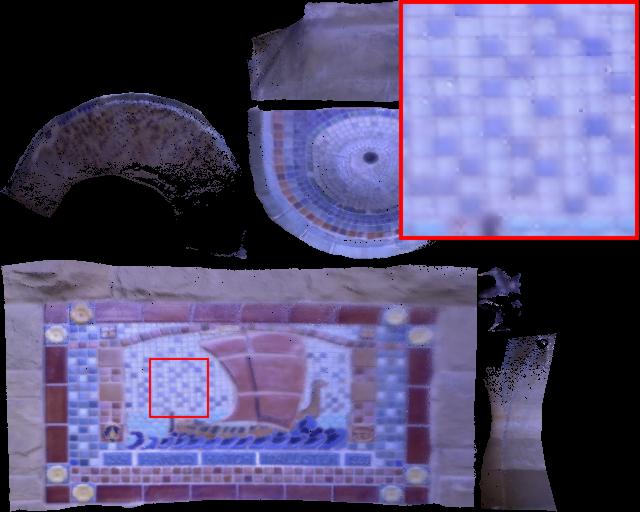}
{\footnotesize  SRResNet}
\end{minipage}

\begin{minipage}[t]{0.32\linewidth}
\centering
\includegraphics[width=1\textwidth]{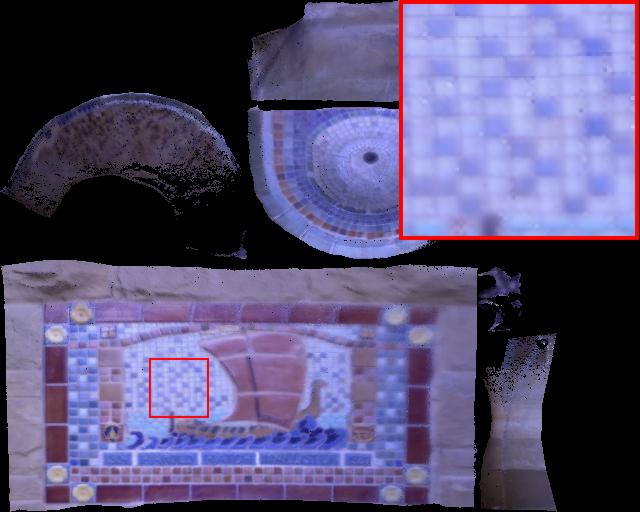}
{\footnotesize  EDSR}
\end{minipage}
\begin{minipage}[t]{0.32\linewidth}
\centering
\includegraphics[width=1\textwidth]{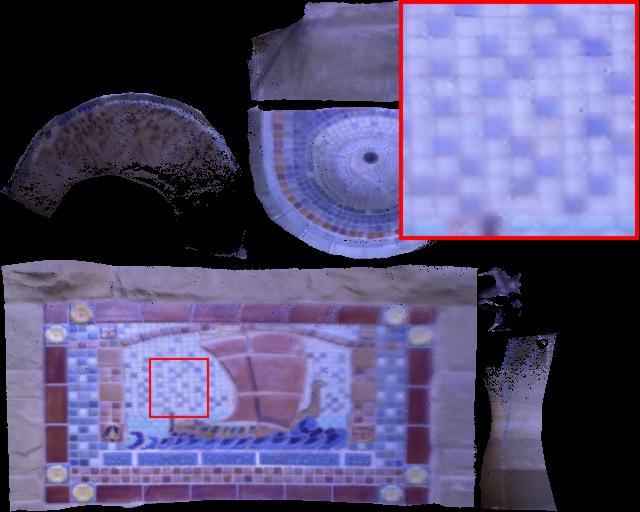}
{\footnotesize  RCAN}
\end{minipage}
\begin{minipage}[t]{0.32\linewidth}
\centering
\includegraphics[width=1\textwidth]{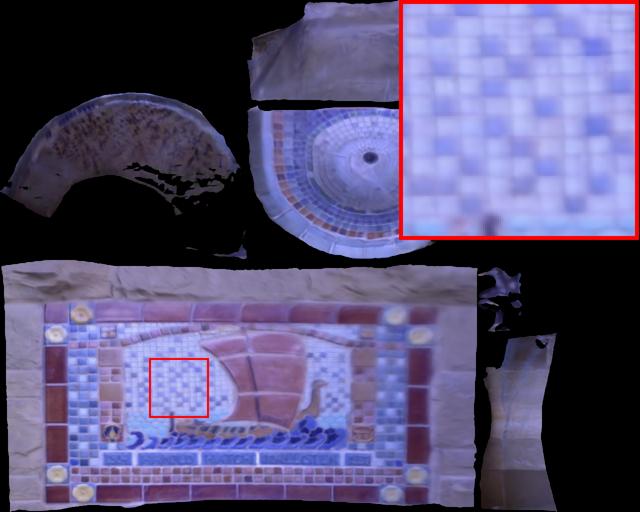}
{\footnotesize  EDSR-FT}
\end{minipage}

\begin{minipage}[t]{0.32\linewidth}
\centering
\includegraphics[width=1\textwidth]{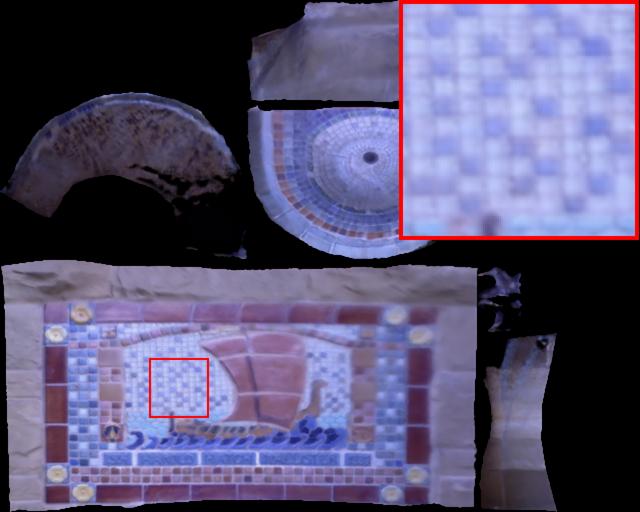}
{\footnotesize  NLR-Sub}
\end{minipage}
\begin{minipage}[t]{0.32\linewidth}
\centering
\includegraphics[width=1\textwidth]{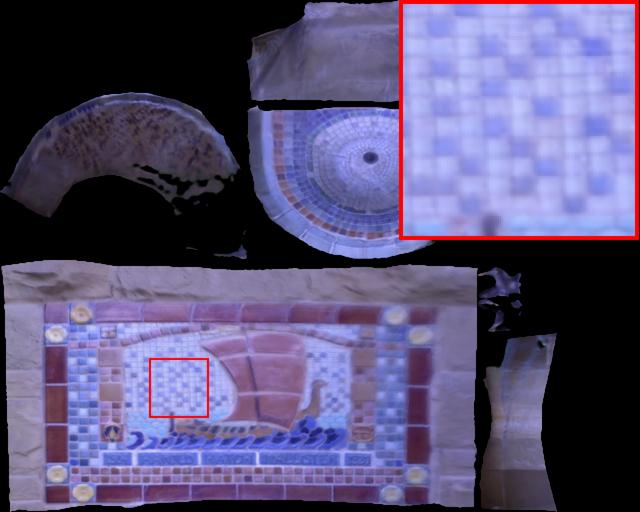}
{\footnotesize  NLR}
\end{minipage}
\begin{minipage}[t]{0.32\linewidth}
\centering
\includegraphics[width=1\textwidth]{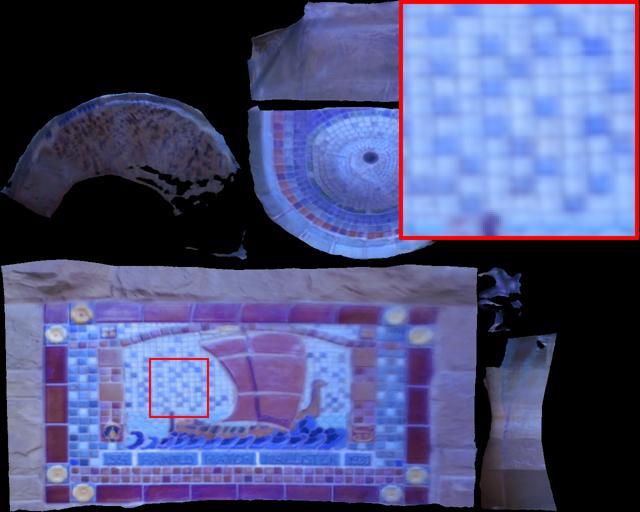}
{\footnotesize  NHR}
\end{minipage}
\caption{Texture map SR results of \emph{Fountain} by different methods.} 
\label{fig:sr_fountain}
\end{figure*}

In Table~\ref{tbl:psnr} and Table~\ref{tbl:ssim}, we show the PSNR and SSIM results of  different methods. Apart from the methods in the main paper, the results of FSRCNN~\cite{dong2016accelerating}, SRResNet~\cite{ledig2017photo}, and RCAN~\cite{zhang2018image} are also provided. For FSRCNN, the pre-trained models provided by the authors are directly used. SRResNet, EDSR, and RCAN are trained on DIV2K~\cite{Agustsson_2017_CVPR_Workshops}. More visual results of \emph{relief}, \emph{facade}, \emph{Buddha}, and \emph{Fountain} for different methods are shown in Fig.~\ref{fig:sr_relief}, Fig.~\ref{fig:sr_facade}, Fig.~\ref{fig:sr_Buddha}, and Fig.~\ref{fig:sr_fountain}, respectively.

\begin{table*}[ht]
  \caption{The PSNR results of different methods for scaling factor $\times 2$, $\times 3$, and $\times 4$.}
  \label{tbl:psnr}
  \centering
  \footnotesize
  \begin{tabular}{c||c|c|c||c|c|c||c|c|c||c|c|c||c|c|c}
    \hline  
    \multirow{2}{*}{Method} & \multicolumn{3}{c||}{ETH3D} & \multicolumn{3}{c||}{Collection} & \multicolumn{3}{c||}{Middlebury} & \multicolumn{3}{c||}{SyB3R} & \multicolumn{3}{c}{Average} \\ \cline{2-16}
    & $\times 2$ & $\times 3$ & $\times 4$ & $\times 2$ & $\times 3$ & $\times 4$ & $\times 2$ & $\times 3$ & $\times 4$ & $\times 2$ & $\times 3$ & $\times 4$ & $\times 2$ & $\times 3$ & $\times 4$  \\ \hline
    Nearest 	& 19.06 	& 16.71 	& 14.68 	& 24.22 	& 19.7 	& 16.92 	& 10.08 	& 7.93 	& 7.08 	& 30.84 	& 27.88 	& 25.82 	& 21.07 	& 18.12 	& 16.0 \\ \hline
Bilinear 	& 20.61 	& 18.24 	& 16.32 	& 26.2 	& 21.48 	& 18.84 	& 11.87 	& 8.88 	& 7.77 	& 31.75 	& 28.83 	& 26.9 	& 22.67 	& 19.6 	& 17.56 \\ \hline
Bicubic 	& 20.21 	& 17.96 	& 15.88 	& 25.67 	& 21.12 	& 18.29 	& 11.32 	& 8.81 	& 7.73 	& 31.77 	& 28.78 	& 26.73 	& 22.28 	& 19.34 	& 17.16 \\ \hline
Lanczos 	& 20.01 	& 17.74 	& 15.69 	& 25.42 	& 20.86 	& 18.07 	& 11.14 	& 8.81 	& 7.81 	& 31.71 	& 28.7 	& 26.63 	& 22.09 	& 19.15 	& 17.0 \\ \hline
HRST 	& 16.18 	& -- 	& 16.12 	& 32.29 	& -- 	& 29.63 	& 22.13 	& -- 	& 20.88 	& 27.9 	& -- 	& 26.34 	& 22.17 	& -- 	& 21.17 \\ \hline
HRST+ 	& -- 	& -- 	& -- 	& 32.24 	& -- 	& 29.9 	& 22.76 	& -- 	& 21.55 	& -- 	& -- 	& -- 	& -- 	& -- 	& -- \\ \hline
FSRCNN 	& 18.09 	& 15.02 	& 13.75 	& 23.58 	& 18.16 	& 16.36 	& 9.62 	& 7.73 	& 7.43 	& 30.23 	& 26.35 	& 25.01 	& 20.27 	& 16.61 	& 15.29 \\ \hline
SRRESNET 	& 17.61 	& 14.89 	& 12.83 	& 22.99 	& 18.1 	& 15.22 	& 8.92 	& 6.98 	& 6.5 	& 30.04 	& 26.88 	& 24.52 	& 19.79 	& 16.53 	& 14.36 \\ \hline
EDSR 	& 16.75 	& 14.08 	& 12.03 	& 21.77 	& 17.2 	& 14.24 	& 8.49 	& 7.13 	& 6.61 	& 29.31 	& 26.18 	& 23.81 	& 18.89 	& 15.79 	& 13.61 \\ \hline
RCAN 	& 16.32 	& 13.62 	& 11.55 	& 21.6 	& 16.73 	& 13.91 	& 8.4 	& 7.05 	& 6.54 	& 29.11 	& 25.86 	& 23.5 	& 18.58 	& 15.38 	& 13.22 \\ \hline
EDSR+ 	& 21.13 	& 19.75 	& 18.44 	& 28.25 	& 25.53 	& 24.19 	& 12.73 	& 11.21 	& 9.9 	& 32.78 	& 29.9 	& 28.31 	& 23.66 	& 21.75 	& 20.4 \\ \hline
NLR- 	& 21.21 	& 20.11 	& 19.2 	& 28.08 	& 25.0 	& 23.27 	& 14.68 	& 12.37 	& 11.11 	& 32.18 	& 28.84 	& 26.64 	& 23.75 	& 21.78 	& 20.47 \\ \hline
NLR 	& 21.31 	& 20.27 	& 19.18 	& 28.38 	& 25.85 	& 24.84 	& 13.67 	& 12.92 	& 12.29 	& 32.57 	& 29.57 	& 27.67 	& 23.85 	& 22.22 	& 21.08 \\ \hline
NHR 	& 25.19 	& 23.95 	& 22.7 	& 30.25 	& 28.41 	& 26.27 	& 17.16 	& 17.21 	& 15.63 	& 30.57 	& 27.42 	& 24.39 	& 26.46 	& 24.94 	& 23.22 \\ \hline
  \end{tabular}
  \vspace{-0.4cm}
\end{table*}

\begin{table*}[ht]
  \caption{The SSIM results of different methods for scaling factor $\times 2$, $\times 3$, and $\times 4$.}
  \label{tbl:ssim}
  \centering
  \footnotesize
  \begin{tabular}{c||c|c|c||c|c|c||c|c|c||c|c|c||c|c|c}
    \hline  
    \multirow{2}{*}{Method} & \multicolumn{3}{c||}{ETH3D} & \multicolumn{3}{c||}{Collection} & \multicolumn{3}{c||}{Middlebury} & \multicolumn{3}{c||}{SyB3R} & \multicolumn{3}{c}{Average} \\ \cline{2-16}

    & $\times 2$ & $\times 3$ & $\times 4$ & $\times 2$ & $\times 3$ & $\times 4$ & $\times 2$ & $\times 3$ & $\times 4$ & $\times 2$ & $\times 3$ & $\times 4$ & $\times 2$ & $\times 3$ & $\times 4$  \\ \hline
Nearest	& 0.81	& 0.74	& 0.68	& 0.88	& 0.77	& 0.7	& 0.54	& 0.43	& 0.39	& 0.91	& 0.84	& 0.79	& 0.82	& 0.74	& 0.67 \\ \hline
Bilinear	& 0.83	& 0.77	& 0.71	& 0.91	& 0.81	& 0.75	& 0.57	& 0.45	& 0.41	& 0.92	& 0.86	& 0.81	& 0.84	& 0.77	& 0.71 \\ \hline
Bicubic	& 0.83	& 0.76	& 0.7	& 0.9	& 0.8	& 0.73	& 0.56	& 0.45	& 0.41	& 0.92	& 0.86	& 0.81	& 0.84	& 0.76	& 0.7 \\ \hline
Lanczos	& 0.82	& 0.75	& 0.68	& 0.9	& 0.79	& 0.71	& 0.55	& 0.45	& 0.4	& 0.92	& 0.86	& 0.81	& 0.83	& 0.75	& 0.68 \\ \hline
HRST	& 0.67	& --	& 0.66	& 0.96	& --	& 0.93	& 0.92	& --	& 0.9	& 0.88	& --	& 0.82	& 0.79	& --	& 0.77 \\ \hline
HRST+	& --	& --	& --	& 0.95	& --	& 0.92	& 0.91	& --	& 0.89	& --	& --	& --	& --	& --	& -- \\ \hline
FSRCNN	& 0.77	& 0.66	& 0.61	& 0.85	& 0.7	& 0.64	& 0.49	& 0.37	& 0.36	& 0.9	& 0.81	& 0.77	& 0.78	& 0.66	& 0.61 \\ \hline
SRRESNET	& 0.78	& 0.7	& 0.63	& 0.86	& 0.74	& 0.66	& 0.5	& 0.41	& 0.38	& 0.91	& 0.84	& 0.78	& 0.79	& 0.7	& 0.64 \\ \hline
EDSR	& 0.76	& 0.67	& 0.6	& 0.82	& 0.71	& 0.64	& 0.49	& 0.41	& 0.37	& 0.91	& 0.83	& 0.77	& 0.77	& 0.68	& 0.61 \\ \hline
RCAN	& 0.75	& 0.66	& 0.58	& 0.84	& 0.7	& 0.63	& 0.49	& 0.4	& 0.37	& 0.9	& 0.83	& 0.77	& 0.77	& 0.67	& 0.6 \\ \hline
EDSR+	& 0.86	& 0.83	& 0.79	& 0.94	& 0.91	& 0.88	& 0.62	& 0.51	& 0.45	& 0.93	& 0.88	& 0.84	& 0.87	& 0.83	& 0.79 \\ \hline
NLR-	& 0.86	& 0.82	& 0.74	& 0.93	& 0.89	& 0.85	& 0.66	& 0.51	& 0.42	& 0.93	& 0.86	& 0.82	& 0.87	& 0.82	& 0.75 \\ \hline
NLR	& 0.86	& 0.83	& 0.8	& 0.94	& 0.91	& 0.89	& 0.65	& 0.58	& 0.54	& 0.93	& 0.87	& 0.83	& 0.87	& 0.83	& 0.8 \\ \hline
NHR	& 0.84	& 0.89	& 0.84	& 0.87	& 0.93	& 0.85	& 0.74	& 0.77	& 0.72	& 0.82	& 0.85	& 0.72	& 0.84	& 0.88	& 0.82 \\ \hline
  \end{tabular}
\end{table*}

\end{document}